\documentclass{article}
\usepackage{jfrExamplee}
\usepackage{graphicx}
\usepackage{apalike}
\usepackage{setspace}

\usepackage{hyperref}
\hypersetup{bookmarksopen,bookmarksnumbered,
pdfpagemode=UseOutlines,
colorlinks=true,
linkcolor=blue,
anchorcolor=blue,
citecolor=blue,
filecolor=blue,
menucolor=blue,
urlcolor=blue
}
\usepackage{upgreek}
\usepackage{amsfonts,amssymb}
\usepackage{bm}
\usepackage{bbm}

\usepackage{amsthm} 
\usepackage{thmtools}
\usepackage{mathtools}
\usepackage{xspace}
\usepackage[font=footnotesize]{subcaption}
\usepackage{units}
\usepackage{booktabs} 
\usepackage{tabularx}
    \newcolumntype{L}{>{\raggedright\arraybackslash}X}
\usepackage{tabulary}
\usepackage[usenames,dvipsnames,table]{xcolor} 
\usepackage{diagbox}
\usepackage[ruled,vlined,linesnumbered]{algorithm2e}  
\usepackage{parskip}
\SetKwComment{Comment}{$\triangleright$\ }{}
\usepackage{ifthen,version}
\usepackage{soul}
\usepackage{csquotes}
\usepackage{microtype}      
\usepackage{lipsum}
\usepackage{tikz}

\usepackage{todonotes}
\usepackage{enumitem}
\usepackage{cite}
\usepackage{gensymb}
\usepackage{footmisc}
\usepackage[title]{appendix}

\newcommand{\xxnote}[3]{}
\ifx\hidenotes\undefined
  \usepackage{color}
  \renewcommand{\xxnote}[3]{\color{#2}{#1: #3}}
\fi

\usepackage{multirow}
\usepackage{pbox}

\usepackage{algorithmic}

\newtheoremstyle{hypstyle}
{3pt} 
{3pt} 
{\itshape} 
{} 
{\bfseries} 
{.} 
{.5em} 
{} 

\theoremstyle{hypstyle}

\DeclareMathOperator*{\argmin}{arg\,min}
\DeclareMathOperator*{\argmax}{arg\,max}

\newcommand{\argmaxprob}[1]{\underset{#1}{\argmax}}
\newcommand{\argminprob}[1]{\underset{#1}{\argmin}}

\newcommand{\abs}[1]{\left|#1 \right|}

\newcommand{\expect}[2]{\mathbb{E}_{#1}\left[#2\right]}

\newcommand{\real}[0]{\mathbb{R}}

\newcommand{\bbm}{\begin{bmatrix}}
\newcommand{\ebm}{\end{bmatrix}}

\newcommand{\Path}[1]{\xi_{#1}}
\newcommand{\cost}[0]{J}
\newcommand{\costFn}[1]{\cost \left( #1 \right)}

\newcommand{\costShot}[0]{\cost_\mathrm{shot}}
\newcommand{\costFnShot}[1]{\costShot \left( #1 \right)}

\newcommand{\costSmooth}[0]{\cost_\mathrm{smooth}}
\newcommand{\costFnSmooth}[1]{\costSmooth \left( #1 \right)}

\newcommand{\costOcc}[0]{\cost_\mathrm{occ}}
\newcommand{\costFnOcc}[1]{\costOcc \left( #1 \right)}

\newcommand{\costObs}[0]{\cost_\mathrm{obs}}
\newcommand{\costFnObs}[1]{\costObs \left( #1 \right)}

\newcommand{\map}[0]{\mathcal{M}}
\newcommand{\grid}[0]{\mathcal{G}}
\newcommand{\context}[0]{\mathcal{C}}

\graphicspath{{figs/}}

\title{Autonomous Aerial Cinematography In Unstructured Environments With Learned Artistic Decision-Making}

\author{
Rogerio Bonatti \\
Robotics Institute\\
Carnegie Mellon University\\
Pittsburgh, PA 15213 \\
\texttt{rbonatti@cs.cmu.edu} \\
\And
Wenshan Wang \\
Robotics Institute\\
Carnegie Mellon University\\
Pittsburgh, PA 15213 \\
\texttt{wenshanw@andrew.cmu.edu} \\
\And
Cherie Ho \\
Robotics Institute\\
Carnegie Mellon University\\
Pittsburgh, PA 15213 \\
\texttt{cherieh@cs.cmu.edu} \\
\And
Aayush Ahuja \\
Robotics Institute\\
Carnegie Mellon University\\
Pittsburgh, PA 15213 \\
\texttt{aahuja2@andrew.cmu.edu} \\
\And
Mirko Gschwindt \\
Department of Computer Science\\
Technische Universit{\"a}t M{\"u}nchen\\
Munich, Germany \\
\texttt{m.gschwindt@tum.de} \\
\And
Efe Camci \\
School of Mechanical and\\
Aerospace Engineering\\
Nanyang Technological University\\
639798, Singapore \\
\texttt{efe001@e.ntu.edu.sg} \\
\And
Erdal Kayacan \\
Department of Engineering\\
Aarhus University\\
DK-8000 Aarhus C, Denmark \\
\texttt{erdal@eng.au.dk} \\
\And
Sanjiban Choudhury \\
School of Computer Science\\
University of Washington\\
Seattle, WA 98195 \\
\texttt{sanjibac@cs.uw.edu} \\
\And
Sebastian Scherer \\
Robotics Institute\\
Carnegie Mellon University\\
Pittsburgh, PA 15213 \\
\texttt{basti@cs.cmu.edu} \\
}

\begin{document}







\maketitle


\begin{abstract}

Aerial cinematography is revolutionizing industries that require live and dynamic camera viewpoints such as entertainment, sports, and security. However, safely piloting a drone while filming a moving target in the presence of obstacles is immensely taxing, often requiring multiple expert human operators. Hence, there is demand for an autonomous cinematographer that can reason about both geometry and scene context in real-time. 
Existing approaches do not address all aspects of this problem; they either require high-precision motion-capture systems or GPS tags to localize targets, rely on prior maps of the environment, plan for short time horizons, or only follow artistic guidelines specified before flight.
\\\\
In this work, we address the problem in its entirety and propose a complete system for real-time aerial cinematography that for the first time combines: (1) vision-based target estimation; (2) 3D signed-distance mapping for occlusion estimation; (3) efficient trajectory optimization for long time-horizon camera motion; and (4) learning-based artistic shot selection. We extensively evaluate our system both in simulation and in field experiments by filming dynamic targets moving through unstructured environments. Our results indicate that our system can operate reliably in the real world without restrictive assumptions.
We also provide in-depth analysis and discussions for each module, with the hope that our design tradeoffs can generalize to other related applications. Videos of the complete system can be found at: \small \href{https://youtu.be/ookhHnqmlaU}{https://youtu.be/ookhHnqmlaU}\normalsize.

\end{abstract}


\section{Introduction}



\begin{figure}[t]
    \center
    \includegraphics[width=0.95\textwidth]{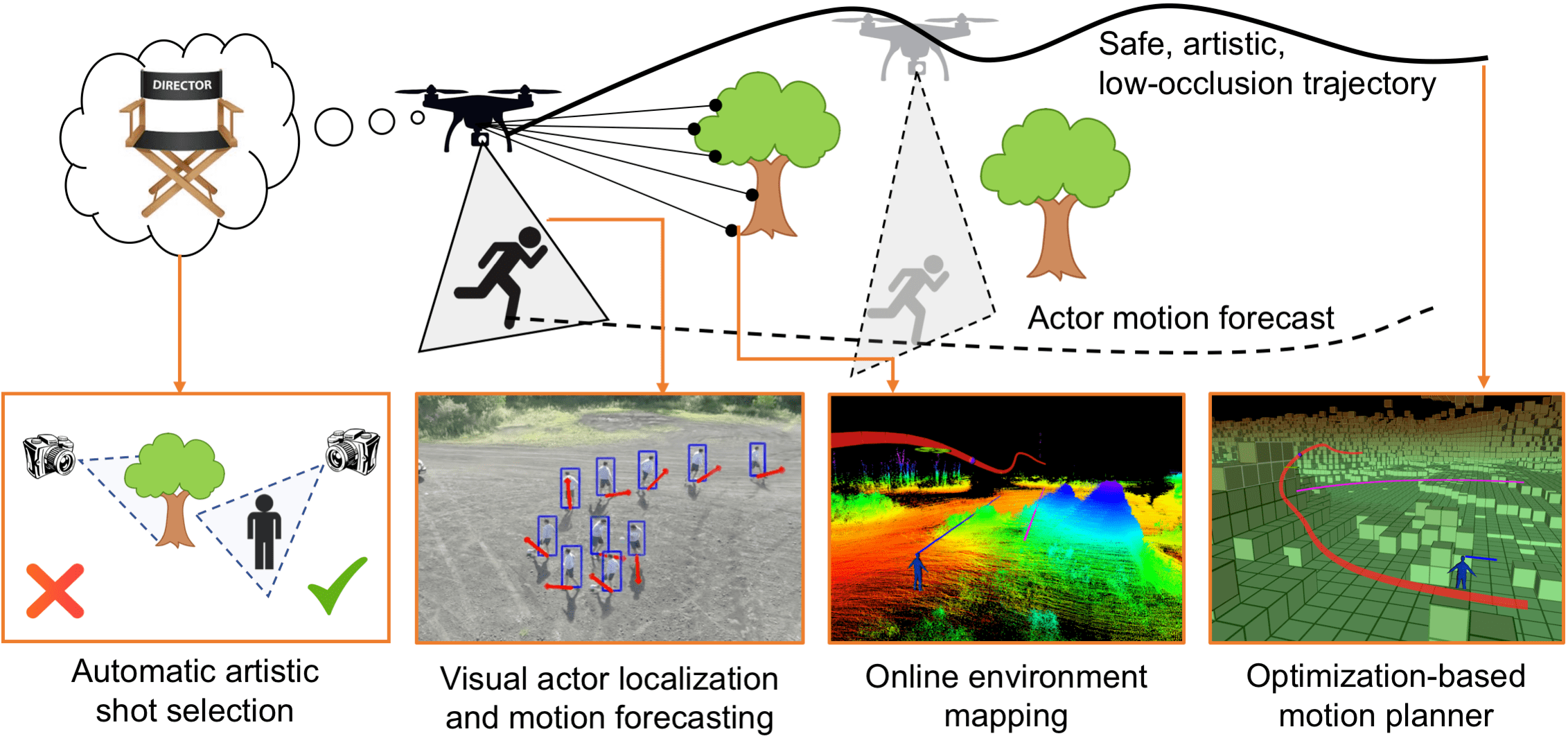}
    \caption{Aerial cinematographer pipeline: the UAV visually detects the actor's motion using a vision-based localization module, maps the environment with an onboard LiDAR sensor, reasons about artistic guidelines, and plans a smooth, collision-free trajectory while avoiding occlusions.
    \label{fig:intro}}
\end{figure}

Manually-operated unmanned aerial vehicles (UAVs) are drastically improving efficiency and productivity in a diverse set of industries and economic activities. In particular, tasks that require dynamic camera viewpoints have been most affected, where the development of small scale UAVs has alleviated the need for sophisticated hardware to manipulate cameras in space. For instance, in the movie industry, drones are changing the way both professional and amateur film-makers can capture shots of actors and landscapes by allowing the composition of aerial viewpoints which are not feasible using traditional devices such as hand-held cameras and dollies \cite{santamarina2018introduction}. In the sports domain, flying cameras can track fast-moving athletes and accompany dynamic movements \cite{la2016drones}. Furthermore, flying cameras show a largely unexplored potential for tracking subjects of interest in security applications \cite{de2018ethics}, which are not possible with static sensors. 

However, manually-operated UAVs often require multiple expert pilots due to the difficulty of executing all necessary perception and motion planning tasks synchronously: it takes high attention and effort to simultaneously identify the actor(s), predict how the scene is going to evolve, control the UAV, avoid obstacles and reach the desired viewpoints. Hence the need for an autonomous aerial cinematography system.

The distinctive challenge of developing an autonomous aerial cinematography system is the need to tightly couple \emph{contextual} and \emph{geometric} threads. 
Contextual reasoning involves processing camera images to detect the actor, understanding how the scene is going to evolve, and selecting desirable viewpoints.
Geometric reasoning considers the 3D configuration of objects in the environment to evaluate the visibility quality of a particular viewpoint and whether the UAV can reach it in a safe manner.
Although these two threads differ significantly in terms of sensing modalities, computational representation and computational complexity, both sides play a vital role when addressing the entirety of the autonomous filming problem.
In this work we present a complete system that combines both threads in a cohesive and principled manner.
In order to develop our autonomous cinematographer, we address several key challenges.

\subsection{Challenges}
\label{subsec:challenges}

Consider a typical filming scenario in Figure~\ref{fig:intro}. The UAV must overcome several challenges: 



\paragraph{Actor pose estimation with challenging visual inputs:} The UAV films a dynamic actor from various angles, therefore it is critical to accurately localize the actor's position and orientation in a 3D environment. In practice, the use of external sensors such as motion capture systems \cite{nageli2017real} and GPS tags \cite{bonatti2018autonomous,joubert2016towards} for pose estimation is highly impractical; a robust system should only rely on visual localization. The challenge is to deal with all possible viewpoints, scales, backgrounds, lighting conditions, motion blur caused by both the dynamic actor and camera.

\paragraph{Operating in unstructured scenarios:} The UAV flies in diverse, unstructured environments without prior information. In a typical mission, it follows an actor across varying terrain and obstacle types, such as slopes, mountains, and narrow trails between trees or buildings. The challenge is to maintain an online map that has a high enough resolution to reason about viewpoint occlusions and that updates itself fast enough to keep the vehicle safe.




\paragraph{Keep actor visible while staying safe:} In cinematography, the UAV must maintain visibility of the actor for as long as possible while staying safe in a partially known environment. When dealing with dynamic targets, the UAV must anticipate the actor's motion and reason about collisions and occlusions generated by obstacles in potential trajectories. The challenge of visibility has been explored in previous works in varying degrees of obstacle complexity \cite{penin2018vision,nageli2017real,galvane2018directing,bonatti2019towards}.


 \paragraph{Understanding scene context for autonomous artistic decision-making:} When filming a movie, the director actively selects the camera pose based on the actor's movement, environment characteristics and intrinsic artistic values. Although humans can make such aesthetic decisions implicitly, it is challenging to define explicit rules to define the ideal artistic choices for a given context.


\paragraph{Making real-time decisions with onboard resources:} Our focus is on unscripted scenarios where shots are decided on the fly; all algorithms must run in real-time with limited computational resources.

\subsection{Contributions}

Our paper revolves around two key ideas. First, we design differentiable objectives for camera motion that can be efficiently optimized for long time horizons. We architect our system to compute these objectives efficiently online to film a dynamic actor. Second, we apply learning to elicit human artistic preferences in selecting a sequence of shots. Specifically, we offer the following contributions:

\begin{enumerate}
  \item We propose a method for visually localizing the actor's position, orientation, and forecasting their future trajectory in the world coordinate frame. We present a novel semi-supervised approach that uses temporal continuity in sequential data for the heading direction estimation problem (Section~\ref{sec:vision});
  \item We propose an incremental signed distance transform algorithm for large-scale real-time environment mapping using a range sensor, \textit{e.g.,} LiDAR (Section~\ref{sec:mapping});
  \item We formalize the aerial filming motion planning problem following cinematography guidelines for arbitrary types of shots and arbitrary obstacle shapes. We propose an efficient optimization-based motion planning method that exploits covariant gradients and Hessians of the objective functions for fast convergence (Section~\ref{sec:planning});
  \item We propose a deep reinforcement learning method that incorporates human aesthetic preferences for artistic reasoning to act as an autonomous movie director, considering the current scene context to select the next camera viewpoints (Section~\ref{sec:artistic});
  \item We offer extensive quantitative and qualitative performance evaluations both for our integrated system and for each module, both in simulation and field tests (Section~\ref{sec:experiments}), along with detailed discussions on experimental lessons learned (Section~\ref{sec:discussion}).
\end{enumerate}

This paper builds upon our previous works that each focuses on an individual component of our framework: visual actor detection, tracking and heading estimation \cite{wang2019heading}, online environment mapping \cite{bonatti2019towards}, motion planning for cinematography \cite{bonatti2018autonomous}, and autonomous artistic viewpoint selection \cite{gschwindt2019can}. In this paper, for the first time we present a detailed description of the unified architecture (Section~\ref{sec:system_arch}), provide implementation details of the entire framework, and offer extensive flight test evaluations of the complete system.

\section{Problem Definition}
\label{sec:problem_form}









The overall task is to control a UAV to film an actor who is moving through an unknown environment.
Let $\Path{q}(t) : [0,t_f] \rightarrow \real^3  \times SO(2)$ be the trajectory of the UAV as a mapping from time to a position and heading, i.e., $\Path{q}(t) = \{x_{q}(t), y_{q}(t), z_{q}(t), \psi_{q}(t)\}$. 
Analogously, let $\Path{a}(t) : [0,t_f] \rightarrow \real^3  \times SO(2)$ be the trajectory of the actor: $\Path{a}(t) = \{x_{a}(t), y_{a}(t), z_{a}(t), \psi_{a}(t)\}$.
In our work, a instantaneous measurement of the actor state $S_{a} : \real^3  \times SO(2)$ is obtained using onboard sensors (monocular camera and LiDAR, as seen in Section~\ref{sec:vision}), but external sensors and motion capture systems could also be employed (Section~\ref{sec:related_work}). Measurements $S_{a}$ are continuously fed into a prediction module that computes $\Path{a}$ (Section~\ref{sec:vision}). 

The UAV also needs to store a representation of the environment. Let grid $\grid: \real^3 \rightarrow [0,1]$ be a voxel occupancy grid that maps every point in space to a probability of occupancy. Let $\map : \real^3 \rightarrow \real$ be the signed distance value of a point to the nearest obstacle. Positive signs are for points in free space, and negative signs are for points either in occupied or unknown space, which we assume to be potentially inside an obstacle. During flight the UAV senses the environment with the onboard LiDAR, updates grid $\grid$, and then updates $\map$ (more details at Section~\ref{sec:mapping}).

We can generically formulate a motion planning problem that aims to minimize a particular cost function $\costFn{\Path{q}}$ for cinematography. Within the filming context, this cost function measures jerkiness of motion, safety, environmental occlusion of the actor and shot quality (artistic quality of viewpoints). This cost function depends on the environment $\grid$ and $\map$, and on the actor forecast $\Path{a}$, all of which are sensed on-the-fly. The changing nature of environment and $\Path{a}$ demands re-planning at a high frequency.

Here we briefly touch upon the four components of the cost function $\costFn{\Path{q}}$ (refer to Section~\ref{sec:planning} for details and mathematical expressions):

\begin{enumerate}
  \item \emph{Smoothness} $\costFnSmooth{\Path{q}}$: Penalizes jerky motions that may lead to camera blur and unstable flight;
  \item \emph{Safety} $\costFnObs{\Path{q},\map}$: Penalizes proximity to obstacles that are unsafe for the UAV;
  \item \emph{Occlusion} $\costFnOcc{\Path{q},\Path{a},\map}$: Penalizes occlusion of the actor by obstacles in the environment;
  \item \emph{Shot quality} $\costFnShot{\Path{q},\Path{a},\Omega_\mathrm{art}}$: Penalizes poor viewpoint angles and scales that deviate from the desired artistic guidelines, given by the set of parameters $\Omega_\mathrm{art}$.
\end{enumerate}

In its simplest form, we can express $\costFn{\Path{q}}$ as a linear composition of each individual cost, weighted by scalars $\lambda_i$. The objective is to minimize $\costFn{\Path{q}}$ subject to initial boundary constraints $\Path{q}(0)$. The solution $\Path{q}^*$ is then tracked by the UAV:

\begin{equation}
\begin{aligned}
\label{eq:main_cost}
\costFn{\Path{q}} &=  \begin{bmatrix}
       1 & \lambda_1 & \lambda_2 & \lambda_3
     \end{bmatrix} 
\begin{bmatrix}
       \costFnSmooth{\Path{q}} \\
       \costFnObs{\Path{q},\map} \\
       \costFnOcc{\Path{q},\Path{a},\map} \\
       \costFnShot{\Path{q},\Path{a},\Omega_{art}}
     \end{bmatrix} \\
\Path{q}^* &= \argminprob{\Path{q}} \quad \costFn{\Path{q}}, \quad \text{s.t.} \; \Path{q}(0) = \{x_0,y_0,z_0,\psi_0\}
\end{aligned}
\end{equation}

We describe in Section~\ref{sec:planning} how (\ref{eq:main_cost}) is solved. The parameters $\Omega_\mathrm{art}$ of the shot quality term $\costShot$ are usually specified by the user prior to takeoff, and assumed constant throughout flight. For instance, based on the terrain characteristics and the type of motion the user expects the actor to do, they may specify a frontal or circular shot with a particular scale to be the best artistic choice for that context. Alternatively, $\Omega_\mathrm{art}$ can change dynamically, either by user's choice or algorithmically.


A dynamically changing $\Omega_\mathrm{art}$ leads to a new challenge: the UAV must make choices that maximize the artistic value of the incoming visual feed. As explained further in Section~\ref{sec:artistic}, artistic choices affect not only the immediate images recorded by the UAV. By changing the positioning of the UAV relative to the subject and obstacles, current choices influence the images captured in future time steps. Therefore, the selection of $\Omega_\mathrm{art}$ needs to be framed as a sequential decision-making process.

Let $v_t = \{I_1,I_2,...,I_k\}$ be a sequence of $k$ observed images captured by the UAV during time step $t$ between consecutive artistic decisions. 
Let $R_\mathrm{art}(v_t)$ be the  user's implicit evaluation reward based on the observed the video segment $v_t$.
The user's choice of an optimal artistic parameter sequence $\{ \Omega_{1}^*, \Omega_{2}^*, ..., \Omega_{n}^* \}$ can be interpreted as an optimization of the following form:

\begin{equation}
\label{eq:art_mapping}
\{ \Omega_{1}^*, \Omega_{2}^*, ..., \Omega_{n}^* \} = \argmaxprob{\{ \Omega_{1}, \Omega_{2}, ..., \Omega_{n} \}} \quad \sum_{t}^{} R_\mathrm{art}(v_{t})
\end{equation}

The optimization from Equation~\ref{eq:art_mapping} is usually left up to the UAV operator's experience and intuition. In Section~\ref{sec:artistic}, we detail a novel method for implicitly learning the selection of artistic parameters depending on the scene's context.

\section{Related Work} 
\label{sec:related_work}






Our work exploits synergies at the confluence of several domains of research to develop an aerial cinematography platform that can follow dynamic targets in unstructured environments using onboard sensors and computing power. Next, we describe related works in different areas that come together under the problem definition described in Section~\ref{sec:problem_form}.

\paragraph{Virtual cinematography:} Camera control in virtual cinematography has been extensively examined by the computer graphics community as reviewed by \cite{christie2008camera}. These methods typically reason about the utility of a viewpoint in isolation, follow artistic principles and composition rules \cite{arijon1976grammar,bowen2013grammar} and employ either optimization-based approaches to find good viewpoints or reactive approaches to track the virtual actor. The focus is typically on through-the-lens control where a virtual camera is manipulated while maintaining focus on certain image features~\cite{gleicher1992through,drucker1994intelligent,lino2011director,lino2015intuitive}.
However, virtual cinematography is free of several real-world limitations such as robot physics constraints and assumes full knowledge of the environment.

Several works analyse the choice of which viewpoint to employ for a particular situation. For example, in
\cite{drucker1994intelligent}, the researchers use an A* planner to move a virtual camera in pre-computed indoor simulation scenarios to avoid collisions with obstacles in 2D. 
More recently, we find works such as \cite{leake2017computational} that 
post-processes videos of a scene taken from different angles
by automatically labeling features of different views. The approach uses high-level user-specified rules which exploit the labels to automatically select the optimal sequence of viewpoints for the final movie. In addition, \cite{wu2018thinking} help editors by defining a formal language of editing patterns for movies.

\paragraph{Autonomous aerial cinematography:}
There is a rich history of work in autonomous aerial filming. 
For instance, several works focus on following user-specified artistic guidelines~\cite{joubert2016towards,nageli2017real,galvane2017automated,galvane2018directing} but often rely on perfect actor localization through a high-precision RTK GPS or a motion-capture system.
Additionally, although the majority of work in the area deals with collisions between UAV and actors\cite{nageli2017real,joubert2016towards,huang2018act}, they do not factor in the environment for safety considerations.
While there are several successful commercial products, they too have certain limitations such as operating in low speed and low clutter regimes  (e.g. DJI Mavic~\cite{mavic2019}) or relatively short planning horizons (e.g. Skydio R1~\cite{skydio2018}). 
Even our previous work~\cite{bonatti2018autonomous}, despite handling environmental occlusions and collisions, assumes a prior elevation map and uses GPS to localize the actor. 
Such simplifications impose restrictions on the diversity of scenarios that the system can handle. 

Several contributions on aerial cinematography focus on keyframe navigation. \cite{roberts2016generating,joubert2015interactive,gebhardt2018optimizing,gebhardt2016airways,xie2018creating} provide user interface tools to re-time and connect static aerial viewpoints to provide smooth and dynamically feasible trajectories, as well as a visually pleasing images. \cite{lan2017xpose} use key-frames defined on the image itself instead of world coordinates.

Other works focus on tracking dynamic targets, and employ a diverse set of techniques for actor localization and navigation. For example, \cite{huang2018act,huang2018through} detect the skeleton of targets from visual input, while others approaches rely on off-board actor localization methods from either motion-capture systems or GPS sensors \cite{joubert2016towards,galvane2017automated,nageli2017real,galvane2018directing,bonatti2018autonomous}. These approaches have varying levels of complexity: \cite{bonatti2018autonomous,galvane2018directing} can avoid obstacles and occlusions with the environment and with actors, while other approaches only handle collisions and occlusions caused by actors. In addition, in our latest work \cite{bonatti2019towards} we made two important improvements on top of \cite{bonatti2018autonomous} by including visual actor localization and online environment mapping.

Specifically on the motion planning side, we note that different UAV applications can influence the choice of motion planning algorithms. The main motivation is that different types of planners can exploit specific properties and guarantees of the cost functions. For example, sampling-based planners \cite{kuffner2000rrt,karaman2011sampling,elbanhawi2014sampling} or search-based planners \cite{Lav06,aine2016multi} should ideally use fast-to-compute costs so that many different states can be explored during search in high-dimensional state spaces. Other categories of planners, based on trajectory optimization \cite{ratliff2009chomp,schulman2013finding}, usually require cost functions to be differentiable to the first or higher orders. We also find hybrid methods that make judicious use of optimization combined with search or sampling \cite{choudhury2016regionally,luna2013anytime}.

Furthermore, different systems present significant differences in onboard versus off-board computation.
We summarize and compare contributions from past works in Table~\ref{tab:related_work}. It is important to notice that none of the previously published approaches provides a complete solution to the generic aerial cinematography problem using only onboard resources.


\begin{table}[h!]
\begin{center}
\caption{Comparison of dynamic aerial cinematography systems. Notes: i) \cite{huang2018act} define artistic selection as the viewpoint that maximizes the projection of the actor on the image; ii) \cite{bonatti2018autonomous} localize the actor visually only for control of the camera gimbal, but use GPS to obtain the actor's position in global coordinates for planning; iii) Cells marked with ``Actor'' for occlusion and obstacle avoidance mean that those approaches only take into account the actors in the scene as ellipsoidal obstacles, and disregard other objects.}
\begin{tabular}{l|lllllll}
\label{tab:related_work}
 &&&&&\\ 
 \textbf{\pbox{0cm}{Reference}}    & \textbf{\pbox{1.8cm}{Online\\art. selec.\\}} & \textbf{\pbox{1.1cm}{Online\\map\\}}  & \textbf{\pbox{0.6cm}{Actor\\localiz.\\}} & \textbf{\pbox{1.3cm}{Onboard\\comp.\\}}    & \textbf{\pbox{0cm}{Avoids \\occl.\\}}   & \textbf{\pbox{0cm}{Avoids\\obst.\\}} & \textbf{\pbox{0.9cm}{Online\\plan\\}} \\
 \hline
 \cite{galvane2017automated} &  \cellcolor{red!25} $\times$ &  \cellcolor{red!25} $\times$          & \cellcolor{red!25} $\times$              & \cellcolor{red!25} $\times$               & \cellcolor{red!25} $\times$              & \cellcolor{red!25} $\times$          & \cellcolor{green!25} \checkmark       \\
 \cite{joubert2016towards}   &  \cellcolor{red!25} $\times$ &  \cellcolor{red!25} $\times$          & \cellcolor{red!25} $\times$              & \cellcolor{red!25} $\times$               & \cellcolor{red!25} $\times$              & \cellcolor{yellow!25}  Actor     & \cellcolor{green!25} \checkmark       \\  
 \cite{nageli2017real}       &  \cellcolor{red!25} $\times$ &  \cellcolor{red!25} $\times$          & \cellcolor{red!25} $\times$              & \cellcolor{red!25} $\times$               &\cellcolor{yellow!25} Actor         & \cellcolor{yellow!25} Actor     & \cellcolor{green!25} \checkmark       \\
 \cite{galvane2018directing} &  \cellcolor{red!25} $\times$ &  \cellcolor{red!25} $\times$          & \cellcolor{red!25} $\times$              & \cellcolor{red!25} $\times$               &\cellcolor{green!25} \checkmark           & \cellcolor{green!25} \checkmark       & \cellcolor{green!25} \checkmark       \\
 
 \cite{huang2018through}     &  \cellcolor{red!25} $\times$ &  \cellcolor{red!25} $\times$          & \cellcolor{green!25} \checkmark          & \cellcolor{green!25} \checkmark           & \cellcolor{red!25} $\times$              & \cellcolor{yellow!25} Actor     & \cellcolor{green!25} \checkmark       \\
 \cite{huang2018act}         &  \cellcolor{yellow!25} Actor proj. &  \cellcolor{red!25} $\times$          & \cellcolor{green!25} $\checkmark$        & \cellcolor{green!25} \checkmark           &\cellcolor{red!25} $\times$               & \cellcolor{yellow!25} Actor     & \cellcolor{green!25} \checkmark       \\
 \cite{bonatti2018autonomous} &  \cellcolor{red!25} $\times$ &  \cellcolor{red!25} $\times$          & \cellcolor{yellow!25} Vision              & \cellcolor{green!25} \checkmark           &\cellcolor{green!25} \checkmark           & \cellcolor{green!25} \checkmark       & \cellcolor{green!25} \checkmark       \\
 \cite{bonatti2019towards}     &  \cellcolor{red!25} $\times$          & \cellcolor{green!25} \checkmark       & \cellcolor{green!25} \checkmark          & \cellcolor{green!25} \checkmark           &  \cellcolor{green!25} \checkmark         & \cellcolor{green!25} \checkmark       & \cellcolor{green!25} \checkmark       \\
 \textbf{\quad\quad\quad\quad\quad\quad Ours} &  \cellcolor{green!25} \checkmark    & \cellcolor{green!25} \checkmark       & \cellcolor{green!25} \checkmark          & \cellcolor{green!25} \checkmark           &  \cellcolor{green!25} \checkmark         & \cellcolor{green!25} \checkmark       & \cellcolor{green!25} \checkmark       \\

\end{tabular}
\end{center}
\end{table}

\paragraph{Making artistic choices autonomously:}
A common theme behind all the work presented so far is that a user must always specify which kind of output they expect from the system in terms of artistic behavior. This behavior is generally expressed in terms of the set of parameters $\Omega_{art}$, and relates to different shot types, camera angles and angular speeds, type of actor framing, etc. If one wishes to autonomously specify artistic choices, two main points are needed: a proper definition of a metric for artistic quality of a scene, and a decision-making agent which takes actions that maximize this quality metric, as explained in Equation~\ref{eq:art_mapping}.

Several works explore the idea of learning a beauty or artistic quality metric directly from data. \cite{karpathy2015deep} learns a measure for the quality of \textit{selfies}; \cite{fang2017creatism} learn how to generate professional landscape photographs; \cite{gatys2016image} learn how to transfer image styles from paintings to photographs. 

On the action generation side, we find works that have exploited deep reinforcement learning \cite{mnih2015human} to train models that follow human-specified behaviors. Closest to our work, \cite{christiano2017deep} learns behaviors for which hand-crafted rewards are hard to specify, but which humans find easy to evaluate. 

Our work, as described in Section~\ref{sec:artistic}, brings together ideas from all the aforementioned areas to create a generative model for shot type selection in aerial filming drones which maximizes an artistic quality metric.

\paragraph{Online environment mapping:}
Dealing with imperfect representations of the world becomes a bottleneck for viewpoint optimization in physical environments. As the world is sensed online, it is usually incrementally mapped using voxel occupancy maps~\cite{thrun2005probabilistic}. To evaluate a viewpoint, methods typically raycast on such maps, which can be very expensive~\cite{isler2016information,Charrow-RSS-15}. Recent advances in mapping have led to better representations that can incrementally compute the \emph{truncated signed distance field (TSDF)}~\cite{newcombe2011kinectfusion,klingensmith2015chisel}, i.e. return the distance and gradient to nearest object surface for a query. TSDFs are a suitable abstraction layer for planning approaches and have already been used to efficiently compute collision-free trajectories for  UAVs~\cite{oleynikova2016voxblox,cover2013sparse}. 


\paragraph{Visual target state estimation:} 

Accurate object state estimation with monocular cameras is critical for many robot applications, including autonomous aerial filming. Two key problems in target state estimation include detecting objects and their orientation. 

Deep learning based techniques have achieved remarkable progress in the area of 2D object detection, such as YOLO (You Only Look Once)~\cite{redmon2016you}, SSD (Single Shot Detector)~\cite{liu2016ssd} and Faster R-CNN method \cite{ren2015faster}. These methods use convolutional neural networks (CNNs) for bounding box regression and category classification. They requires powerful GPUs, and cannot achieve real-time performance when deployed to the onboard platform. Another problem with off-the-shelf models trained on open datasets is that they do not generalize well to the areal filming scenario due to mismatches in data distribution due to angles, lighting, distances to actor and motion blur. Later in Section~\ref{sec:vision} we present our approach for obtaining a real-time object detector for our application.

Another key problem in the actor state estimation for aerial cinematography is estimating the heading direction of objects in the scene. Heading direction estimation (HDE) has been widely studied especially in the context of humans and cars as target objects. There have been approaches that attach sensors including inertial sensors and GPS to the target object to obtain the object's \cite{liu2016novel,deng2017heading}\cite{vista2015design} heading direction. While these sensors provide reliable and accurate estimation, it is highly undesirable for the target actor to carry these extra sensors. Thus, we primarily focus on vision-based approaches for our work that don’t require the actor to carry any additional equipment.

In the context of Heading Direction Estimation using visual input, there have been approaches based on classical machine learning techniques. Based on a probabilistic framework, \cite{flohr2015probabilistic} present a joint pedestrian head and body orientation estimation method, in which they design a HOG based linear SVM pedestrian model. Learning features directly from data rather than handcrafting them has proven more successful, especially in the domain of computer vision. We therefore leverage learning based approaches that ensure superior generalizability and improved robustness. 

Deep learning based approaches have been successfully applied  to the area of 2D pose estimation \cite{toshev2014deeppose, cao2017realtime} which is a related problem. However, the 3D heading direction cannot be trivially recovered from 2D points because the keypoint's depth remains undefined and ambiguous. Also, these approaches are primarily focused on humans and don't address other objects including cars.

There are fewer large scale datasets for 3D pose estimation \cite{h36m_pami, raman2016direction, liu2013accurate, Geiger2013IJRR} and the existing ones generalize poorly to our aerial filming task, again due to mismatch in the data distribution. Thus, we look for approaches that can be applied in limited labeled data setting. The limited dataset constraint is common in many robotics applications, where the cost of acquiring and labeling data is high. Semi-supervised learning (SSL) is an active research area in this domain. However, most of the existing SSL works are primarily focused on classification problems \cite{weston2012deep,hoffer2016semi,rasmus2015semi,dai2017good}, which assume that different classes are separated by a low-density area and easy to separate in high dimensional space. This assumption is not directly applicable to regression problems. 

In the context of cinematography, temporal continuity can be leveraged to formulate a semi-supervised regression problem. \cite{mobahi2009deep} developed one of the first approaches to exploit temporal continuity in the context of deep convolutional neural networks. The authors use video temporal continuity over the unlabeled data as a pseudo-supervisory signal and demonstrate that this additional signal can improve object recognition in videos from the COIL-100 dataset \cite{nene1996columbia}. There are other works that learn feature representations by exploiting temporal continuity \cite{zou2012deep, goroshin2015unsupervised, stavens2010unsupervised, srivastava2015unsupervised, wang2015unsupervised}. \cite{zou2012deep} included the video temporal constraints in an autoencoder framework and learn invariant features across frames. \cite{wang2015unsupervised} designed a Siamese-triplet network which can be trained in an unsupervised manner with a large amount of video data, and  showed that the unsupervised visual representation can achieve competitive performance on various tasks, compared to its ImageNet-supervised counterpart. 
Inspired by these approaches, our recent work \cite{wang2019heading} aims to improve the learning of a regression model from a small labeled dataset by leveraging unlabeled video sequences to enforce temporally smooth output predictions. 

After the target's location and heading direction is estimated on the image plane, we can project it onto the world coordinates and use different methods to estimate the actor's future motion. Motion forecast methods can range from filtering methods such as Kalman filters and extended Kalman filters \cite{thrun2005probabilistic}, which are based solely on the actor's dynamics, to more complex methods that take into account environmental features as well. As an example of the latter, \cite{urmson2008autonomous} use traditional motion planner with handcrafted cost functions for navigation among obstacles, and \cite{zhang2018integrating} use deep inverse reinforcement learning to predict the future trajectory distribution vehicles among obstacles.

\section{System Overview} 
\label{sec:system_arch}

In this section we detail the design hypotheses (Subsec.~\ref{subsec:hyp}) that influenced the system architecture layout (Subsec.~\ref{subsec:arch}), as well as our hardware (Subsec.~\ref{subsec:hardware}) and simulation (Subsec.~\ref{subsec:sim_platform}) platforms.

\subsection{Design Hypotheses}
\label{subsec:hyp}


Given the application challenges (Subsec.~\ref{subsec:challenges}) and problem definition (Sec.~\ref{sec:problem_form}), we defined three key hypotheses to guide the layout of the system architecture for the autonomous aerial cinematography task. These hypotheses serve as high-level principles for our choice of sub-systems, sensors and hardware. We evaluate the hypotheses later in Section~\ref{sec:experiments}, where we detail our simulation and field experiments.

\begin{enumerate}[label=Hyp. \arabic*]
    \item \textit{Onboard sensors can provide sufficient information for good cinematography performance.}\\
    This is a fundamental assumption and a necessary condition for development of real-world aerial cinematography systems that do not rely on ground-truth data from off-board sensors. We hypothesize our system can deal with noisy measurements and extract necessary actor and obstacle information for visual actor localization, mapping and planning.

    \item \textit{Decoupling gimbal control from motion planning improves real-time performance and robustness to noisy actor measurements.}\\
    We assume that an independent 3-DOF camera pose controller can compensate for noisy actor measurements. We expect advantages in two sub-systems: (i) the motion planner can operate faster and with a longer time horizon due to the reduced trajectory state space, and (ii) visual tracking will be more precise because the controller uses direct image feedback instead of a noisy estimate of the actor's location. We use a gimbaled camera with 3-DOF control, which is a reasonable requirement given today's UAV and camera technology.

    \item \textit{Analogous to the role of a movie director, the artistic intent sub-system should provide high-level guidelines for camera positioning, but not interfere directly on low-level controls.}\\
    We hypothesize that a hierarchical structure to guide artistic filming behavior employing high-level commands is preferable to an end-to-end low-level visio-motor policy because: (i) it's easier to ensure overall system safety and stability by relying on more established motion planning techniques, and (ii) it's more data-efficient and easier to train a high-level decision-making agent than an end-to-end low-level policy. 

\end{enumerate}

\subsection{System Architecture}
\label{subsec:arch}

\begin{figure}[t]
    \centering
    \includegraphics[width=0.8\textwidth]{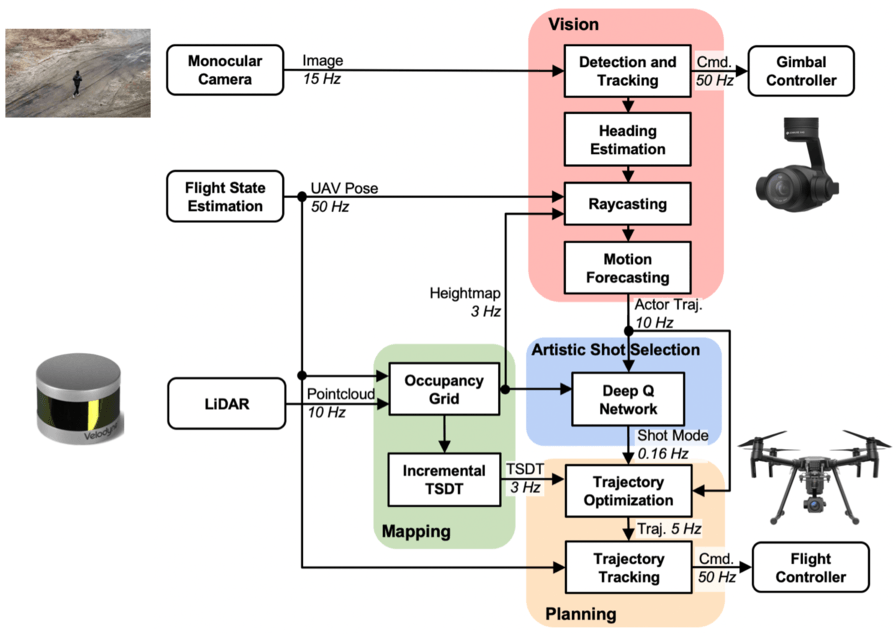}
    \caption{The system consists of 4 main modules running in parallel: Vision, Mapping, Planning and Artistic Shot Selection. The system takes in visual, LiDAR and GPS inputs to output gimbal and flight controller commands. 
    }
    \label{fig:block}
\end{figure}

Taking into account the design hypotheses, we outline the software architecture in Figure \ref{fig:block}. The system consists of 4 main modules: Vision, Mapping, Planning and Artistic Shot Selection. The four modules run in parallel, taking in camera, LiDAR and GPS inputs to output gimbal and flight controller commands for the UAV platform.

\paragraph{Vision (Section \ref{sec:vision}): } The module takes in monocular images to compute a predicted actor trajectory for the Shot Selection and Planning Module. Following \textit{Hyp. 2}, the vision module also controls the camera gimbal independently of the planning module.

\paragraph{Mapping (Section \ref{sec:mapping}): } The module registers the accumulated LIDAR point cloud and outputs different environment  representations: obstacle height map for raycasting and shot selection, and truncated signed distance transform (TSDT) map for the motion planner.

\paragraph{Artistic Shot Selection (Section \ref{sec:artistic}): } Following \textit{Hyp. 3} the module acts as an \textit{artistic movie director} and defines high-level inputs for the motion planner defining the most aesthetic shot type (left, right, front, back) for a given scene context, composed of actor trajectory and obstacle locations. 

\paragraph{Planning (Section \ref{sec:planning}): } The planning module takes in the predicted actor trajectory, TSDT map, and the desired artistic shot mode to compute a trajectory that balances safety, smoothness, shot quality and occlusion avoidance. Using the UAV pose estimate, the module outputs velocity commands for the UAV to track the computed trajectory.




\subsection{Hardware}
\label{subsec:hardware}

\begin{figure}[h!]
    \centering
    \begin{subfigure}[b]{0.35\textwidth}
        \includegraphics[width=\textwidth]{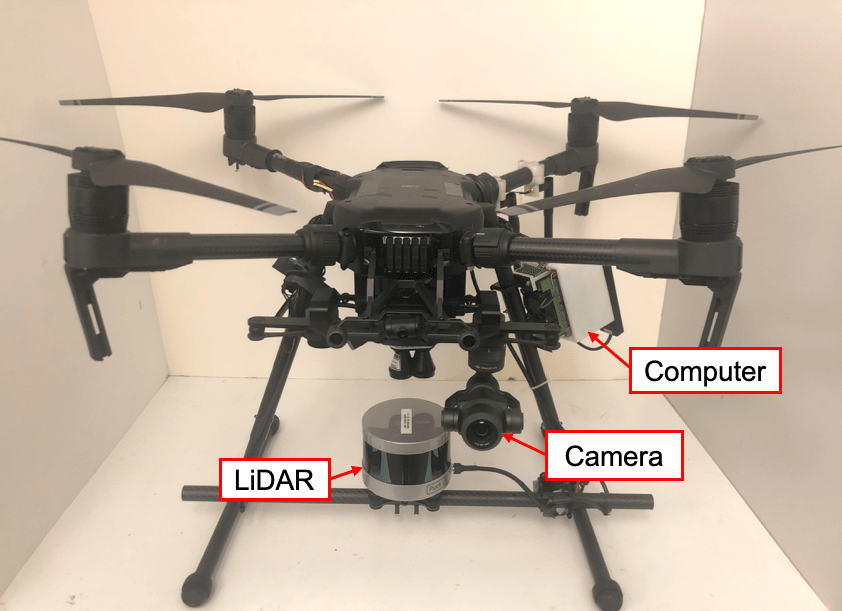}
    \end{subfigure}
    \caption{System hardware: DJI M210 drone equipped with Nvidia TX2 computer, Velodyne VLP-16 Puck Lite LiDAR and Zenmuse X4S camera gimbal.}
    \label{fig:hardware}
\end{figure}

Our base platform is the DJI M210 quadcopter, shown in Figure \ref{fig:hardware}. The UAV fuses GPS, IMU and compass for state estimation, which can be accessed via DJI's SDK. The M210 has a maximum payload capacity of $2.30$ kg \footnote{\label{note1} https://www.dji.com/products/compare-m200-series}, which limits our choice of batteries and onboard computers and sensors.

Our payload is composed of (weights are summarized in Table~\ref{tab:weights}): 
\begin{itemize}
  \item DJI TB50 Batteries, with maximum flight time of $13$ minutes at full payload;
  \item DJI Zenmuse X4S gimbaled camera, whose 3-axis gimbal can be controlled independently of the UAV's motion with angular precision of $\pm 0.01^{\circ}$, and counts with a vibration-dampening structure. The camera records high-resolution videos, up to 4K at 60 FPS;
  \item NVIDIA Jetson TX2 with Astro carrier board. 8 GB of RAM, 6 CPU cores and 256 GPU cores for onboard computation;
  \item Velodyne Puck VLP-16 Lite LiDAR, with $\pm15^{\circ}$ vertical field of view and $100$ m max range.
\end{itemize}
 

\begin{table}[h!]
    \caption{System and payload weights.}
    \centering
    \begin{tabular}{l|l}
    \label{tab:weights}
    Component & Weight (kg) \\ \hline
    DJI M210 & 2.80 \\
    DJI Zenmuse X4S &  0.25\\
    DJI TB 50 Batteries $\times$ 2 & 1.04\\
    NVIDIA TX2 w/ carrier board & 0.30 \\
    VLP-16 Lite& 0.59\\
    Structure Modifications & 0.63\\
    Cables and connectors & 0.28\\

    \hline
    \textbf{Total:} & 5.89 $\leq$ 6.14 (maximum takeoff weight \footref{note1})
    \end{tabular}
\end{table}

\subsection{Photo-realistic Simulation Platform}
\label{subsec:sim_platform}
\begin{figure}[t!]
    \centering
    \includegraphics[width=1.0\textwidth]{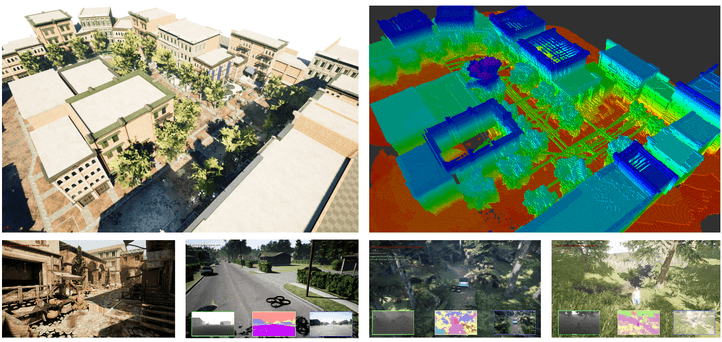}
    \caption{Simulation platform: Airsim combines a physics engine with photo-realistic renderings of various environments and actors (human and vehicles). Here we show the urban and forest environments we used for testing our framework. We build a point cloud and an occupancy map in the simulation. The simulation provides ground truth data for the actor's pose, which is used to evaluate the performance of the vision pipeline. }
    \label{fig:airsim}
\end{figure}

We use the Microsoft AirSim simulation platform \cite{airsim2017fsr} to test our framework and to collect training data for the shot selection module, as explained in detail in Section~\ref{sec:artistic}. Airsim offers a high-fidelity visual and physical simulation for quadrotors and actors (such as humans and cars), as shown in Figure~\ref{fig:airsim}. We built a custom ROS~\cite{quigley2009ros} interface for the simulator, so that our system can switch between the simulation and the real drone seamlessly. All nodes from the system architecture are written in C++ and Python languages, and communicate using the ROS framework.


\section{Visual Actor Localization and Heading Estimation}
\label{sec:vision}

\begin{figure}[t!]
    \center
    \includegraphics[width=1.0\textwidth]{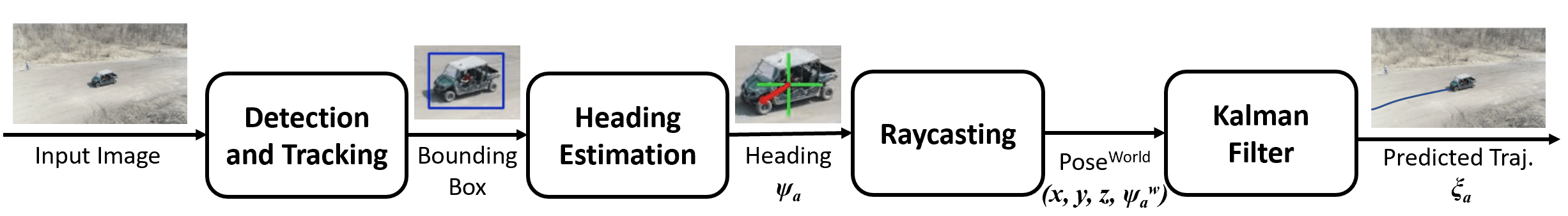}
    \caption{Vision sub-system. We detect and track the actor's bounding box, estimate its heading, and project its pose to world coordinates. A Kalman Filter predicts the actor's forecasted trajectory $\Path{a}$.}
    \label{fig:vision_pipeline}
\end{figure}

The vision module is responsible for two critical roles in the system: to estimate the actor's future trajectory and to control the camera gimbal to keep the actor within the frame. Figure \ref{fig:vision_pipeline} details the four main sub-modules: actor detection and tracking, heading direction angle estimation, global position ray-casting, and finally a filtering stage for trajectory forecasting. Next, we detail each sub-module.

\subsection{Detection and Tracking}

As we discussed in Section~\ref{sec:related_work}, the state-of-the-art object detection methods require large computational resources, which are not available on our onboard platform, and do not perform well in our scenario due to the data distribution mismatch. Therefore, we develop two solutions: first, we build a custom network structure and train it on both the open and context-specific datasets in order to improve speed and accuracy; second, we combine the object detector with a real-time tracker for stable performance.

The deep learning based object detectors are composed of a feature extractor followed by a classifier or regressor. Different feature extractors could be used in each detector to balance efficiency and accuracy. Since the onboard embedded GPU is less powerful, we can only afford feature extractor with relatively fewer layers. We compare several lightweight publicly available trained models for people detection and car detection. 

Due to good real-time inference speed and low memory usage, we combine the MobileNet \cite{howard2017mobilenets} for feature extraction and the Faster-RCNN \cite{ren2015faster} architecture. Our feature extractor consists of 11 depth-wise convolutional modules, which contains 22 convolutional layers. Following the Faster-RCNN structure, the extracted feature then goes to a two-stage detector, namely a region proposal stage and a bounding box regression and classification stage.  While the size of the original Faster-RCNN architecture with VGG is 548 MB, our custom network's is 39 MB, with average inference time of $300$ ms. 

\begin{figure}[h!]
    \center
    \includegraphics[width=0.68\textwidth]{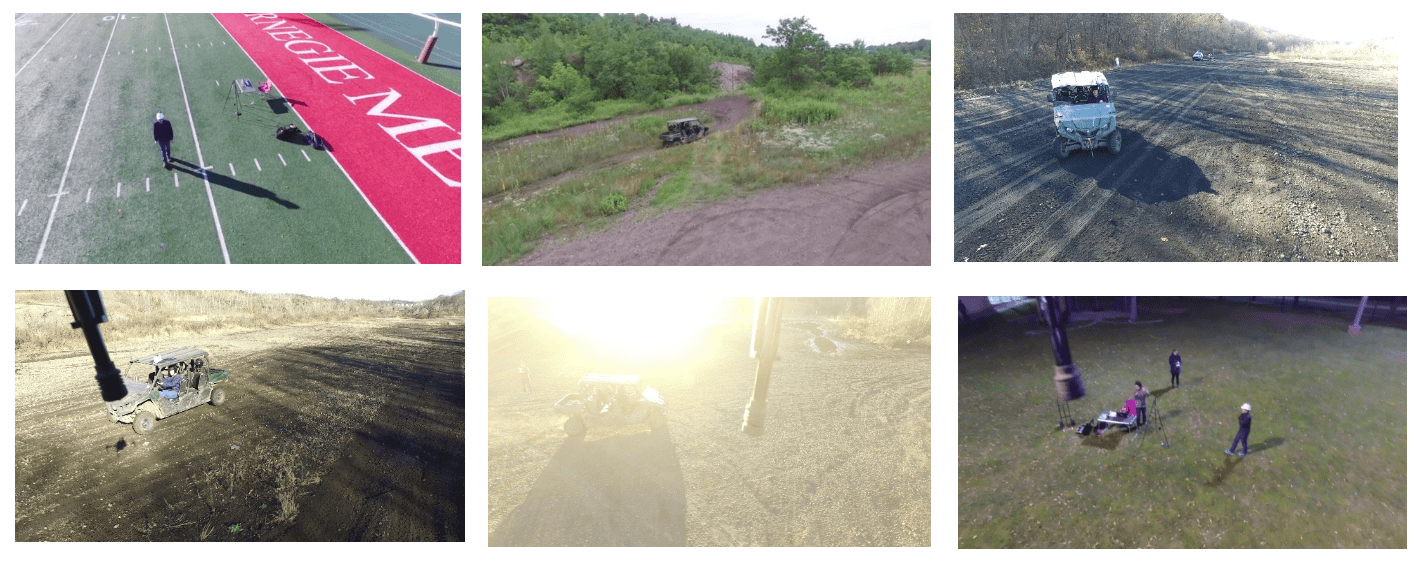}
    \caption{Examples of challenging images for actor detection in the aerial filming task. Large relative tilt angles to the ground, variable lighting, large distance to actor and heavy motion blur make bounding box detection harder than in images from open datasets.}
    \label{fig:challenge_data}
\end{figure}

The distribution of images in the aerial filming task differs significantly from the usual images found in open accessible datasets, due to highly variable relative yaw and tilt angles to the actors, large distances, varying lighting conditions and heavy motion blur. Figure~\ref{fig:challenge_data} displays examples of challenging situations faced in the aerial cinematography problem. Therefore, we trained our network with images from two sources: a custom dataset of 120 challenging images collected from field experiments, and images from the COCO \cite{lin2014microsoft} dataset, in a 1:10 ratio. We limited the detection categories only to person, car, bicycle, and motorcycle, which are object types that commonly appear as actors in aerial filming.

The detection module receives the main camera's monocular image as inputs, and outputs a bounding box. 
We use this initial bounding box to initialize a template tracking process, and re-initialize detection whenever the tracker's confidence falls below acceptable limits. We adopt this approach, as opposed to detecting the actor in every image frame, because detection is a computationally heavy process, and the high rate of template tracking provides more stable measurements for subsequent calculations. We use Kernelized Correlation Filters \cite{henriques2015high} to track the template over the next incoming frames. 

\begin{figure}[h]
    \center
    \includegraphics[width=0.25\textwidth]{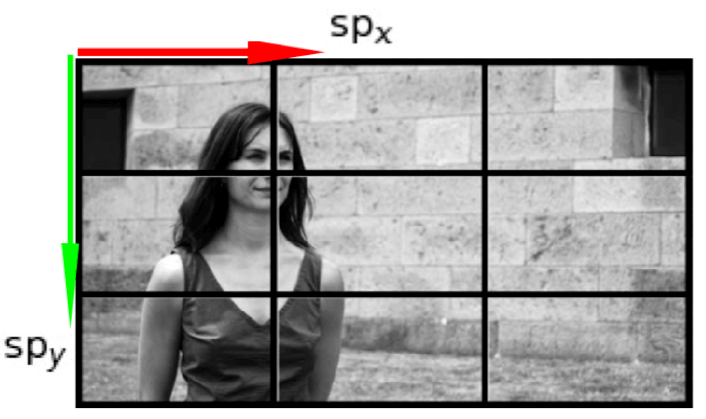}
    \caption{Desired screen position of the actor projection, defined by parameters: $\mathrm{sp_x},\mathrm{sp_y} \in [0,1]$. Typically the user uses one of the thirds of the screen to set the actor's position, or centers the actor on the frame~\cite{bowen2013grammar}.}
    \label{fig:rule_of_thirds}
\end{figure}

As mentioned in Section~\ref{sec:system_arch}, we actively control camera gimbal independently of the UAV's motion to maintain visibility of the target. We use a PD controller to frame the actor on the desired screen position, following the commanded artistic principles from the operator. Typically, the operator centers the target on the middle of the image space, or uses visual composition rules such as the rule of thirds~\cite{bowen2013grammar}, as seen on Figure~\ref{fig:rule_of_thirds}.

\subsection{Heading Estimation}
When filming a moving actor, heading direction estimation (HDE) plays a central role in motion planning. Using the actor's heading information, the UAV can position itself within the desired shot type determined by the user's artistic objectives, \textit{e.g:} back, front, left and right side shots, or within any other desired relative yaw angle. 


Estimating the heading of people and objects is also an active research problem in many other applications, such as pedestrian collision risk analysis \cite{tian2014estimation}, human-robot interaction \cite{vazquez2015parallel} and activity forecasting \cite{kitani2012activity}. Similar to challenges in bounding box detection, models obtained in other datasets do not easily generalize to the aerial filming task, due to a mismatch in the types of images from datasets to our application. In addition, when the trained model is deployed on the UAV, errors is compounded because the HDE relies on a imperfect object detection module, increasing the mismatch \cite{ristani2016MTMC,Geiger2013IJRR}.

No current dataset satisfies our needs for aerial HDE, creating the need for us to create a custom labeled dataset for our application. As most deep learning approaches, training a network is a data-intensive process, and manually labeling a large enough dataset for conventional supervised learning is a laborious and expensive task. The process is further complicated as multiple actor types such as people, cars and bicycles can appear in footages. 

These constraints motivated us to formulate a novel semi-supervised algorithm for the HDE problem \cite{wang2019heading}. To drastically reduce the quantity of labeled data, we leverage temporal continuity in video sequences as an unsupervised signal to regularize the model and achieve better generalization. We apply the semi-supervised algorithm in both training and testing phases, drastically increasing inference performance, and show that by leveraging unlabeled sequences, the amount of labeled data required can be significantly reduced.



\subsubsection{Defining the loss for temporal continuity} 

\begin{figure}[h!]
    \centering
    \includegraphics[width=0.45\textwidth]{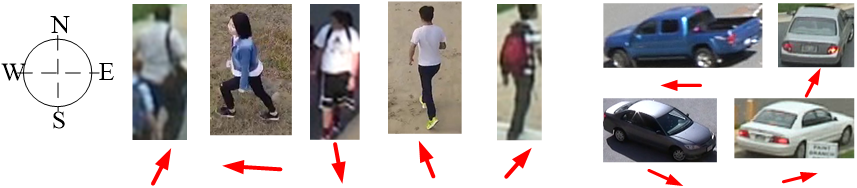}
    \caption{Example of actor bounding boxes and their respective heading angles $\psi_{a}$ in image space. Given the images, our objective is to predict the heading direction, shown as red arrows.}
    \label{fig:heading_problem}
\end{figure}

We define the pose of the actor as a vector $[x, y, z, \psi_a^w]$ on the ground surface. In order to estimate the actor's heading direction in the world frame  $\psi_a^w$, we first predict the actor's heading $\psi_{a}$ in the image frame, as shown in Figure~\ref{fig:heading_problem}. 
Once $\psi_{a}$ is estimated, we project this direction onto the world frame coordinates using the camera's intrinsic and extrinsic matrices.  

The HDE module outputs the estimated heading angle $\psi_{a}$ in image space. Since $\psi_{a}$ is ambiguously defined at the frontier between $-\pi$ and $\pi$, we define the inference as a regression problem that outputs two continuous values: $[\cos(\psi_{a}), \sin(\psi_{a})]$. This avoids model instabilities during training and inference. 

We assume access to a relatively small labeled dataset $D=\{(x_i, y_i)\}_{i=0}^{n}$, where $x_i$ denotes input image, and $y_i=[cos(\psi_i), sin(\psi_i)]$ denotes the angle label. In addition, we assume access to a large unlabeled sequential dataset $U=\{q_j\}_{j=0}^{m}$, where $q_j=\{x_0, x_1,...,x_t\}$ is a sequence of temporally-continuous image data. 

The HDE module's main objective is to approximate a function $y=f(x)$, that minimizes the regression loss on the labeled data $ \sum_{(x,y)\in{D}}L_l(x_l,y_l) = \sum_{(x,y)\in{D}}||y_i-f(x_i)||^2$. One intuitive way to leverage unlabeled data is to add a constraint that the output of the model should not have large discrepancies over a consecutive input sequence. Therefore, we train the model to jointly minimize the labeled loss $L_l$ and some continuity loss $L_u$. We minimize the combined loss:

\begin{equation}
    L_{tot}  = \min{\sum _{(x_l,y_l)\in{L}}L_l(x_l,y_l)+\lambda \sum_{q_u\in{U}}{L_u(q_u)}}
\end{equation}



We define the unsupervised loss using the idea that samples closer in time should have smaller differences in angles than samples further away in time. A similar continuity loss is also used by \cite{wang2015unsupervised} when training an unsupervised feature extractor:

\begin{equation}
  \begin{aligned}
    L_u(q_u) &= \sum_{x_1,x_2,x_3} \max[0,D(x_1,x_2;f)-D(x_1,x_3;f)],\\
    \text{where:} \quad & D(x_1,x_2;f) = ||f(x_1)-f(x_2)||_2,\\
    \text{and:} \quad & x_1,x_2,x_3\in{q_u}
  \end{aligned}
\end{equation}





\subsubsection{Network structure}

\begin{figure}[h!]
    \centering
    \includegraphics[width=0.7\textwidth]{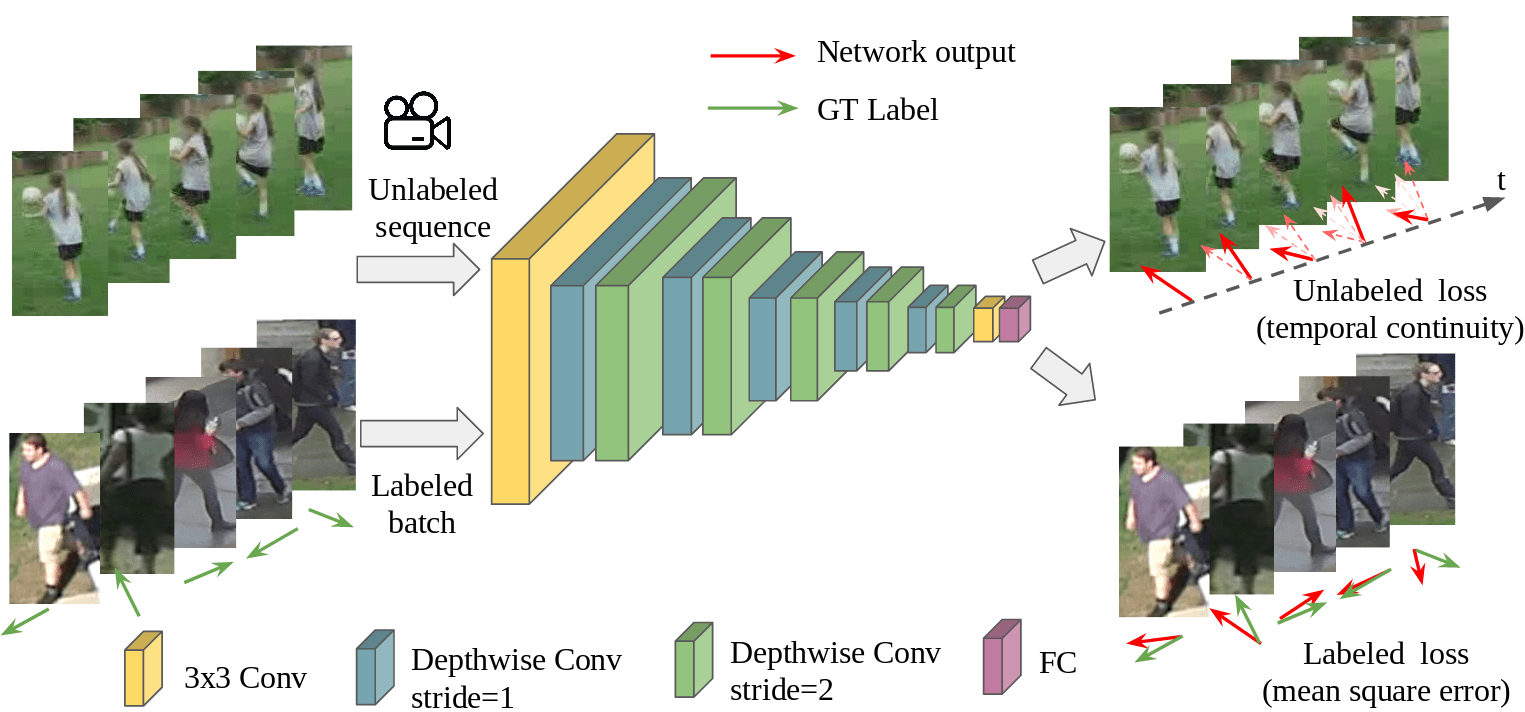}
    \caption{Our network architecture for predicting the actor's heading direction. We use a Mobilenet-based feature extractor followed by a convolutional layer and a fully connected layer to regress to angular values. The network is trained using both labeled and unsupervised losses.}
    \label{fig:heading_structure}
\end{figure}

For lower memory usage and faster inference time in the onboard computer, we design a compact CNN architecture based on MobileNet \cite{howard2017mobilenets}. The input to the network is a cropped image of the target's bounding box, outputted by the detection and tracking modules. The cropped image is padded to a square shape and resized to 192 x 192 pixels. After the 10 group-wise and point-wise convolutional blocks from the original MobileNet architecture, we add another convolutional layer and a fully connected layer that output two values representing the cosine and sine values of the angle. Figure~\ref{fig:heading_structure} illustrates the architecture. 


During each training iteration, one shuffled batch of labeled data and one sequence of unlabeled data are passed through the network. The labeled loss and unlabeled losses are computed and backpropagated through the network. 


\subsubsection{Cross-dataset semi-supervised fine-tuning}

Due to data distribution mismatch between the aerial cinematography task and open datasets, we train our network on a combination of images from both sources. Later in Subsection~\ref{subsec:experiments_heading} we evaluate the impact of fine-tuning the training process with unsupervised videos from our application.

\subsection{Ray-casting}

The ray-casting module convert the detection/tracking and HDE results from image space to coordinates and heading in the world frame $[x, y, z, \psi_a^w]$. 
Given the actor's bounding box, we project its center-bottom point onto a height map of the terrain, provided by the mapping module. The intersection of this line with the height map provides the $[x, y, z]$ location of the actor. 

Assuming that the camera gimbal's roll angle is fixed at zero degrees by the active gimbal controller, we can directly obtain the actor's heading direction on the world frame $\psi_a^w$ by transforming the heading $\psi$ from the image space with the camera's extrinsic matrix in world coordinates.

\begin{figure}[h]
    \centering
    \includegraphics[width=0.45\textwidth]{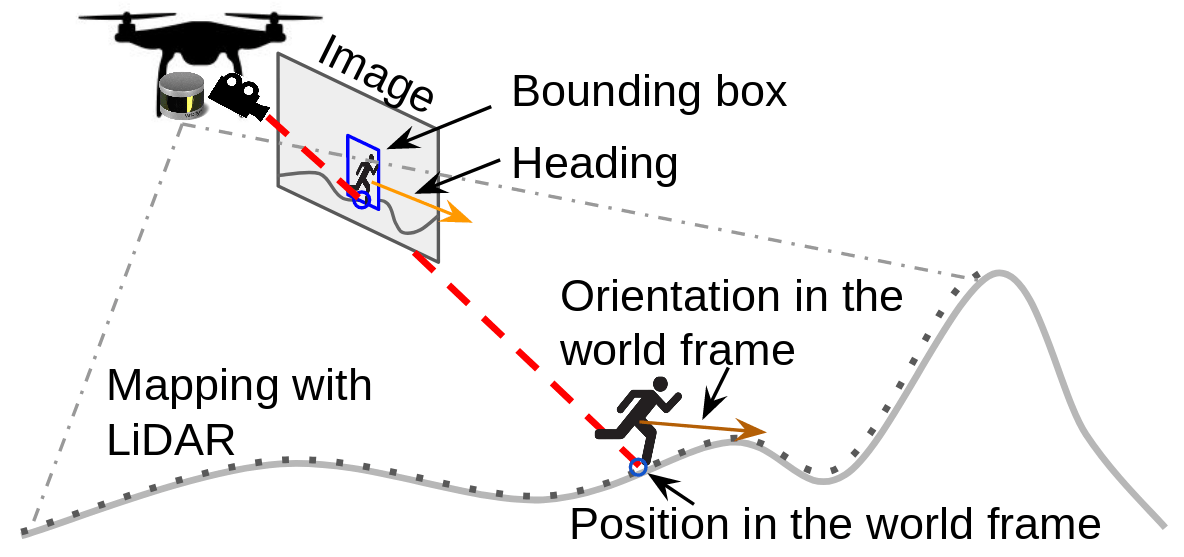}
    \caption{Raycasting module uses the actor's bounding box, estimated heading angle, environment height map and camera matrices to obtain pose of actor in the world frame $[x, y, z, \psi_a^w]$.}
    \label{fig:raycasting}
\end{figure}




\subsection{Motion Forecasting}
\label{subsec:forecast}

Given a sequence of actor poses in the world coordinates, we estimate the actor's future trajectory based on motion models. The motion planner later uses the forecast to plan non-myopically over long time horizons.

We use two different motion models depending on the actor types. For people, we apply a linear Kalman filter with a two-dimensional motion model. Since a person's movement direction can change drastically, we use no kinematic constraints applied to the motion model, and just assume constant velocity. We assume no control inputs for state $[x, y, \dot{x}, \dot{y}]$ in the prediction step, and use the next measurement of $[x, y, z]$ in the correction step. When forecasting the motion of cars and bicycles we apply an extended Kalman filter with a kinematic bicycle model. For both cases we use a $10$ s horizon for prediction.

\section{Online Environment Mapping}
\label{sec:mapping}

As explained in Section~\ref{sec:problem_form}, the motion planner requires signed distance values $\map$ to solve the optimization problem that results in the final UAV trajectory. The main role of the mapping subsystem described here is to register LiDAR points from the onboard sensor, update the occupancy grid $\grid$, and incrementally update the signed distance $\map$.

\subsection{LiDAR Registration} 
During our filming operation, we receive approximately $300,000$ points per second from the laser sensor mounted at the bottom of the aircraft. We register the points in the world coordinate system using a rigid body transform between the sensor and the aircraft plus the UAV's state estimation, which fuses GPS, barometer, internal IMUs and accelerometers. For each point we also store the corresponding sensor coordinate, which is used for the occupancy grid update. 

LiDAR points can be categorized either as hits, which represent successful laser returns within the maximum range of $100$ m, or as misses, which represent returns that are either non-existent, beyond the maximum range, or below a minimum sensor range. We filter all expected misses caused by reflections from the aircraft's own structure. Finally, we probabilistically update all voxels from $\grid$ between the sensor and its LiDAR returns, as described in Subsection~\ref{subsec:grid_update}. 

\subsection{Occupancy Grid Update}
\label{subsec:grid_update}
The mapping subsystem holds a rectangular grid that stores the likelihood that any cell in space is occupied. In this work we use a grid size of $250\times250\times100$ m, with $1$ m square voxels that store an $8$-bit integer value between $0-255$ as the occupancy probability, where $0$ is the limit for a fully free cell, and $255$ is the limit for a fully occupied cell. All cells are initialized as \textit{unknown}, with value of $127$. 

Algorithm~\ref{alg:ray} covers the grid update process.
The inputs to the algorithm are the sensor position $p_\mathrm{sensor}$, the LiDAR point $p_\mathrm{point}$, and a flag $\texttt{is\_hit}$ that indicates whether the point is a hit or miss. The endpoint voxel of a hit will be updated with log-odds value $l_\mathrm{occ}$, and all cells in between sensor and endpoint will be updated by subtracting value $l_\mathrm{free}$. We assume that all misses are returned as points at the maximum sensor range, and in this case only the cells between endpoint and sensor are updated $l_\mathrm{free}$.

As seen in Algorithm~\ref{alg:ray}, all voxel state changes to \textit{occupied} or \textit{free} are stored in lists $V^\mathrm{change}_\mathrm{occ}$ and $V^\mathrm{change}_\mathrm{free}$. State changes are used for the signed distance update, as explained in Subsection~\ref{subsec:idt_update}. 

\small
\begin{algorithm}[ht]
\caption{\small Update $\grid$ ($p_\mathrm{sensor}$, $p_\mathrm{point}$, \texttt{is\_hit})}
\label{alg:ray}
\SetAlgoLined
 Initialize $V^\mathrm{change}_\mathrm{occ}$, $V^\mathrm{change}_\mathrm{free}$ \Comment{list of changed voxels}\ 
 Initialize $l_\mathrm{free}$, $l_\mathrm{occ}$ \Comment{log-odds probabilistic updates}\
 \For{\textrm{\textbf{each}} voxel $v$ between $p_\mathrm{sensor}$ and $p_\mathrm{point}$}{
  $v \gets v - l_\mathrm{free}$\;
  \If{$v$ was occupied or unknown and now is free}{
   Append($v$, $V^\mathrm{change}_\mathrm{free}$)\;
   \For{\textrm{\textbf{each}} unknown neighbor $v_\mathrm{unk}$ of $v$}{
    Append($v_\mathrm{unk}$, $V^\mathrm{change}_\mathrm{occ}$)
   }
  }
  \If{$v$ is the endpoint and \normalfont{\texttt{is\_hit}} is true}{
   $v \gets v + l_\mathrm{occ}$\;
   \If{$v$ was free or unknown and now is occupied}{
   Append($v$, $V^\mathrm{change}_\mathrm{occ}$)
   }
  }
 }
 \textbf{return} $V^\mathrm{change}_\mathrm{occ}$, $V^\mathrm{change}_\mathrm{free}$
\end{algorithm}
\normalsize

\subsection{Incremental Distance Transform Update}
\label{subsec:idt_update}

We use the list of voxel state changes as input to an algorithm, modified from \cite{cover2013sparse}, that calculates an incremental truncated signed distance transform (iTSDT), stored in $\map$.
The original algorithm described by \cite{cover2013sparse} initializes all voxels in $\map$ as free, and as voxel changes arrive in sets $V^{change}_{occ}$ and $V^\mathrm{change}_\mathrm{free}$, it incrementally updates the distance of each free voxel to the closest occupied voxel using an efficient wavefront expansion technique within some limit (therefore truncated). 

Our problem, however, requires a \textit{signed} version of the DT, where the \textit{inside} and \textit{outside} of obstacles must be identified and given opposite signs (details of this requirement are given in the description of the occlusion cost function detailed in Section~\ref{sec:planning}). 
The concept of regions inside and outside of obstacles cannot be captured by the original algorithm, which provides only a iTDT (with no sign).
Therefore, we introduced two important modifications:

\paragraph{Using the borders of obstacles.} The original algorithm uses only the occupied cells of $\grid$, which are incrementally pushed into $\map$ using set $V^\mathrm{change}_\mathrm{occ}$. We, instead, define the concept of \textit{obstacle border} cells, and push them incrementally as $V^\mathrm{change}_\mathrm{occ}$.

Let $v_\mathrm{border}$ be an obstacle border voxel, and $V_\mathrm{border}$ be the set of all border voxels in the environment. We define $v_\mathrm{border}$ as any voxel that is either a direct hit from the LiDAR (lines $13-15$ of Alg.~\ref{alg:ray}), or as any \textit{unknown} voxel that is a neighbor of a \textit{free} voxel (lines $5-9$ of Alg.~\ref{alg:ray}). In other words, the set $V_\mathrm{border}$ will represent all cells that separate the known free space from unknown space in the map, whether this unknown space is part of cells inside an obstacle or cells that are actually free but just have not yet been cleared by the LiDAR. 

By incrementally pushing $V^\mathrm{change}_\mathrm{occ}$ and $V^\mathrm{change}_\mathrm{free}$ into $\map$, its data structure will maintain the current set of border cells $V_\mathrm{border}$. By using the same algorithm described in \cite{cover2013sparse} but now with this distinct type of data input, we can obtain the distance of any voxel in $\map$ to the closest obstacle border. One more step is required to obtain the sign of this distance.

\begin{figure}[h]
    \center
    \includegraphics[width=0.5\textwidth]{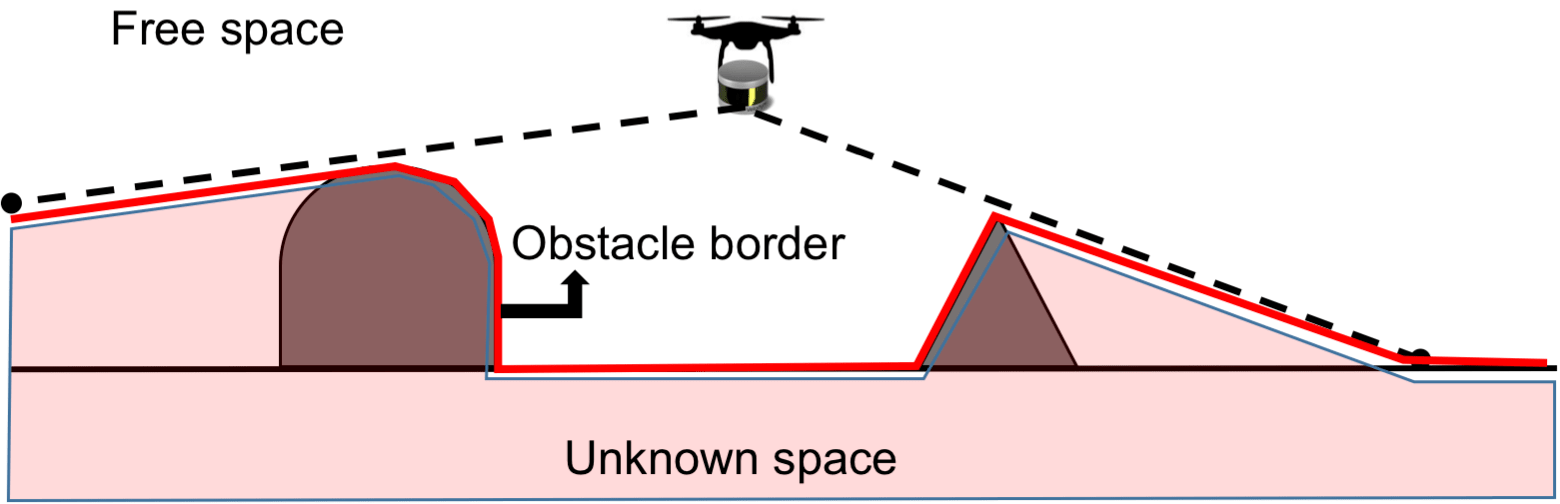}
    \caption{Diagram with our obstacle representation. Unknown voxels (red shade) are either inside of obstacles, or in zones occluded from the sensor's field of view. The red line displays the obstacle border, which is at the interface between LiDAR hits and free space, and at the interface between unknown and free space. In this conceptual figure we assume the sensor to be onmidirectional, so the ground and obstacles below aircraft were captured as hits; however in practice the sensor has field of view limitations. }
    \label{fig:mapping}
\end{figure}

\paragraph{Querying $\grid$ for the sign.} The data structure of $\map$ only stores the distance of each cell to the nearest obstacle border. Therefore we query the value of $\grid$ to attribute the sign of the iTSDT, marking free voxels as positive, and unknown or occupied voxels as negative (Figure~\ref{fig:mapping}).

\subsection{Building a Height Map}

Despite keeping a full 3D map structure as the representation used for planning (Section~\ref{sec:planning}), we also incrementally build a height map of the environment that is used for both the raycasting procedure when finding the actor position in world coordinates (Section~\ref{sec:vision}), and for the online artistic choice selection procedure (Section~\ref{sec:artistic}).

\begin{figure}[h]
    \center
    \includegraphics[width=0.4\textwidth]{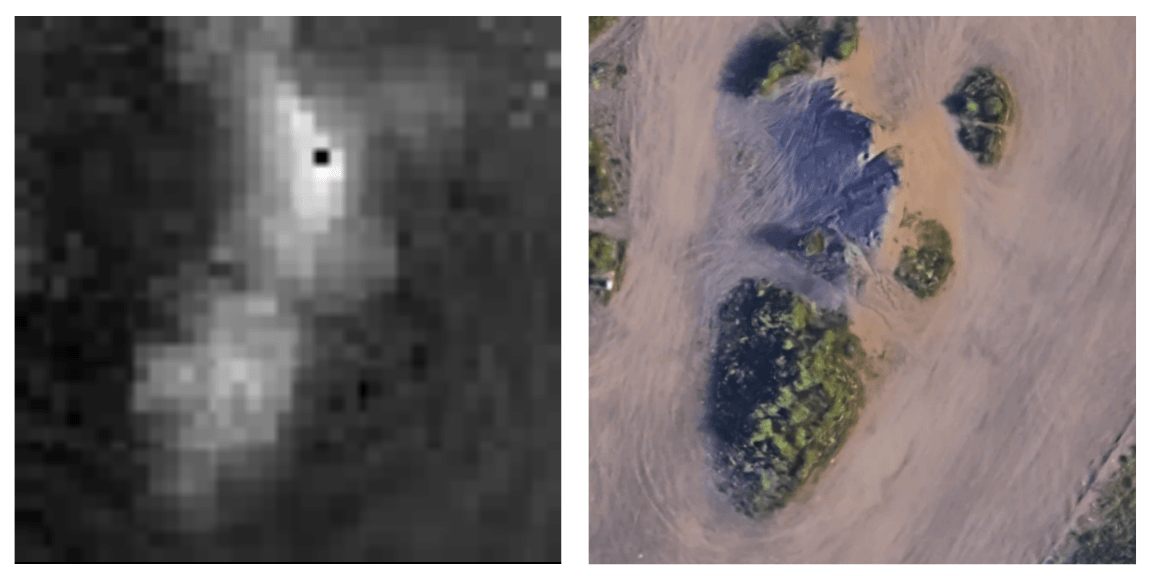}
    \caption{The left figure shows the height map accumulated over flight over a small mountain. Color scale goes from 0 ($-10$m) to 255 ($+10$m), where the zero reference of $127$ is taken at the UAV's initial takeoff height. The right figure is the top-down view of the mountain of the same place. 
    }
    \label{fig:height_map}
\end{figure}

The height map is a 2D array where the value of each cell corresponds to a moving average of the height of the LiDAR hits that arrive in each position. All cells are initialized with $0 m$ of height, relative to the world coordinate frame, which is taken from the UAV's takeoff position. An example height map is shown in Figure~\ref{fig:height_map}.


\section{Motion Planning}
\label{sec:planning}

The motion planner's objective is to calculate trajectories for the UAV to film the moving actor. Next we detail the definition of our trajectory, cost functions, the trajectory optimization algorithm, and implementation details.

\subsection{UAV Trajectory Definition}

Recall Section~\ref{sec:problem_form}, where we defined $\Path{q}(t) : [0,t_f] \rightarrow \real^3  \times SO(2)$ as the UAV trajectory and $\Path{a}(t) : [0,t_f] \rightarrow \real^3  \times SO(2)$ as the actor trajectory, where $\Path{q}(t) = \{x_{q}(t), y_{q}(t), z_{q}(t), \psi_{q}(t)\}$ and $\Path{a}(t) = \{x_{a}(t), y_{a}(t), z_{a}(t), \psi_{a}(t)\}$. High-frequency measurements of the actor's current position generate the actor's motion forecast $\Path{a}(t)$ in the vision module (Subsection~\ref{subsec:forecast}), and it is the motion planner's objective to output the UAV trajectory $\Path{q}(t)$. 

Since the gimbal controller can position the camera independently of the UAV's body motion, we purposefully decouple the UAV body's heading $\psi(t)$ from the main motion planning algorithm. We set $\psi_q(t)$ to always point from $\Path{q}(t)$ towards $\Path{a}(t)$ at all times, as seen in Equation~\ref{eq:heading}.

\begin{equation}
  \label{eq:heading}
  \psi_q(t) = atan2(y_{a}(t)-y_{q}(t), x_{a}(t)-x_{q}(t))
\end{equation}

This assumption significantly reduces the complexity of the planning problem by removing four degrees of freedom (three from the camera and one from the UAV's heading), and improves filming performance because the camera can be controlled directly from image feedback, without the accumulation of errors from the raycasting module (Section~\ref{sec:vision}).

Now, let $\Path{q}(t)$ represent the UAV's trajectory in a continuous time-parametrized form, and let $\Path{q}$ represent the same trajectory in a finite discrete form, with total time length $t_f$. Let point $p_0$ represents the contour conditions of the beginning of the trajectory. $\Path{q}$ contains a total of $n-1$ waypoints of the form $p_i$, where $i=1,...,n-1$, as shown in Equation~\ref{eq:traj}.

\begin{equation}
  \label{eq:traj}
  \begin{aligned}
    \Path{q} = 
    \begin{bmatrix}
      p_1\\p_2\\\vdots\\p_{n-1}
    \end{bmatrix} = 
    \begin{bmatrix}
      p_{1x} & p_{1y} & p_{1z}\\p_{2x} & p_{2y} & p_{2z}\\\vdots&\vdots&\vdots\\p_{n-1\ x} & p_{n-1\ y} & p_{n-1\ z}\\
    \end{bmatrix}
  \end{aligned}
\end{equation}

\subsection{Planner Requirements}

As explained in Section~\ref{sec:problem_form}, in a generic aerial filming framework we want trajectories which are smooth ($\costSmooth{}$), capture high quality viewpoints ($\costShot{}$), avoid occlusions ($\costOcc{}$) and keep the UAV safe ($\costObs{}$). Each objective can then be encoded in a separate cost function, and the motion planner's objective is to find the trajectory that minimizes the overall cost, assumed to be a linear combination of individual cost functions, subject to the initial condition constraints. For the sake of completeness, we repeat Equation~\ref{eq:main_cost} below as Equation~\ref{eq:main_cost_again}:

\begin{equation}
\begin{aligned}
\label{eq:main_cost_again}
\costFn{\Path{q}} &=  \begin{bmatrix}
       1 & \lambda_1 & \lambda_2 & \lambda_3
     \end{bmatrix} 
\begin{bmatrix}
       \costFnSmooth{\Path{q}} \\
       \costFnObs{\Path{q},\map} \\
       \costFnOcc{\Path{q},\Path{a},\map} \\
       \costFnShot{\Path{q},\Path{a},\Omega_{art}}
     \end{bmatrix} \\
\Path{q}^* &= \argminprob{\Path{q}} \quad \costFn{\Path{q}}, \quad \text{s.t.} \; \Path{q}(0) = \{x_0,y_0,z_0,\psi_0\}
\end{aligned}
\end{equation}


Our choice of cost functions and planning is dictated by two main observations. First, filming requires the UAV to reason over a longer horizon than reactive approaches, usually in the order of $\sim10$s. The UAV not only has to avoid local obstacles such as small branches or light posts, but also consider how larger obstacles such as entire trees, buildings, and terrain elevations may affect image generation. Note that the horizons are limited by how accurate the actor prediction is. Second, filming requires a high planning frequency. The actor is dynamic, constantly changing direction and velocity. The map is continuously evolving based on sensor readings. Finally, since jerkiness in trajectories have significant impact on video quality, the plans need to be smooth, free of large time discretization.

Based on these observations, we chose local trajectory optimization techniques to serve as the motion planner. Optimizations are fast and reason over a smooth continuous space of trajectories. In addition, locally optimal solutions are almost always of acceptable quality, and plans can be incrementally updated across planning cycles.

A popular optimization-based approach that addresses the aerial filming requisites is to cast the problem as an unconstrained cost optimization, and apply covariant gradient descent~\cite{zucker2013chomp,ratliff2009learning}. This is a quasi-Newton method, and requires that some of the objectives have analytic Hessians that are easy to invert and that are well-conditioned. With the use of first and second order information about the problem, such methods exhibit fast convergence while being stable and computationally inexpensive. The use of such quasi-Newton methods requires a set of differentiable cost functions for each objective, which we detail next.



\subsection{Definition of Cost Functions}




\subsubsection{Smoothness}

We measure smoothness as the cumulative sum of \textit{n-th order} derivatives of the trajectory, following the rationale of \cite{ratliff2009chomp}. Let $D$ be a discrete difference operator. The smoothness cost is:

\begin{equation}
 \label{eq:smooth_cont}
		\costFnSmooth{\Path{q}(t)} = \frac{1}{t_{f}} \frac{1}{2} \int_0^{t_{f}} \sum_{d=1}^{d_\mathrm{max}} \alpha_n (D^d \Path{q}(t))^2 dt,
\end{equation}

where $\alpha_n$ is a weight for different orders, and $d_\mathrm{max}$ is the number of orders. In practice, we penalize derivatives up to the third order, setting $\alpha_n = 1, d_\mathrm{max} = 3$.

Appendix~\ref{subsec:appendix_smooth} expands upon this cost function and reformulates it in matrix form using auxiliary matrices $A_\mathrm{smooth}$, $b_\mathrm{smooth}$, and $c_\mathrm{smooth}$. We state the cost, gradient and Hessian for completeness:

\begin{equation}
\begin{aligned}
 \costFnSmooth{\Path{q}} &= 
 \frac{1}{2(n-1)} Tr(\Path{q}^T A_\mathrm{smooth} \Path{q} + 2\Path{q}^T b_\mathrm{smooth} + c_\mathrm{smooth}) \\
   \nabla \costFnSmooth{\Path{q}} &= \frac{1}{(n-1)}(A_\mathrm{smooth}\Path{q}+b_\mathrm{smooth})\\
  \nabla^2 \costFnSmooth{\Path{q}} &= \frac{1}{(n-1)} A_\mathrm{smooth}
\end{aligned}
\end{equation}

\subsubsection{Shot quality}

\begin{figure}[h]
    \centering
    \includegraphics[width=0.7\textwidth]{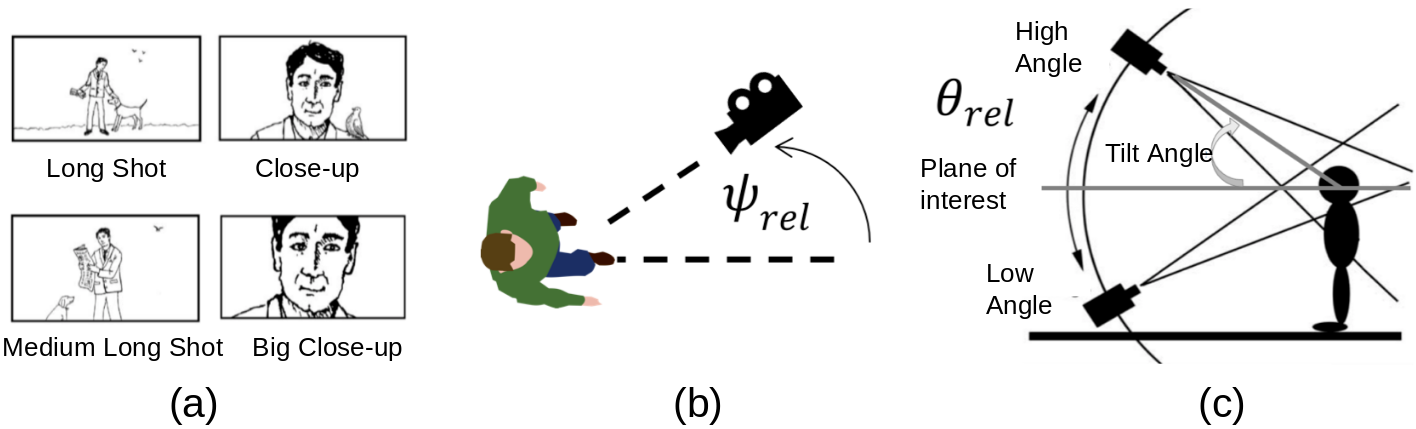}
    \caption{
    Shot parameters $\Omega_\mathrm{art}$ for shot quality cost function, adapted from \cite{bowen2013grammar}: a) shot scale $\rho$ corresponds to the size of the projection of the actor on the screen; b) line of action angle $\psi_\mathrm{rel} \in [0,2\pi]$; c) tilt angle $\theta_\mathrm{rel} \in [-\pi,\pi].$ 
    }
    \label{fig:cine_params}
\end{figure}

First, we analytically define the the artistic shot parameters. Based on cinematography literature \cite{arijon1976grammar,bowen2013grammar}, we select a minimal set of parameters that compose most of the shots possible for single-actor, single-camera scenarios. We define $\Omega_\mathrm{art}$ as a set of three parameters: $\Omega_\mathrm{art} = \{\rho, \psi_\mathrm{rel}, \phi_\mathrm{rel} \}$, where: (i) $\rho$ is the shot scale, which can be mapped to the distance between actor and camera, (ii) $\psi_\mathrm{rel}$ is the relative yaw angle between actor and camera, and (iii) $\phi_\mathrm{rel}$ is the relative tilt angle between the actor's current height plane and the camera. Figure~\ref{fig:cine_params} depicts the components of $\Omega_\mathrm{art}$.



Given a set $\Omega_\mathrm{art}$, we can now define a desired cinematography path $\Path{shot}(t)$:

\begin{equation}
	\begin{aligned}
		&\Path{shot}(t) = \Path{a}(t) + \rho 
		\begin{bmatrix}
		 \cos(\psi_{a} + \psi_{rel}) \sin(\theta_{rel})\\
		 \sin(\psi_{a} + \psi_{rel}) \cos(\theta_{rel})\\
		 \cos(\theta_{rel})
		\end{bmatrix}\\
	\end{aligned}
\end{equation}

Next, we can define an analytical expression for the shot quality cost function as the distance between the current camera trajectory and the the desired cinematography path:

\begin{equation}
  \costFnShot{\Path{q},\Path{a}} = \frac{1}{t_{f}} \frac{1}{2} \int_0^{t_{f}} ||\Path{q}(t)-\Path{shot}(\Path{a}(t))||^2 dt
\end{equation}

Appendix~\ref{subsec:appendix_shotqual} expands upon this cost function and reformulates it in matrix form using auxiliary matrices $A_\mathrm{shot}$, $b_\mathrm{shot}$, and $c_\mathrm{shot}$. Again, we state the cost, gradient and Hessian for completeness:

\begin{equation}
\begin{aligned}
  \costFnShot{\Path{q},\Path{a}} &= 
  \frac{1}{2(n-1)} Tr(\Path{q}^T A_\mathrm{shot} \Path{q} + 2\Path{q}^T b_\mathrm{shot} + c_\mathrm{shot}) \\
  \nabla \costFnShot{\Path{q}} &= \frac{1}{(n-1)}(A_\mathrm{shot}\Path{q}+b_\mathrm{shot})\\
  \nabla^2 \costFnShot{\Path{q}} &= \frac{1}{(n-1)} A_\mathrm{shot}
\end{aligned}
\end{equation}

We note that although the artistic parameters of the shot quality cost described in this work are defined for single-actor single-camera scenarios, the extension of $\costShot$ to multi-actor scenarios is trivial. It can be achieved by defining an artistic guideline $\Path{shot}$ using multi-actor parameters such as the angles with respect to the line of action \cite{bowen2013grammar}, or geometric center of the targets. We detail more possible extensions of our work in Section~\ref{sec:conclusion}.

\subsubsection{Safety}

Given the online map $\grid$, we can obtain the truncated signed distance (TSDT) map $\map: \real^3 \to \real$ as described in Section~\ref{sec:mapping}. Given a point $p$, we adopt the obstacle avoidance function from \cite{zucker2013chomp}. This function linearly penalizes the intersection with obstacles, and decays quadratically with distance, up to a threshold $\epsilon_\mathrm{obs}$:

\begin{equation}
c(p) = 
    \begin{cases}
      -\map(p)+\frac{1}{2} \epsilon_\mathrm{obs} & \map(p)<0\\
      \frac{1}{2\epsilon_\mathrm{obs}}(\map(p)-\epsilon_\mathrm{obs})^2 & 0<\map(p)\leq\epsilon_\mathrm{obs}\\
      0&\text{otherwise}
    \end{cases}
\end{equation}

Similarly to \cite{zucker2013chomp}, define a safety cost function for the entire trajectory:

\begin{equation}
    \costFnObs{\Path{q}, \map} = \int_{t=0}^{t_{f}} c(\Path{q}(t)) \abs{\frac{d}{dt}\Path{q}(t)}{} dt
\end{equation}



We can differentiate $\costObs$ with respect to a point at time $t_i$ to obtain the cost gradient (note that $\hat{v} = \frac{v}{|v|}$ denotes a normalized vector):

\begin{equation}
\begin{aligned}
    \nabla \costFnObs{\Path{q}(t_i),\map} &= \nabla \costFnObs{p_i,\map} = \abs{\dot{p_i}}{} \left [ \; \left ( I - \hat{\dot{p_i}} \hat{\dot{p_i}}^T \right ) \nabla c(p_i) -c(p_i) \kappa \;\right ] \;,\\
    &\text{where:}\quad \kappa = \frac{1}{\abs{\dot{p_i}}{}^2}(I-\hat{\dot{p_i}} \hat{\dot{p_i}}^T) \ddot{p_i}
\end{aligned}
\end{equation}

In practice, we use discrete derivatives to calculate $\map$, the velocities $\dot{p_i}$, and accelerations $\ddot{p_i}$.

\subsubsection{Occlusion avoidance} 
\begin{figure}[h!]
    \center
    \includegraphics[width=0.9\textwidth]{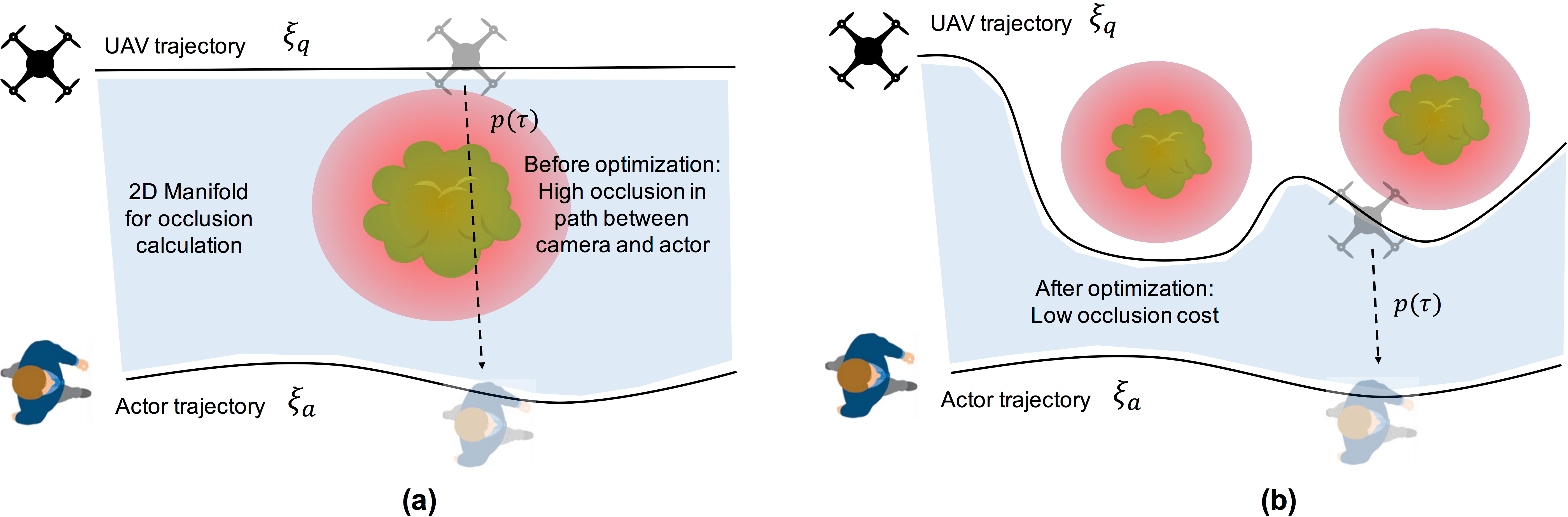}
    \caption{Occlusion cost representation. (a) For every pair of UAV-actor position, we integrate the penalization $c$ on the signed distance field over a 2D manifold. (b) After optimization, occlusion gradients pull the UAV trajectory towards regions with full visibility of the actor.}
    \label{fig:occlusion}
\end{figure}

Even though the concept of occlusion is binary, \textit{i.e}, we either have or don't have visibility of the actor, a major contribution of our past work \cite{bonatti2018autonomous} was to define a differentiable cost that expresses a viewpoint's occlusion intensity among arbitrary obstacle shapes. The fundamental idea behind this cost is that it measures along how much obstacle blockage the best possible camera viewpoints of $\Path{q}$ would go through, assuming the camera pointed directly at the actor's true position at all times. For illustration purposes, Figure~\ref{fig:occlusion} shows the concept of occlusion for motion in a 2D environment, even though our problem fully defined in 3D.

Mathematically, we define occlusion as the integral of the TSDT cost $c$ over a 2D manifold connecting both trajectories $\Path{q}$ and $\Path{a}$. The manifold is built by connecting each UAV-actor position pair at time $t$ using the parametrized path $p(\tau)$, where $p(\tau=0) = \Path{q}(t)$ and $p(\tau=1) = \Path{a}(t)$:

\begin{equation}
\costFnOcc{\Path{q},\Path{a},\map}
 = \int_{t=0}^{t_{f}} \left( \int_{\tau=0}^{1}\; c(p(\tau)) \abs{\frac{d}{d\tau} p(\tau)}{} d\tau \right)\, \abs{\frac{d}{dt}\Path{q}(t)}{} dt
\end{equation}




 We can derive the functional gradient with respect to a point $p_i$ at time $t_i$, resulting in:

\begin{equation}
\nabla \costFnOcc{\Path{q},\Path{a},\map}(t_i) = 
 \int_{\tau=0}^{1} \nabla c(p(\tau)) |L| |\dot{q}| \left [I - (\hat{\dot{q}}+\tau(\frac{\dot{a}}{|\dot{q}|}-\hat{\dot{q}}) ) \hat{\dot{q}}^T \right ]
- c(p(\tau)) |\dot{q}| \left [ \hat{L}^T + \frac{\hat{L}^T \dot{L} \hat{\dot{q}}{}^T}{|\dot{q}|} + |L|\kappa^T \right]
d\tau,
\end{equation}



where:

\begin{equation}
q=\Path{q}(t_i), \quad a=\Path{a}(t_i),\quad p(\tau) = (1-\tau)q + \tau a, , \quad L = a-q, \quad \hat{v} = \frac{v}{|v|}, \quad 
\kappa = \frac{1}{|\dot{q}|^2}(I-\hat{\dot{q}} \hat{\dot{q}}^T) \ddot{q}
\end{equation}


Intuitively, the term multiplying $\nabla c(p(\tau))$ is related to variations of the signed distance gradient in space, with the rest of the term acting as a lever to deform the trajectory. The term $c(p(\tau))$ is linked to changes in path length between camera and actor.

\subsection{Trajectory Optimization Algorithm}
\label{subsec:opt_algorithm}


Our objective is to minimize the total cost function $\costFn{\Path{q}}$ (\ref{eq:main_cost}). We do so by covariant gradient descent, using the gradient of the cost function $\nabla J(\Path{q})$, and a analytic approximation of the Hessian $\nabla^2 J(\Path{q}) \approx (A_\mathrm{smooth} + \lambda_3 A_\mathrm{shot})$:
\begin{equation}
  \Path{q}^+ = \Path{q} - \frac{1}{\eta} (A_\mathrm{smooth} + \lambda_1 A_\mathrm{shot})^{-1} \nabla J(\Path{q})
\end{equation}

In the optimization context, $\nabla^2 J(\Path{q})$ acts as a metric to guide the solution towards a direction of steepest descent on the functional cost. This step is repeated until convergence. We follow two conventional stopping criteria for descent algorithms based on current cost landscape curvature and relative cost decrease \cite{boyd2004convex}, and limit the maximum number of iterations. We use the current trajectory as initialization for the next planning problem, appending a segment with the same curvature as the final points of the trajectory for the additional points until the end of the time horizon.

Note in Algorithm~\ref{alg:optimize} one of the main advantages of the CHOMP algorithm \cite{ratliff2009chomp}: we only perform the Hessian matrix inversion once, outside of the main optimization loop, rendering good convergence rates \cite{bonatti2018autonomous}. By fine-tuning hyper-parameters such as trajectory discretion level, trajectory time horizon length, optimization convergence thresholds, and relative weights between costs, we can achieve a re-planning frequency of approximately $5$Hz considering a $10$s horizon. These are adequate parameters for safe and non-myopic operations in our environments, but lower or higher frequencies can be achieved with the same underlying algorithm depending on application-specific requirements.

\begin{algorithm}[H]
\label{alg:optimize}
\SetAlgoLined
 $M_{inv} \gets (A_\mathrm{smooth} + \lambda_3 A_\mathrm{shot})^{-1}$\;
 \For{$i=0,1,...,i_\mathrm{max}$}{
  \If{$(\nabla J(\Path{q}{}_{i})^T M_\mathrm{inv} \nabla J(\Path{q}{}_{i}))^2/2 < \epsilon_0$ or $(J(\Path{q}{}_{i}) - J(\Path{q}{}_{i-1})) < \epsilon_1$}{
   \textbf{return} $\Path{q}{}_{i}$\;
   }
  $\Path{q}{}_{i+1} = \Path{q}{}_{i} - \frac{1}{\eta} M_{inv} \nabla J(\Path{q}{}_{i})$\;
 }
 \caption{\small Optimize ($\Path{q}$)}
 \textbf{return} $\Path{q}{}_{i}$\;
\end{algorithm}

The resulting trajectory from the most recent plan is appended to the UAV's trajectory tracker, which uses a PD controller to send velocity commands to the aircraft's internal controller.

\section{Learning Artistic Shot Selection}
\label{sec:artistic}

In this section, we introduce a novel method for online artistic shot type selection. Parameter selection which specifies the shot type can be set before deployment with a fixed set of parameters $\Omega_\mathrm{art}$. However, using a fixed shot type renders undesirable results during operation since the UAV does not adapt to different configurations of obstacles in the environment. Instead, here we design an algorithm for selecting adaptive shot types, depending on the context of each scene.

\subsection{Deep Reinforcement Learning Problem Formulation}

\begin{figure}[h!]
    \centering
    \includegraphics[width=0.5\columnwidth]{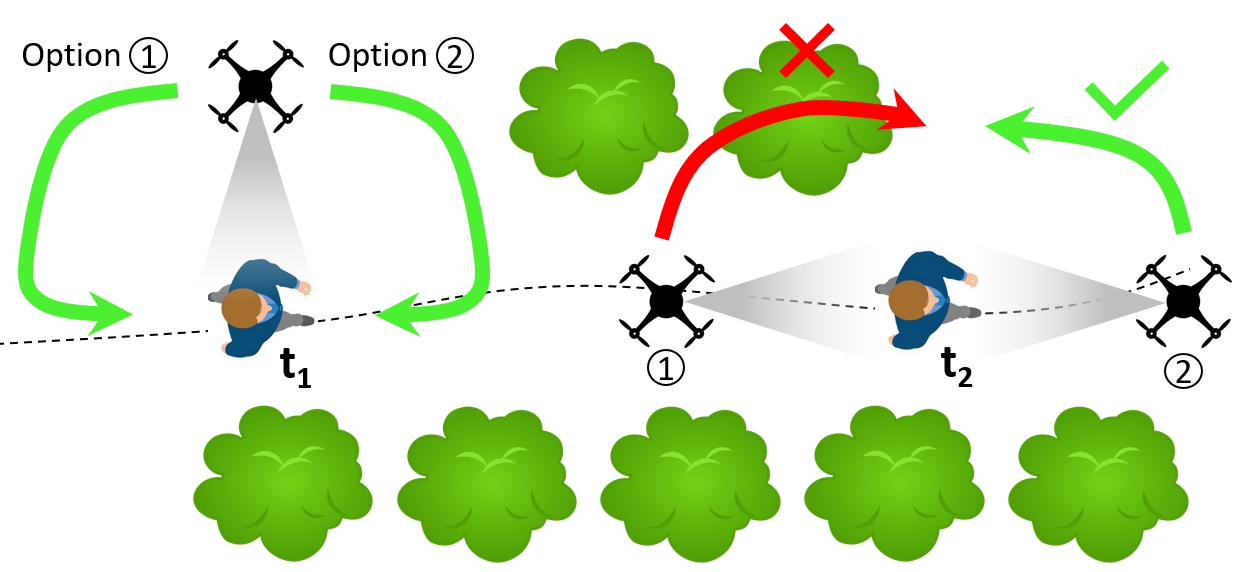}
    \caption{Example of artistic parameter selection as a sequential decision-making problem. The choice of frontal or back shots at time step $t_1$ influence the quality of left side shot choice at time step $t_2$.}
    \label{fig:rl_reason}
\end{figure}

As introduced in Section \ref{sec:problem_form}, the choice of artistic parameters is a time-dependent sequential decision-making problem. Decisions taken at the current time step influence the quality of choices in future states. Figure~\ref{fig:rl_reason} exemplifies the sequential nature of the problem.

We define the problem as a Contextual Markov Decision Process (C-MDP) \cite{krishnamurthy2016contextual}, and use an RL algorithm to find an optimal shot selection policy. Our goal is to learn a policy $\pi_{\theta}(a_t | c_t)$, parametrized by $\theta$, that allows an agent to choose action $a_t$, given scene context $c_t$, to select among a discrete set of UAV artistic parameters $\Omega_\mathrm{art}$. Our action set is defined as four discrete values of $\Omega$ relative to left, right, back, and frontal shots. These shot types define the relative yaw angle $\phi_\mathrm{rel}$, which is fed into the UAV's motion planner, as explained in Section~\ref{sec:planning}. 

We define state $c_t$ as the scene context, which is an observation drawn from the current MDP state $s_t$. 
The true state of the MDP is not directly observable because, to maintain the Markovian assumption, it encodes a diverse set of information such as: the UAV's full state and future trajectory, the actor's true state and future trajectory, the full history of shot types executed in past choices, a set of images that the UAV's camera has recorded so far, ground-truth obstacle map, environmental properties such as lighting and wind conditions, etc.
Therefore, our definition of context $c_t$ can be seen as a lower dimensional compression of $s_t$, given by a concatenation of following three elements:
\begin{enumerate}
  \item \emph{Height map:} a local 2.5D map containing the elevation of the obstacles near the actor;
  \item \emph{Current shot type:} four discrete values corresponding to the current relative position of the UAV with respect to the actor;
  \item \emph{Current shot count:} number of time steps the current shot type has been executed consecutively.
\end{enumerate}
We assume that states evolve according to the system dynamics: $s_{t+1} \sim p(s_t, a_t)$. 
Finally, we define the artistic reward $R_\mathrm{art}(v_t)$ where $v_t(s_t, a_t) = \{I_1,I_2,...,I_k\}$ is the video taken after the UAV executed action $a_t$ at state $s_t$. Our objective is to find the parameters of the optimal policy, which maximizes the expected cumulative reward:

\begin{equation} 
\label{eq_objective}
  \theta^* = \argmaxprob{\theta} \quad \expect{}{\sum_{t=1}^{T} R_\mathrm{art}(v_t)},
\end{equation}



where the expectation accounts for all randomness in the model and the policy. A major challenge for solving Equation~\ref{eq_objective} is the difficulty of explicitly modeling the state transition function $p(s_t, a_t)$. This function is dependent on variables such as the quadrotor and actor dynamics, the obstacle set, the motion planner's implicit behavior, the quadrotor and camera gimbal controllers, and the disturbances in the environment. In practice, we cannot derive an explicit model for the transition probabilities of the MDP. Therefore, we use a model-free method for the RL problem, using an action-value function $Q(c_t,a_t)$ to compute the artistic value of taking action $a_t$ given the current context $c_t$:

\begin{equation} \label{eq_qfunction}
Q(c_t,a_t)= \sum_{t'=t}^{T} \expect{\pi_{\theta}}{R_\mathrm{art}(v(c_{t'},a_{t'}))|c_t,a_t}
\end{equation}

The large size and complexity of the state space for our application motivates us to use a deep neural network with parameters $\theta$ to approximate $Q$: $Q(c_t,a_t) \approx f(c_t,a_t,\theta)$ \cite{sutton1998reinforcement,mnih2013playing}.

\subsection{Reward Definition}
\label{subsec:reward}

Now we define the artistic reward function $R\textsubscript{art}$. At a high level, we define the following basic desired aesthetical criteria for an incoming shot sequence: 
\begin{itemize}
    \item Keep the actor within camera view for as much time as possible;
    \item Maintain the tilt viewing angle $\theta\textsubscript{t}$ within certain bounds; neither too low nor too high above the actor;
    \item Vary the relative yaw viewing angle over time, in order to show different sides of the actor and backgrounds. Constant changes keep the video clip interesting. However, too frequent changes don't leave the viewer enough time to get a good overview of the scene; 
    \item Keep the drone safe, since collisions at a minimum destabilize the UAV, and usually cause complete loss of actor visibility due to a crash.
\end{itemize}

While these basic criteria represent essential aesthetical rules, they cannot account for all possible aesthetical requirements. The evaluation of a movie clip can be highly subjective, and depend on the context of the scene and background of the evaluator. Therefore, in this work we compare two different approaches for obtaining numerical reward values. In the first approach we hand-craft an arithmetical reward function $R_\mathrm{art}$, which follows the basic aesthetics requirements outlined above. In addition, we explore an alternative approach for obtaining $R_\mathrm{art}$ directly from human supervision. Next, we describe both methods.    


\paragraph{Hand-crafted reward:} The reward calculation from each control time step involves the analysis and evaluation of each frame of the video clip. Since our system operates with steps that last $6$s, the reward value depends on the evaluation of $180$ frames, given that images arrive at $30$Hz. We define $R_\mathrm{frame}$ as the sub-reward relative to each frame, and compute it using the following metrics:


\textit{Shot angle:} 
$R_\mathrm{frame}^\mathrm{shot}$ considers the current UAV tilt angle $\theta_\mathrm{rel}$ in comparison with an optimal value $\theta_\mathrm{opt}=15^{\circ}$ and an accepted tolerance $\theta_\mathrm{tol}=\pm10^{\circ}$ around it\footnote{The optimal value and bounds were determined by using standard shot parameters for aerial cinematography.}.
The shot angle sub-reward decays linearly and symmetrically between $1.0$ and $0.0$ from $\theta_\mathrm{opt}$ to the tolerance bounds.
Out of the bounds, we assign a negative penalty of $R_\mathrm{frame}^\mathrm{shot} = -0.5$.

\textit{Actor's presence ratio:} considers the screen space occupied by the actor's body. We set two bounds $pr_\mathrm{min}=0.05$ and $pr_\mathrm{max}=0.10$ based on a desired long-shot scale, actor size of $1.8$m, and the camera's intrinsic matrix. 
If the presence ratio lies within the bounds, we set the value of $R_\mathrm{frame} = R_\mathrm{frame}^\mathrm{shot}$. Otherwise, this parameter indicates that the current frame contains very low aesthetics value, with the actor practically out of the screen or occupying an exorbitant proportion of it. In that case, we set a punishment $R_\mathrm{frame} = -0.5$.


We average the resulting $R_\mathrm{frame}$ over all frames present in one control step to obtain an intermediate reward $R_\mathrm{step} = \frac{1}{N}\sum_{i=1}^{N} R_\mathrm{frame,\ i}$. Next, we consider the interaction between consecutive control steps to discount $R_\mathrm{step}$ using a third metric: shot type duration.

\begin{figure}[h]
\centering
\includegraphics[width=0.35\columnwidth]{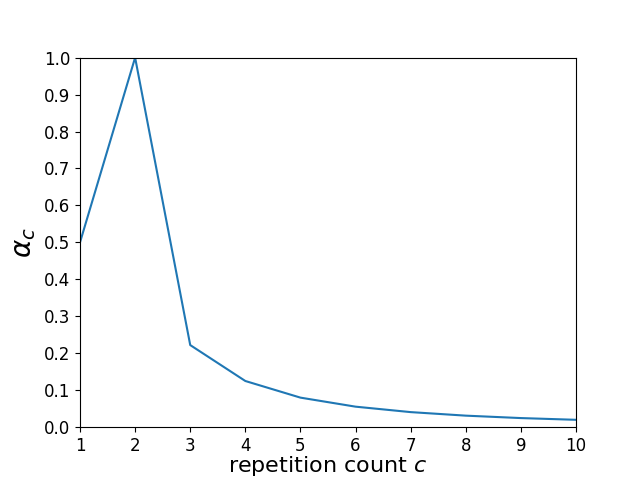}
\caption{Values of the variable discount parameter $\alpha_{c}$ over shot repetition counts $c$.}
\label{fig:reward_discount}
\end{figure}

\textit{Shot type duration:} considers the duration of the current shot type, given by the count of steps $c$ in which the same action was selected sequentially.
We use the heuristic that the ideal shot type has a length $12$s, or $c_\mathrm{opt} = 2$ time steps\footnote{This heuristics choice was based on informal tests with shot switching frequencies.}, and define a variable discount parameter $\alpha_{c}$, as seen in Fig.~\ref{fig:reward_discount}. High repetition counts are penalized quadratically to maintain the viewers interested in the video clip.

Eq.~\eqref{eq:reward_discounted} shows how we obtain the final artistic reward $R_\mathrm{art}$ for the current movie clip. If $R_\mathrm{step}$ is positive, $\alpha_{c}$ serves as a discount factor, with the aim of guiding the learner towards the optimal shot repetition count. 
In the case of negative $R_\mathrm{step}$, we multiply the reward by the inverse $\alpha_{c}$, with the objective of accentuating the penalization, and to incentivize the policy to quickly recover from executing bad shot types repetitively. 

\begin{equation} 
\label{eq:reward_discounted}
R_\mathrm{art} = \begin{cases}
             R_\mathrm{step} \cdot \alpha_{c} , & \text{if } R_\mathrm{step}\geq 0\\
            \frac{R_\mathrm{step}}{\alpha_{c}}, & \text{otherwise.}
        \end{cases}
\end{equation}

In the eventual case of a UAV collision during the control step we override the reward calculation procedure to only output a negative reward of $R_\mathrm{art} = -1.0$.

\begin{figure}[h!]
    \centering
    \begin{minipage}{0.4\columnwidth}
    \centering
    \includegraphics[width=0.7\columnwidth]{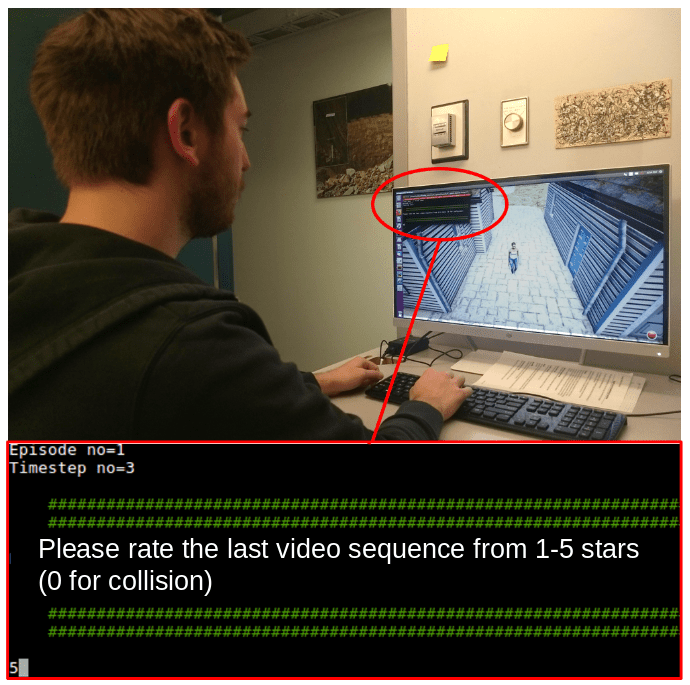}
    \end{minipage}
    \caption{Evaluators rate video clips during the training procedure using an interface on the screen.}
    \label{fig:human_reward}
\end{figure}

\paragraph{Human supervision reward:} We also explore a reward scheme for video segments based solely on human evaluation. 
We create an interface (Fig.~\ref{fig:human_reward}) in which the user gives an aesthetics score between 1 (worst) and 5 (best) to the video generated by the previous shot selection action. 
The score is then linearly mapped to a reward $R_\mathrm{art}$ between $-0.5$ and $1.0$ to update the shot selection policy in the RL algorithm. In the case of a crash during the control step, we override the user's feedback with a penalization of $R_\mathrm{art}=-1.0$.

\subsection{Implementation Details}

\begin{figure}[h!]
\centering
\includegraphics[width=0.6\columnwidth]{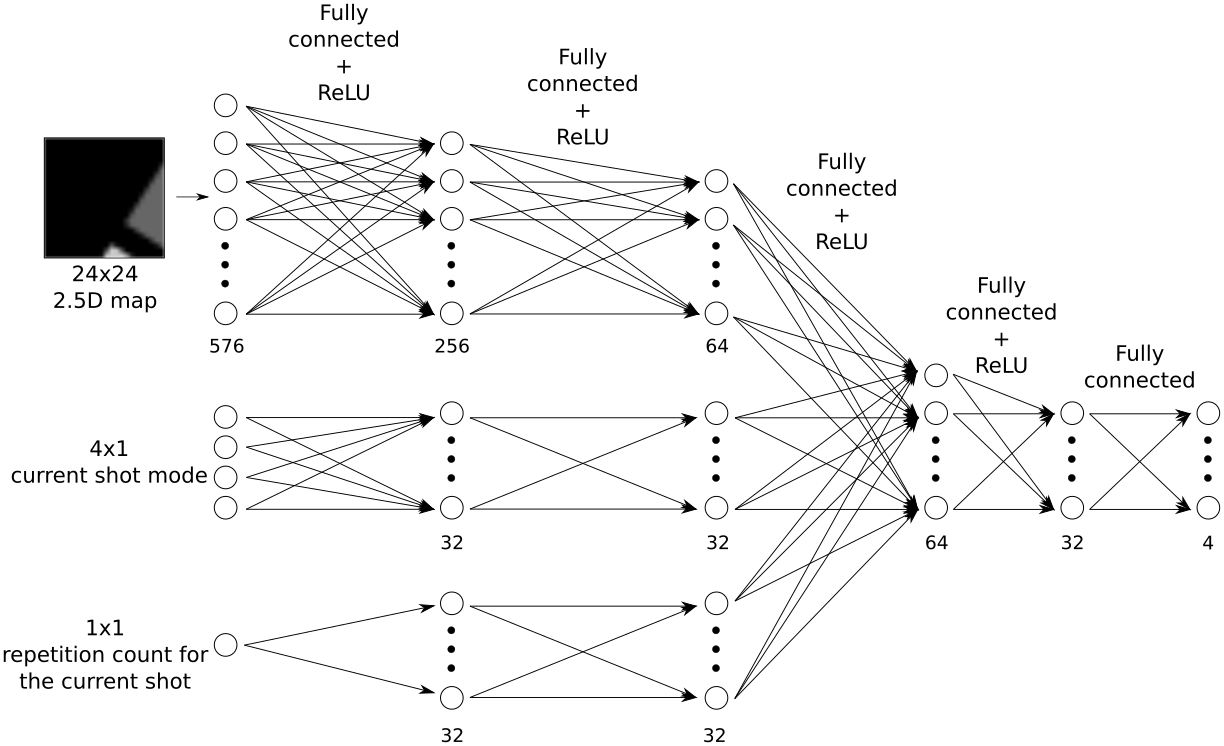}
\caption{DQN architecture consisting of fully connected layers. Each portion of state is first fed to separate mini-networks of two layers. Then, the outputs of these are combined in three consecutive layers whose output is the Q values.}
\label{fig:dqn_arch}
\end{figure}

Our DQN architecture is composed of different linear layers which combine the state inputs, as seen in Figure~\ref{fig:dqn_arch}. We use ReLU functions after each layer except for the last, and use the Adam optimizer \cite{kingma2014adam} with Huber loss \cite{huber1964robust} for creating the gradients. We use an experience replay (ER) buffer for training the network, such as the one described by \cite{mnih2015human}.

\section{Experimental Results} 
\label{sec:experiments}

In this section we detail integrated experimental results, followed by detailed results on each subsystem.


\subsection{Integrated System Results}
\label{subsec:experiments_integrated}

\begin{figure}[h!]
    \centering
    \includegraphics[width=0.8\textwidth]{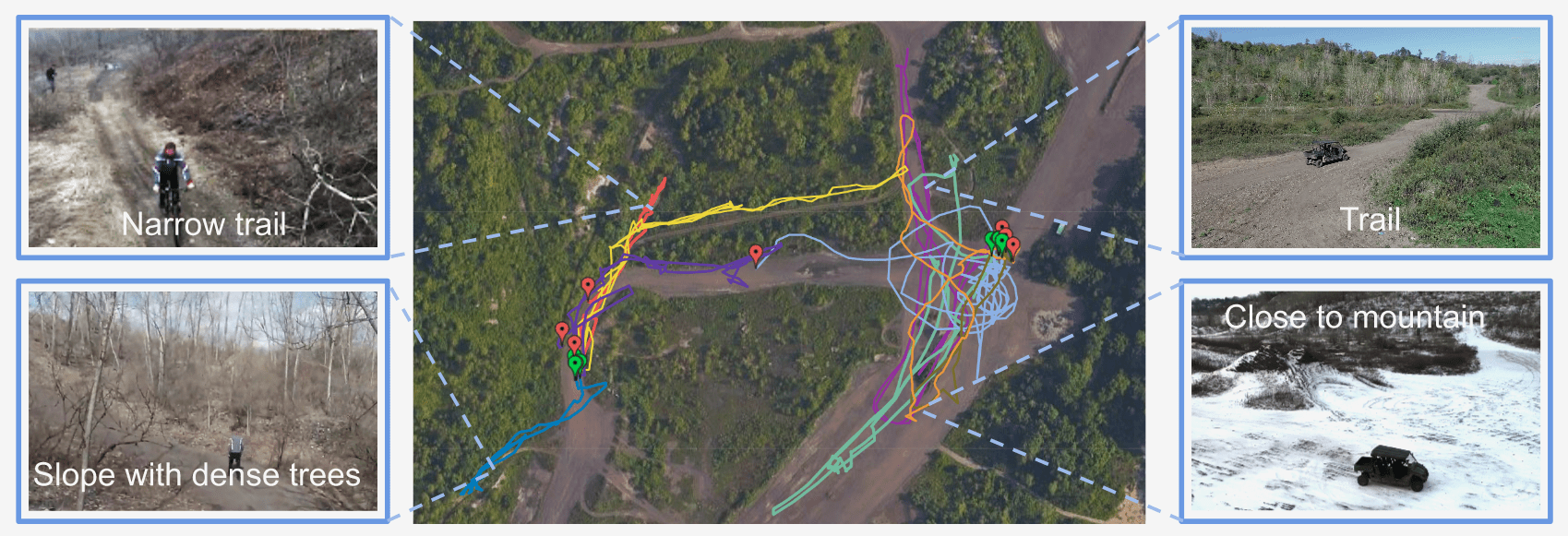}
    \caption{Testing facility. The middle figure shows a top-down satellite view of testing terrain, overlayed with UAV positions from different trials. We accumulated over $2$h of flight time, and a total distance of almost $6$km. The side figures show the diverse types of terrains in our experiments. The figures also depict different actors and different seasons. }
    \label{fig:gascola_tests}
\end{figure}

We conducted a series of field trials to evaluate our integrated system in real-life conditions. We used a large open facility named Gascola in Penn Hills, PA, located about $20$ min east of Pittsburgh, PA. The facility has a diverse set of obstacle types and terrain types such as several off-road trails, large mounds of dirt, trees, and uneven terrain. Figure~\ref{fig:gascola_tests} depicts the test site and shows the different areas where the UAV flew during experiments. We summarize the test's objectives and results in Table~\ref{tab:int_results}, and indicate which results explain our initial hypotheses from Subsection~\ref{subsec:hyp}. 

\begin{table}[h!]
\centering
\caption{Objectives and results for integrated experiments.}
\begin{tabular}[t]{l|ll}   
\hline
\textbf{Objectives} & \textbf{Results}\\
\hline
\begin{tabular}[t]{@{}l@{}}Stay safe among unstructured  \\ obstacles sensed online \\\textit{(addresses Hyp.1)} \end{tabular} & 
\begin{tabular}[t]{@{}l@{}} Avoided all obstacles successfully, including trees,  \\  dirt mound, slopes, posts. See Figs. \ref{fig:exp_mural}-\ref{fig:exp_time_lapse}. \end{tabular} \\
\hline   
\begin{tabular}[t]{@{}l@{}}Avoid occlusions among \\ any obstacle shape \\\textit{(addresses Hyp.1)}\end{tabular} &
\begin{tabular}[t]{@{}l@{}}Planner maintained actor visibility. \\ See Fig.~\ref{fig:exp_mural}. More results in Subsec~\ref{subsec:experiments_planner}. \end{tabular}\\
\hline
\begin{tabular}[t]{@{}l@{}}Process data fully onboard \\\textit{(addresses Hyp.1)}\end{tabular} & 
\begin{tabular}[t]{@{}l@{}} Data processed solely on onboard computer. \\ Table~\ref{tab:statistics} and Fig.~\ref{fig:dt_time} show system statistics.\end{tabular} 
\\
\hline
\begin{tabular}[t]{@{}l@{}}Operate with different types \\ of actors at different speeds \\\textit{(addresses Hyps.2-3)}\end{tabular}
& \begin{tabular}[t]{@{}l@{}} Person, car, bikes shot at high-speed\\ chases. See Figs. \ref{fig:exp_mural}-\ref{fig:exp_rl_real}. \end{tabular}\\  
\hline
\begin{tabular}[t]{@{}l@{}}Execute different shot types \\\textit{(addresses Hyp.3)} \end{tabular} & 
\begin{tabular}[t]{@{}l@{}}Successful recording of back, right, front, circling \\shots (Fig. \ref{fig:exp_mural}). Smooth shot transitions (Fig.~\ref{fig:exp_rl_real}).\end{tabular}
\\   
\hline
\begin{tabular}[t]{@{}l@{}}Automatically select artistically \\ meaningful shot types \\\textit{(addresses Hyp.3)}\end{tabular}   & 
\begin{tabular}[t]{@{}l@{}} Policy adapted to current context to\\ produce a visually aesthetic video. Fig.~\ref{fig:exp_rl_real}. \end{tabular} \\
\hline
\end{tabular}
\label{tab:int_results}  
\end{table}

\begin{figure}[h!]
    \centering
    \includegraphics[width=1.0\textwidth]{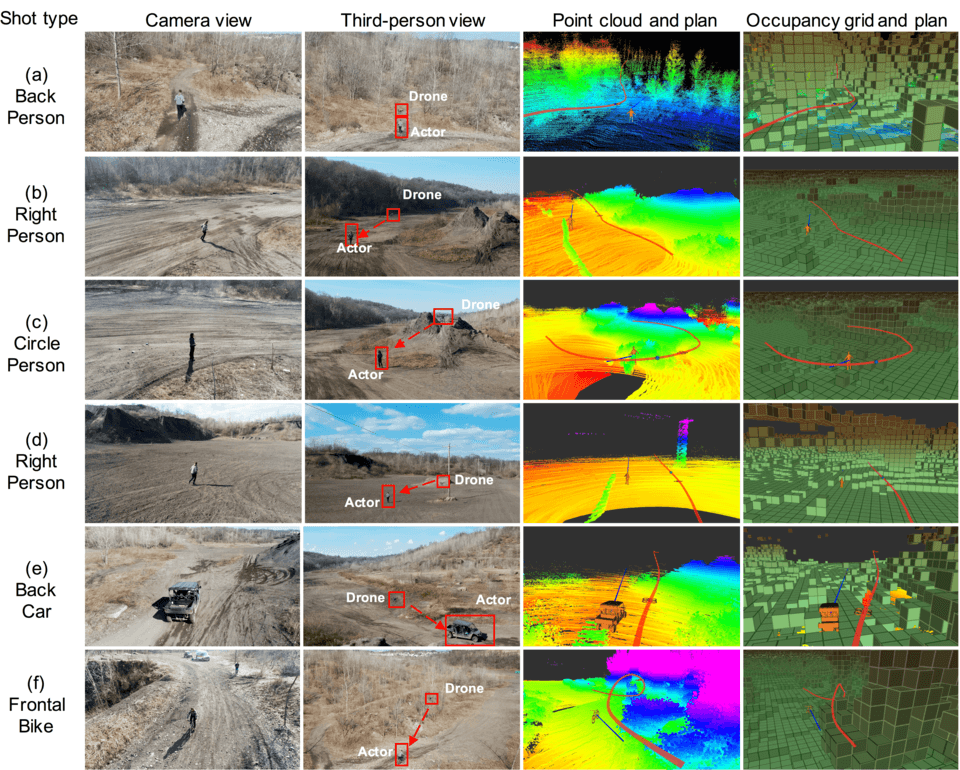}
    \caption{Field results with different fixed shot types in multiple environment types, following different actor types. The UAV trajectory (red) tracks the actor's forecasted motion (blue), and stays safe while avoiding occlusions from obstacles. We display accumulated point clouds of LiDAR hits and the occupancy grid: a) Back shot following runner in narrow tree trail; b) Right side shot following a runner close to dirt mound; c) Circular shot on person close to dirt mound; d) Right side shot below the 3D structure of an electric wire. Note that LiDAR registration is noisy close to the pole in row due to large electromagnetic interference with the UAV's compass; e) Right side shot following car close to dirt mound; f) Frontal shot on biker going downhill on a trail with tall trees.}
    \label{fig:exp_mural}
\end{figure}

\begin{figure}[h!]
    \centering
    \includegraphics[width=0.9\textwidth]{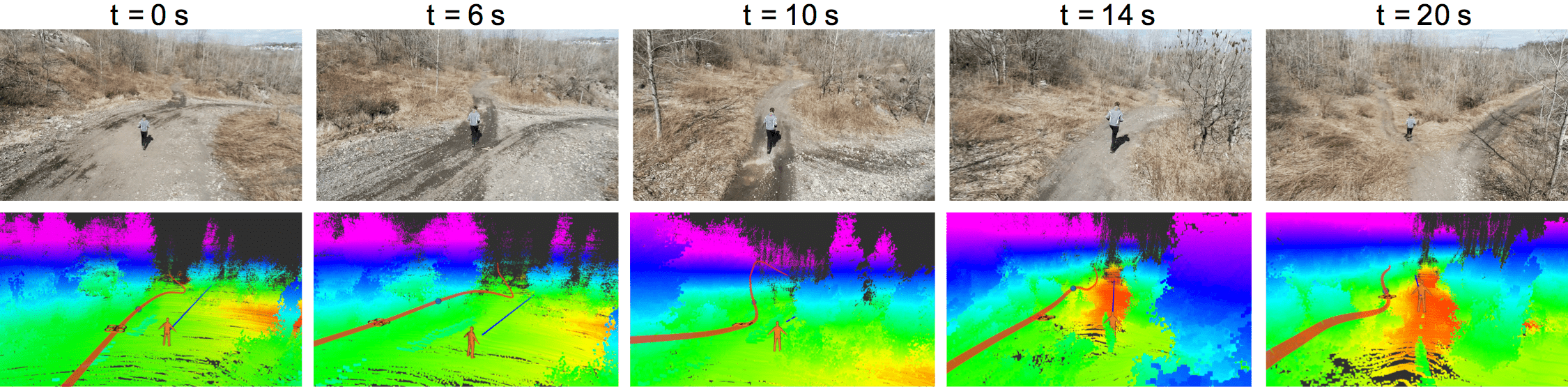}
    \caption{Detailed time lapse of back shot following a runner in a narrow trail with trees. As the UAV approaches the trees at $t=6$s, the trajectory bends to keep the vehicle safe, maintain target visibility, and follow the terrain's downwards inclination.}
    \label{fig:exp_time_lapse}
\end{figure}

Figure~\ref{fig:exp_mural} summarizes our experiments conducted with fixed shot types.  We employ a variety of shot types and actors, while operating in a wide range of unstructured environments such as open fields, in proximity to a large mound of dirt, on narrow trails between trees and on slopes. In addition, Figure~\ref{fig:exp_time_lapse} provides a detailed time lapse of how the planner's trajectory output evolves during flight through a narrow trail between trees.

\begin{figure}[h!]
    \centering
    \includegraphics[width=0.9\textwidth]{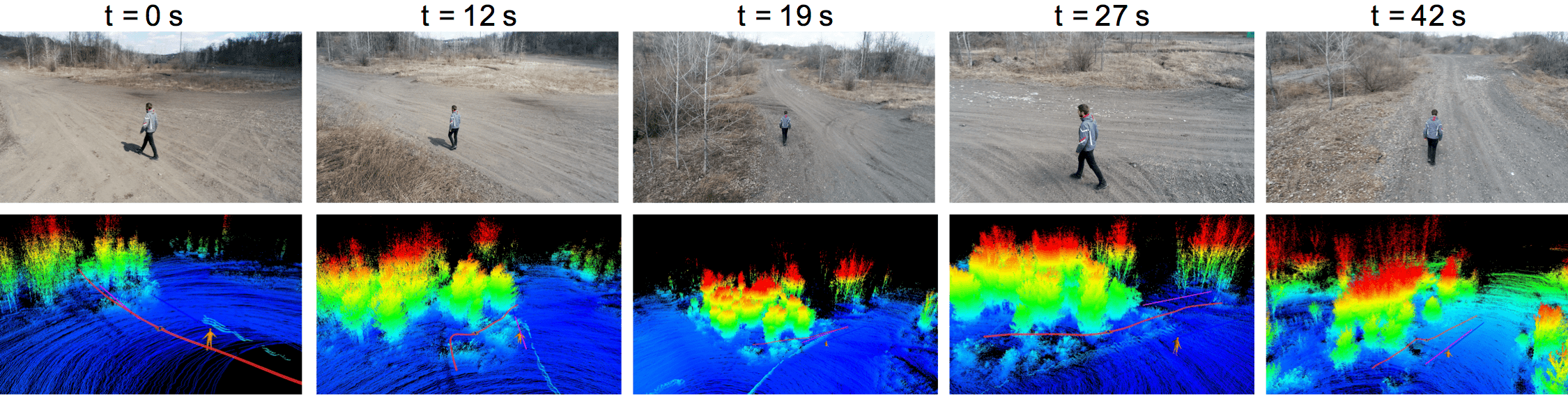}
    \caption{Field test results using the online artistic shot selection module. UAV trajectory is shown in red, actor motion forecast in blue, and desired shot reference in pink. The UAV initially does a left side shot at $t=0$s in the open field, but as a line of trees appear, it switches to a back shot at $t=12$s. When an opening appears among the tree lines, the UAV selects a left side shot again at $t=27$s, and when the clearance ends, the module selects a back shot again.}
    \label{fig:exp_rl_real}
\end{figure}

Figure~\ref{fig:exp_rl_real} shows experiments where we employed the online automatic artistic selection module. In-depth results of this module are described in Subsection~\ref{subsec:experiments_arts}.

\begin{figure}[h!]
    \centering
    \includegraphics[width=0.40\textwidth]{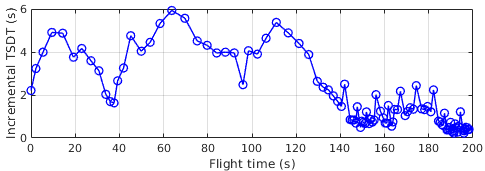}
    \caption{Incremental distance transform computation time over flight time. The first operations take significantly more time because of our map initialization scheme where all cells are initially considered as unknown instead of free, causing the first laser scans to update a significantly larger number of voxels than later scans. During calculation the planner is not blocked: it can still access TSDT values from the latest version of $\map$.}
    \label{fig:dt_time}
\end{figure}

We also summarize our integrated system's runtime performance statistics in Table~\ref{tab:statistics}, and discuss details of the online mapping performance in Figure~\ref{fig:dt_time}. Videos of the system in action can be found attached with the submission, or online at \small \href{https://youtu.be/ookhHnqmlaU}{https://youtu.be/ookhHnqmlaU}\normalsize.

\begin{table}[h!]
\centering
\caption{System statistics recorded during flight time on onboard computer.}
\begin{tabular}{l|llll|l}
\label{tab:statistics}
 \textbf{\pbox{0.2cm}{System}} & \textbf{\pbox{1.0cm}{Module}}& \textbf{\pbox{2.0cm}{CPU\\Thread (\%)}} & \textbf{\pbox{2.0cm}{RAM (MB)}}  & \textbf{\pbox{2.0cm}{Runtime (ms)}} & \textbf{\pbox{2.0cm}{Target freq. (Hz)}}\\
\hline
  & Detection & 57 & 2160  & 145 & \\

Vision  & Tracking & 24 & 25  & 14.4 & 15\\

 & Heading & 24 & 1768  & 13.9 & \\

  & KF & 8 & 80  & 0.207 & \\

\hline

 & Grid & 22 & 48  & 36.8 & \\

Mapping & TSDF & 91 & 810  & 100-6000 & 10\\

  & LiDAR & 24 & 9 & 100  &  \\

\hline

Planning  & Planner & 98 & 789  & 198 & 5\\

\hline

Controls  & DJI SDK & 89 & 40  & NA & 50\\

\hline

Shot selection  & DQN & 4 & 1371  & 10.0 & 0.16\\

\hline

\end{tabular}
\end{table}

From the field experiments, we verify that our system achieved all system-level objectives in terms of safely and robustly executing a diverse set of aerial shots with different actors and environments. Our data also confirms the questions raised to validate our hypotheses: onboard sensors and computing power sufficed to provide smooth plans, producing artistically appealing images.

Next we present detailed results on the individual sub-systems of the aircraft.

\subsection{Visual Actor Localization and Heading Estimation}
\label{subsec:experiments_heading}

Here we detail dataset collection, training and testing of the different subcomponents of the vision module. We summarize the vision-specific test's objectives and results in Table~\ref{tab:vision_objectives}.

\begin{table}[h!]
\caption{Objectives and results for vision-specific experiments.}
\centering
\begin{tabular}{l|l}
\hline   
\textbf{Objectives} & \textbf{Results}\\
\hline
\begin{tabular}[c]{@{}l@{}}Compare object detection \\network architectures\end{tabular} 
& \begin{tabular}[c]{@{}l@{}} Faster-RCNN showed significantly better\\ performance than SSD architecture (Fig. \ref{fig:rec_pre_curve}).\end{tabular} \\
\hline
\begin{tabular}[c]{@{}l@{}}Compare supervised and semi-supervised\\ methods for heading estimation \end{tabular} 
& \begin{tabular}[c]{@{}l@{}} Semi-supervised method has smoother output\\ and higher accuracy(Table~\ref{tab:semi_finetune} and Fig.~\ref{fig:comp_finetune} )\end{tabular}
\\
\hline
\begin{tabular}[c]{@{}l@{}}Analyze the amount of labeled data \\ needed for semi-supervised training \end{tabular} 
& \begin{tabular}[c]{@{}l@{}} Loss increased by less than $\sim8\%$ when we trained\\ the model with $1/10$ of labeled data (Fig.~\ref{fig:semi_compare}) \end{tabular}
\\
\hline
\begin{tabular}[c]{@{}l@{}} Validate integrated 3D pose estimator \\using image projections \end{tabular}   & 
\begin{tabular}[c]{@{}l@{}} Error of less than $1.7$m in estimated actor path length\\ over a $40$m long ground-truth actor trajectory (Fig.~\ref{fig:vision_results}).\end{tabular}
\\
\hline
\end{tabular}
\label{tab:vision_objectives}
\end{table}

\subsubsection{Object detection network}

\textit{Dataset collection.}
We trained the network on the COCO dataset~\cite{lin2014microsoft}, and fine-tuned it with a custom aerial filming dataset. To test, we manually labeled 120 images collected from our aerial filming experiments, with bounding box over people and cars. 

\begin{figure}[h!]
    \centering
    \includegraphics[width=0.4\textwidth]{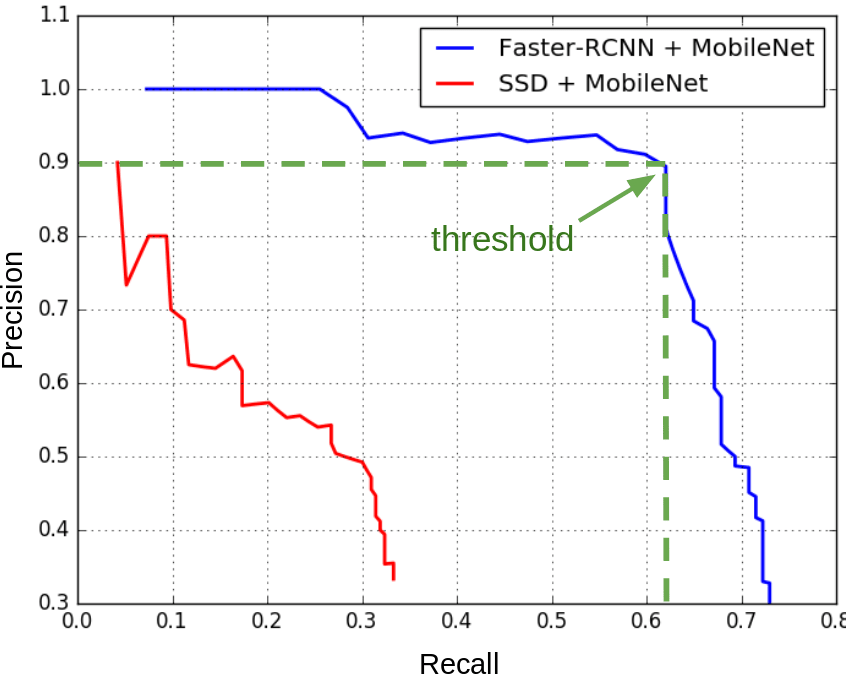}
    \caption{Precision recall curve for object detection. The results are tested on the filming testing data, which contains many challenging cases. 
    }
    \label{fig:rec_pre_curve}
\end{figure}

\textit{Training procedure:} We trained and compared two architectures: one based on Faster-RCNN, another on SSD. As mentioned in Section~\ref{sec:vision}, we simplify feature extraction with MobileNet-based structure to improve efficiency. First we train both structures on the COCO dataset. While the testing performance is good on the COCO testing dataset, the performance shows a significant drop, when tested on our aerial filming data. The network has a low recall rate (lower than 33\%) due to big angle of view, distant target, and motion blur. 
To address the generalization problem, we augmented the training data by adding blur, resizing and cropping images, and modifying colors. After training on a mixture of COCO dataset \cite{lin2014microsoft} and our own custom dataset as described in Section~\ref{sec:vision}, 
Figure~\ref{fig:rec_pre_curve} show the recall-precision curve of the two networks when tested on our filming testing data. The SSD-based network has difficulties detecting small objects, an important need for aerial filming. Therefore, we use Faster-RCNN-based network in our experiments and set the detection threshold to precision=0.9, as shown with the green arrow in Figure~\ref{fig:rec_pre_curve}. 

\subsubsection{Heading direction estimation (HDE) network}

\paragraph{Dataset collection:} We collected a large number of image sequences from various sources. For the person HDE, we used two surveillance datasets: VIRAT \cite{oh2011large} and DukeMCMT~\cite{ristani2016MTMC}, and one action classification dataset: UCF101 \cite{soomro2012ucf101}. We manually labeled 453 images in the UCF101 dataset as ground-truth HDE. As for the surveillance datasets, we adopted a semi-automatic labeling approach where we first detected the actor in every frame, then computed the ground-truth heading direction based on the derivative of the subject's position over a sequence of consecutive time frames. For the car HDE we used two surveillance datasets, VIRAT and Ko-PER~\cite{strigel2014ko}, in addition to one driving dataset, PKU-POSS \cite{wang2016probabilistic}. Table~\ref{heading_datasets} summarizes our data.

\begin{table}[h!]
\caption{Datasets used in HDE study. *MT denotes labeling by motion tracking, HL denotes hand labels. 
}
\label{heading_datasets}
\begin{center}
\begin{tabular}{p{2.2cm} p{2.0cm} p{2.0cm} p{1.5cm} p{1.5cm}}
\hline
Dataset & Target & GT & No. Seqs & No. Imgs \\
\hline
VIRAT  & Car/Person &  MT*     & 650   &  69680 \\
UCF101 & Person     & HL(453)* & 940   & 118027  \\
DukeMCMT & Person   & MT*      & 4336  & 274313 \\
Ko-PER & Car   & $\checkmark$  & 12    & 18277 \\
PKU-POSS & Car & $\checkmark$  & -     & 28973 \\
\hline
\end{tabular}
\end{center}
\end{table}


\textit{Training the network:}

We first train the HDE network using only labeled data from the datasets shown in Table~\ref{heading_datasets}. Rows 1-3 of Table~\ref{tab:semi_finetune} display the results. Then, we fine-tune the model with unlabeled data to improve generalization.

\begin{figure}[h!]
    \centering
    \includegraphics[width=0.5\textwidth]{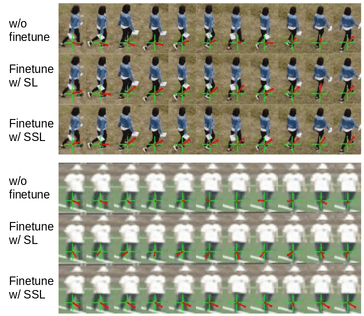}
    \caption{Three models are tested on the sequential data. Two testing sequences are shown in this figure. The top row of each testing sequence shows the results that directly employ the model trained on other open accessible datasets to the aerial filming task. It generalizes poorly due to the distribution difference. The middle row and bottom row show the results after finetuning the model on the filming data with and without continuity loss, respectively. The model using continuity loss for finetuning (bottom row) outputs more accurate and smooth results. }
    \label{fig:comp_finetune}
\end{figure}

We collected 50 videos, each contains approximately 500 sequential images. For each video, we manually labeled 6 images. The HDE model is finetuned with both labeled loss and continuity loss, same as the training process on the open accessible datasets. 
We qualitatively and quantitatively show the results of HDE using semi-supervised finetuning in Figure~\ref{fig:comp_finetune} and Table~\ref{tab:semi_finetune}. The experiment verifies our model could generalize well to our drone filming task, with a average angle error of $0.359$ rad. Compared to the pure supervised learning, utilizing unlabeled data improves generalization and results in more robust and stable performance.

\textit{Baseline comparisons:}
We compare our HDE approach against two baselines. The first baseline Vanilla-CNN is a simple CNN inspired by \cite{choi2016human}. The second baseline CNN-GRU implicitly learns temporal continuity using a GRU network inspired by \cite{liu2017weighted}. One drawback for this model is that although it models the temporal continuity implicitly, it needs large number of labeled sequential data for training, which is very expensive to obtain. 

We employ three metrics for quantitative evaluation: 1) Mean square error (MSE) between the output $(\cos\theta, \sin\theta)$ and the ground truth $(\hat{\cos \theta}, \hat{\sin \theta})$. 2) Angular difference (AngleDiff) between the output and the ground truth. 3) Accuracy obtained by counting the percentage of correct outputs, which satisfies AngleDiff$<\pi/8$. We use the third metric, which allows small error, to alleviate the ambiguity in labeling human heading direction. 

Vanilla-CNN \cite{choi2016human} and CNN-GRU \cite{liu2017weighted} baselines trained on open datasets don't transfer well to drone filming dataset with accuracy below 30\%. Our SSL based model trained on open datasets achieves 48.7\% accuracy. By finetuning on labelled samples of drone filming, we improve this to 68.1\%. Best performance is achieved by finetuning on labelled and  unlabeled sequences of the drone filming data with accuracy of 72.2\% (Table~\ref{tab:semi_finetune}).

\begin{table}[h!]
\caption{Semi-Supervised Finetuning results}
\label{tab:semi_finetune}
\begin{center}
\begin{tabular}{c p{2.0cm} p{2.5cm} p{2.5cm}}
\hline
Method & MSE loss & AngleDiff (rad) & Accuracy (\%) \\
\hline 
Vanilla-CNN w/o finetune & 0.53 & 1.12 & 26.67\\
CNN-GRU w/o finetune & 0.5 & 1.05 & 29.33 \\
SSL w/o finetune & 0.245 & 0.649 & 48.7  \\
SL w/ finetune  & 0.146 & 0.370 & 68.1 \\
SSL w/ finetune  & \textbf{0.113} & \textbf{0.359} & \textbf{72.2} \\
\hline
\end{tabular}
\end{center}
\end{table}

\textit{Reduction in required labeled data using semi-supervised learning:} Following Section~\ref{sec:vision}, we show how semi-supervised learning can significantly decrease the number of labeled data required for the HDE task. 

\begin{figure}[h!]
    \centering
    \includegraphics[width=0.6\textwidth]{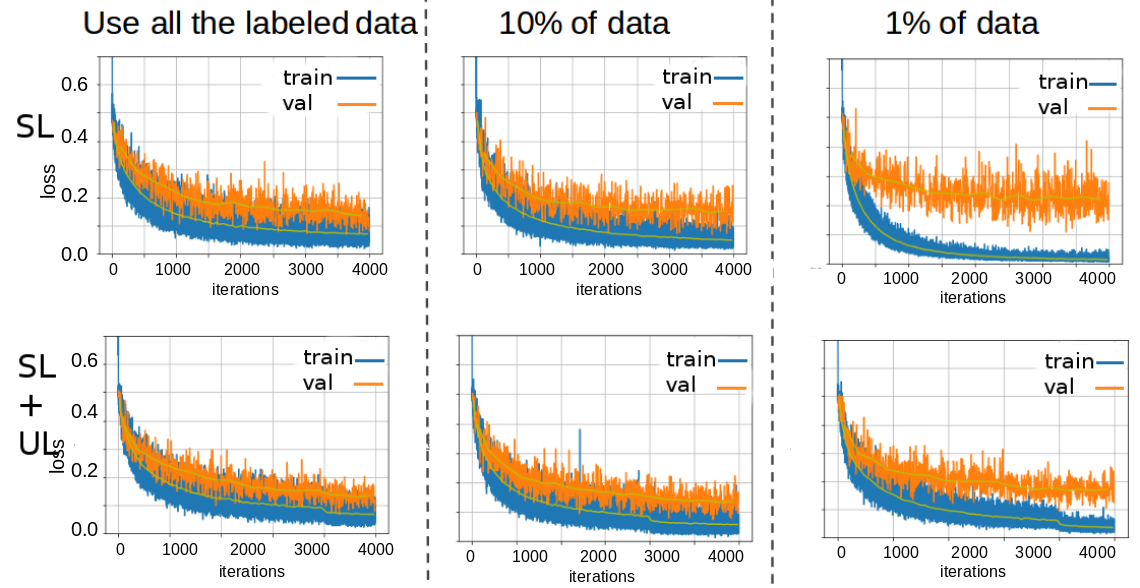}
    \caption{The top row shows training and validation loss for supervised learning using different number of labeled data. The validation performance drops from 0.13 to 0.22, when decreasing the number of labeled data from 100\% to 1\%. The bottom row shows results with semi-supervised learning. The validation losses are 0.13, 0.14 and 0.17 respectively for 100\%, 10\% and 1\% labeled data. }
    \label{fig:semi_compare}
\end{figure}

In this experiment, we train the HDE network on the DukeMCMT dataset, which consists of 274k labeled images from 8 different surveillance cameras. We use the data from 7 cameras for training, and one for testing (about 50k). Figure~\ref{fig:semi_compare} compares result from the proposed semi-supervised method with a supervised method using different number of labeled data. We verify that by utilizing unsupervised loss, the model generalizes better to the validation data than the one with purely supervised loss. 

As mentioned, in practice, we only use 50 unlabeled image sequences, each containing approximately 500 sequential images, and manually labeled 300 of those images. We achieve comparable performance with purely supervised learning methods, which require more labeled data. 

\subsubsection{3D pose estimation}

Based on the detected bounding box and the actor's heading direction in 2D image space, we use ray-casting method to calculate the 3D pose of the actor, given the online occupancy map and the camera pose. We assume the actor is in a upward pose, in which case the pose is simplified as $(x, y, z, \psi_a^w)$, which represents the position and orientation in the world frame. 

\begin{figure}[h!]
    \centering
    \includegraphics[width=0.6\textwidth]{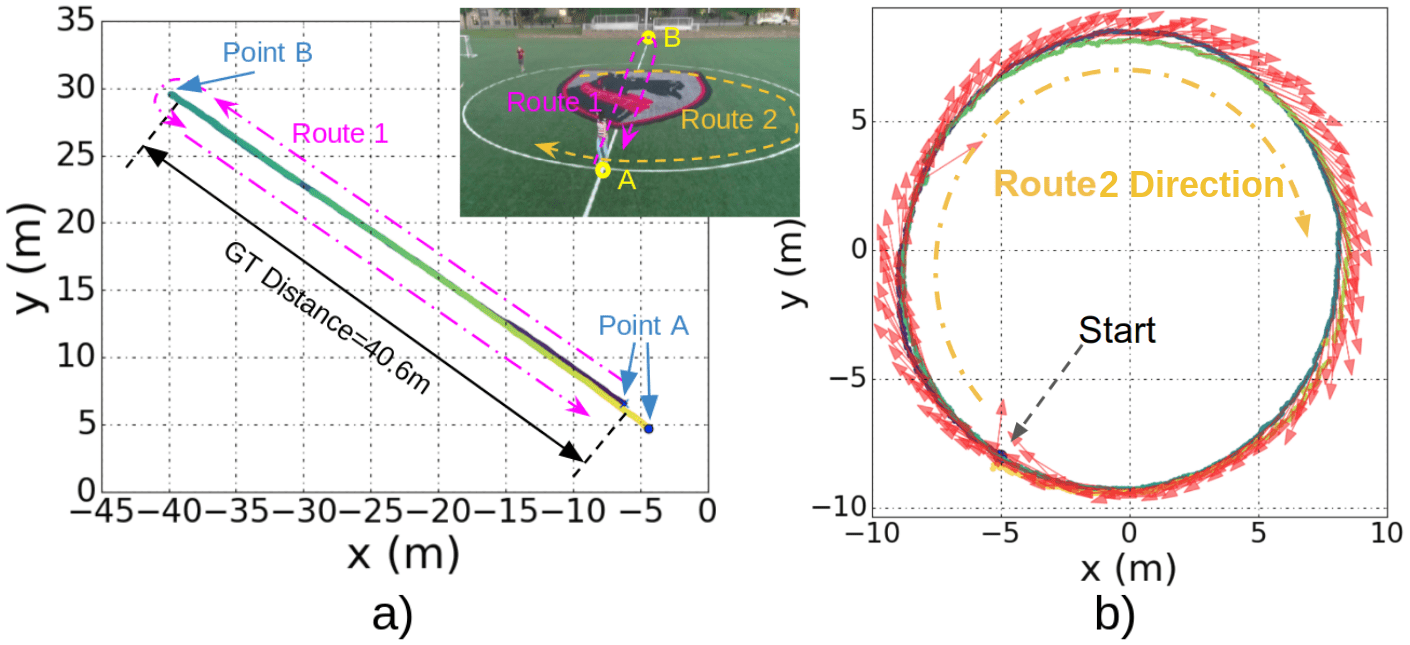}
    \caption{Pose and heading estimation results. a) Actor walks on a straight line from points A-B-A. Ground-truth trajectory length is 40.6m, while the estimated motion length is 42.3m. b) The actor walks along a circle. Ground-truth diameter is 18.3m, while the estimated diameter from ray casting is 18.7m. Heading estimation appears tangential to the ground circle.}
    \label{fig:vision_results}
\end{figure}

We validate the precision of our 3D pose estimation in two field experiments where the drone hovers and visually tracks the actor. First, the actor walks between two points along a straight line, and we compare the estimated and ground truth path lengths. Second, the actor walks in a circle at the center of a football field, and we compute the errors in estimated position and heading direction. Figure~\ref{fig:vision_results} shows our estimation error is less than 5.7\%. 

\subsection{Planner Evaluation}
\label{subsec:experiments_planner}


Next we present detailed results on different aspects of the UAV's motion planner. Table~\ref{tab:userstudy_objectives} summarizes the experiments' objectives and results.

\begin{table}[h!]
\caption{Objectives and results for detailed motion planner experiments.}
\centering
\begin{tabular}{l|l}
\hline   
\textbf{Objectives} & \textbf{Results}\\
\hline
\begin{tabular}[c]{@{}l@{}}Performance comparison between online\\  vs. ground-truth map\end{tabular} 
& \begin{tabular}[c]{@{}l@{}} Similar path quality, with increase in planning \\ time. See Fig.~\ref{fig:comp_map} and Table~\ref{tab:comp}.\end{tabular}\\
\hline
\begin{tabular}[c]{@{}l@{}}Performance comparison between noisy actor \\ forecast vs. ground-truth actor positioning\end{tabular} 
& \begin{tabular}[c]{@{}l@{}} Similar path quality: smoothness cost handles\\  noisy inputs. See Fig.~\ref{fig:comp_actor}.\end{tabular}
\\
\hline
Confirm ability to operate in full 3D environments  & 
\begin{tabular}[c]{@{}l@{}} Can fly under 3D obstacles, not\\  only height maps. See Fig.~\ref{fig:exp_mural}-d.\end{tabular}
\\
\hline
\begin{tabular}[c]{@{}l@{}}Performance comparison between different\\  planning time horizons\end{tabular} 
& \begin{tabular}[c]{@{}l@{}} Longer horizons significantly improve\\   path quality. See Fig.~\ref{fig:planning_horizons} and Table~\ref{tab:horizon}.\end{tabular}
\\
\hline
\begin{tabular}[c]{@{}l@{}}Test impact of occlusion cost\\  function on actor visibility\end{tabular} 
& \begin{tabular}[c]{@{}l@{}} Occlusion cost significantly improves path\\ quality. See Fig.~\ref{fig:stats}, Fig.~\ref{fig:occlusion_3}, and Table~\ref{tab:occlusion}.\end{tabular}
 \\
\hline
\end{tabular}
\label{tab:userstudy_objectives}
\end{table}





\begin{figure}[h!]
    \centering
    \includegraphics[width=0.3\textwidth]{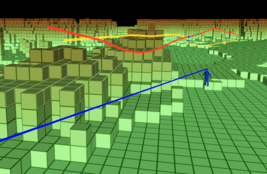}
    \caption{Performance comparisons between planning with full knowledge of the map (yellow) versus with online mapping (red), displayed over ground truth map grid. Online map trajectory is less smooth due to imperfect LiDAR registration and new obstacle discoveries as flight progresses.}
    \label{fig:comp_map}
\end{figure}

\textit{Ground-truth obstacle map vs. online map:} We compare average planning costs between results from a real-life test where the planner operated while mapping the environment in real time with planning results with the same actor trajectory but with full knowledge of the map beforehand. Results are averaged over 140 s of flight and approximately 700 planning problems. Table~\ref{tab:comp} shows a small increase in average planning costs with online map, and Fig~\ref{fig:comp_map} shows that qualitatively both trajectories differ minimally. The planning time, however, doubles in the online mapping case due to mainly two factors: extra load on CPU from other system modules, and delays introduced by accessing the map that is constantly being updated. Nevertheless, computation time is low enough such that the planning module can still operate at the target frequency of $5$ Hz.

\begin{table}[h!]
\centering
\caption{Performance comparison between ground-truth and online map.}
\begin{tabular}{l|llllll}
\label{tab:comp}
 \textbf{\pbox{4cm}{Planning condition}} & \textbf{\pbox{4cm}{Avg. plan time(ms)}}& \textbf{\pbox{2cm}{Avg. cost}} & \textbf{\pbox{3cm}{Median cost}}  \\
\hline
Ground-truth map  & 32.1 & 0.1022 & 0.0603 \\
Online map  & 69.0 & 0.1102 & 0.0825 \\
\hline
\end{tabular}
\end{table}

\begin{figure}[h!]
    \centering
    \includegraphics[width=0.3\textwidth]{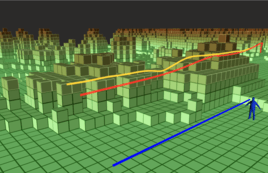}
    \caption{Performance comparison between planning with perfect ground truth of actor's location (red) versus noisy actor estimate with artificial noise of 1m amplitude (yellow). The planner is able to handle noisy actor localization well due to smoothness cost terms, with final trajectory similar to ground-truth case.}
    \label{fig:comp_actor}
\end{figure}

\textit{Ground-truth actor pose versus noisy estimate:} We compare the performance between simulated flights where the planner has full knowledge of the actor's position versus artificially noisy estimates with $1$m amplitude of random noise. The qualitative comparison with the actor's ground-truth trajectory shows close proximity of both final trajectories, as seen in Fig~\ref{fig:comp_actor}.

\textit{Operation on unstructured 3D map:} As seen in Figure~\ref{fig:exp_mural}d, our current system is able to map and avoid unstructured obstacles in 3D environments such as wires and poles. This capability is a significant improvement over previous work that only deals with ellipsoidal obstacle representations \cite{nageli2017real,joubert2016towards,huang2018act}, or a height map assumption \cite{bonatti2018autonomous}.

\begin{figure}[h!]
    \centering
    \includegraphics[width=1.0\textwidth]{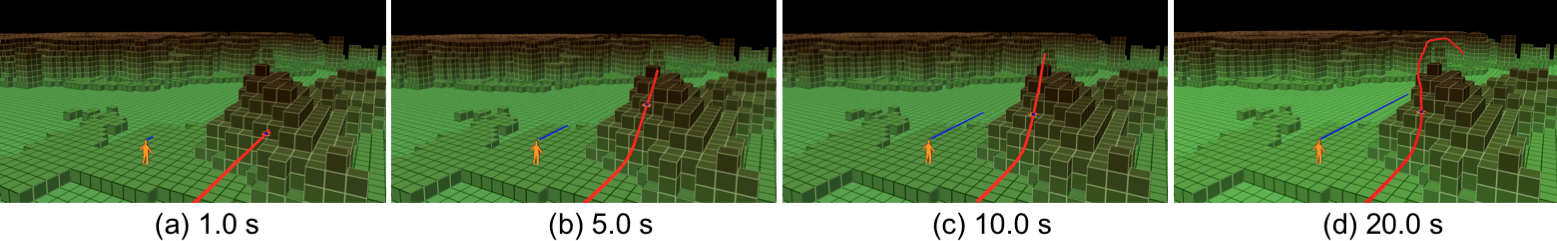}
    \caption{Planner behavior with different time horizons of $1$ (a), $5$ (b), $10$ (c), and $20$ (d) seconds for same actor trajectory and environment. The shortest time horizon of $1$s is not sufficient for the planner to find a trajectory that avoids the mound, and the vehicle gets stuck in a bad local minimum of solutions. Longer horizons let the UAV plan more intelligent trajectories, reasoning about obstacle shapes long before the UAV reaches those positions.}
    \label{fig:planning_horizons}
\end{figure}

\textit{Advantage of longer planning horizons:} We evaluate the overall system behavior when using different planning time horizons between $1$ and $20$ seconds, as seen in Table~\ref{tab:horizon}. Short horizons reason myopically about the environment, and cannot render robust and safe behavior in complex scenes, thus increasing the normalized cost per time length of the resulting trajectory. Figure~\ref{fig:planning_horizons} displays the qualitative difference between trajectories, keeping all variables except planning horizon constant.

\begin{table}[h!]
\centering
\caption{Performance of motion planner with varying planning time horizons for the environment shown in Fig~\ref{fig:planning_horizons}. Longer planning horizons allow better reasoning for safety and occlusion avoidance, lowering the normalized planning cost. However, longer horizons naturally increase planning computing time.}
\begin{tabular}{l|llllll}
\label{tab:horizon}
\textbf{Planning horizon length [s]:} & $\ $\textbf{1.0} & $\ $\textbf{5.0} & $\ $\textbf{10.0} & $\ $\textbf{20.0} \\
\hline
\text{Normalized trajectory cost [$J$/$t_f$]}  & 0.0334 & 0.0041 & 0.0028 & 0.0016\\
\text{Computing time [ms]}  & 0.0117 & 0.0131 & 0.0214 & 0.0343 \\
\hline
\end{tabular}
\end{table}

\begin{figure}[h!]
    \centering
    \includegraphics[width=0.8\textwidth]{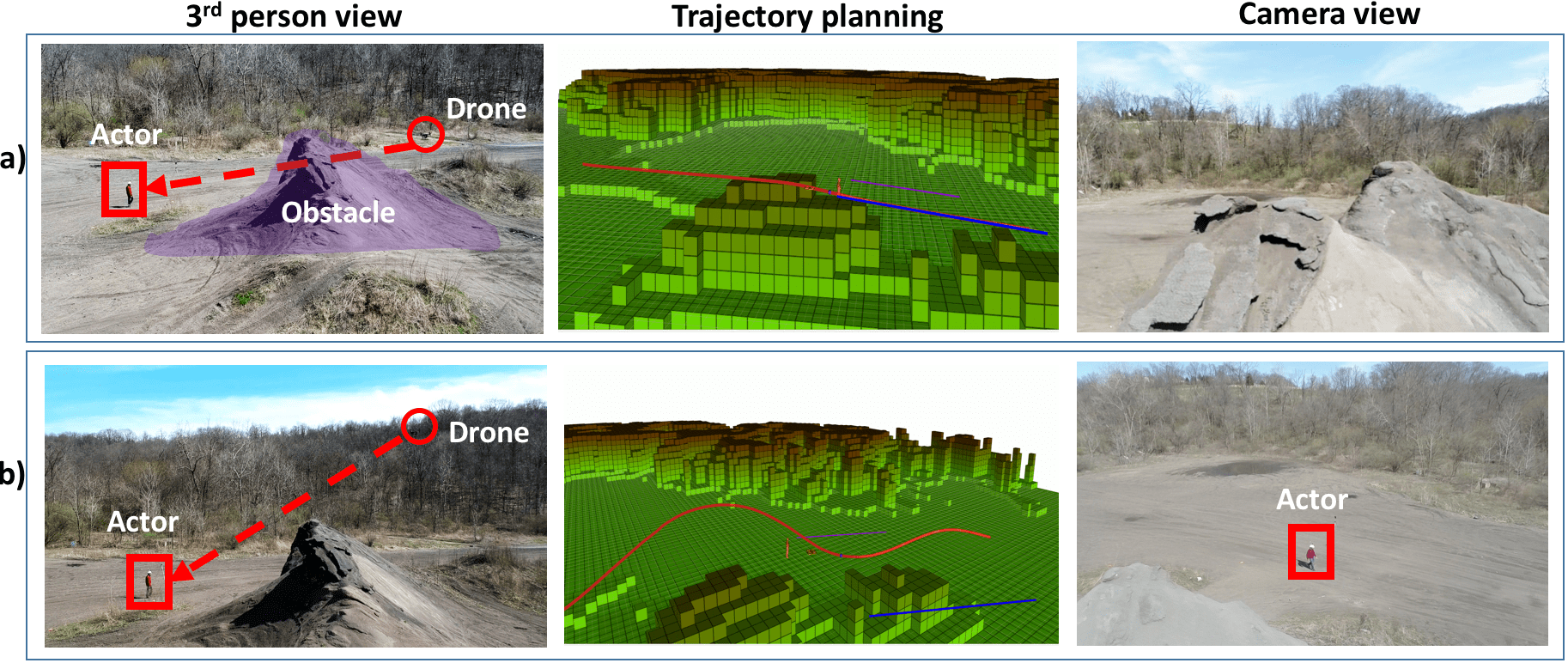}
    \caption{Comparison of planning a) without and b) with occlusion cost function in a special setup where actor positioning comes from a GPS tag. The occlusion cost function significantly improves the quality of the camera image in comparison with pure obstacle avoidance, for same shot type.}
    \label{fig:occlusion_3}
\end{figure}

\textit{Qualitative role of the occlusion cost function:} For this experiment, unlike the tests with visual actor localization, we detect the actor using a ground-truth GPS tag. We set up the motion planner to calculate the UAV paths with and without the occlusion cost function, keeping all other scenario variables equal. As seen in Figure~\ref{fig:occlusion_3}, our proposed occlusion cost significantly improves the aesthetics of the resulting image, keeping the actor visibility. In addition to aesthetics, maintaining actor visibility is a vital feature of our architecture, allowing vision-based actor localization.

\begin{figure}[h!]
    \centering
    \includegraphics[width=0.64\textwidth]{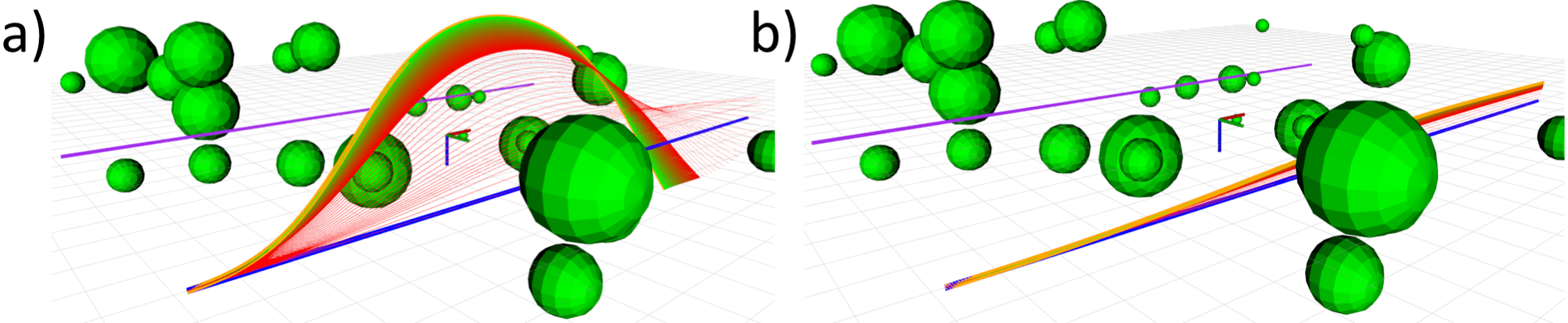}
    \caption{Randomized environment with obstacles to evaluate planner robustness. UAV trajectory initialization shown in blue, actor trajectory in purple, and planner iterations vary from red to green (final solution). a) Solution including occlusion cost function, and b) Pure obstacle avoidance. }
    \label{fig:stats}
\end{figure}

\textit{Quantitative role of the occlusion cost function:}
We evaluate our planning algorithm on environments with increasing levels of randomized clutter, as seen in Figure~\ref{fig:stats}. Table~\ref{tab:occlusion} summarizes the planner performance in different environments in terms of actor visibility and the average distance to the artistically desired trajectory. By using the occlusion cost function, we improve actor visibility by over 10\% in comparison with pure obstacle avoidance in environments with 40 random spheres; however, the trade-off is an increase in the average distance to the desired artistic trajectory.

\begin{table}[h!]
\centering
\caption{Evaluation of motion planner performance in the randomized environment from Fig~\ref{fig:stats}. Statistics computed using $100$ random configurations for each environment complexity level. }
\label{tab:occlusion}
\begin{tabular}{p{3cm}p{3cm}p{1.9cm}p{1.9cm}p{1.9cm}}
\noalign{\smallskip}
  & & \multicolumn{3}{c}{\textbf{Num. of spheres in environment}} \\
\textbf{Success metric} & \textbf{Cost functions} & $\qquad$ 1 & $\qquad$ 20 & $\qquad$ 40 \\
\noalign{\smallskip}\hline\noalign{\smallskip}
Actor visibility & $J_\mathrm{occ}+J_\mathrm{obs}$ & $99.4 \pm 2.2\%$ & $94.2 \pm 7.3\%$  & $86.9 \pm 9.3\%$  \\
along trajectory & $J_\mathrm{obs}$ & $98.8 \pm 3.0\%$ & $87.1 \pm 8.5\%$ & $75.3 \pm 11.8\%$ \\
\hline\noalign{\smallskip}
Avg. dist. to $\xi_\mathrm{shot}$, & $J_\mathrm{occ}+J_\mathrm{obs}$ & $0.4 \pm 0.4$ & $6.2 \pm 11.2$ & $10.7 \pm 13.2$ \\
in m & $J_\mathrm{obs}$ & $0.05 \pm 0.1$ & $0.3 \pm 0.2$ & $0.5 \pm 0.3$  
\end{tabular}
\end{table}

These detailed results allow us to draw insights into the planner performance under different conditions: it can operate smoothly, in full 3D maps, even under noise of real-time environment mapping and noisy actor inputs. We can also verify the importance of the efficient optimization algorithm for planning: by allowing longer time horizons, we can generate significantly higher quality plans. Finally, we demonstrate the essential role of our occlusion cost, which is defined for arbitrary obstacle shapes in the environment, in maintaining actor visibility.

\subsection{Artistic Shot Selection}
\label{subsec:experiments_arts}

Next we present detailed results on training and testing the artistic shot selection module, as well as experiments that provide insights to understand which artistic concepts this subsystem is learning. Table~\ref{tab:artistic_summary} summarizes our test objectives and results.

\begin{figure}[h!]
    \centering
    \includegraphics[width=0.9\columnwidth]{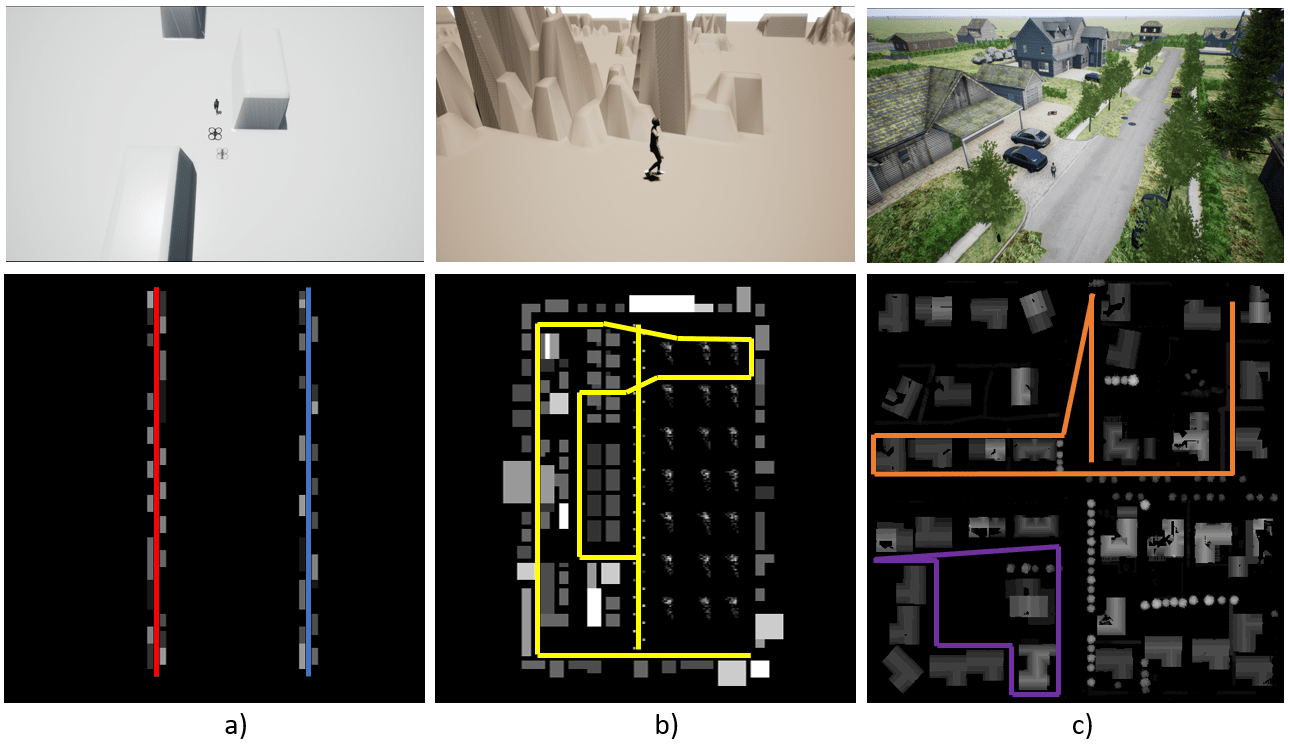}
    \caption{Rendering and height maps of AirSim environments with training routes. a) BlockWorld 1 (red), BlockWorld 2 (blue) b) Bigmap (yellow) c) Neighborhood 1 (orange), Neighborhood 2 (purple).}
    \label{heightmaps}
\end{figure}

We trained our agent exclusively in simulation, using the Microsoft AirSim release (see Section~\ref{subsec:sim_platform}).
We organize our environments in three categories:

\begin{itemize}
  \item \textit{BlockWorld}: It is generated from a height map, and the actor walks on a path with alternating blocks on the left and right sides. Blocks have varying heights and lengths (Figure~\ref{heightmaps}a).
  \item \textit{BigMap:} It is generated from a height map, and significantly more complex. It is separated into three zones: one that resembles the BlockWorld environment, a second zone with alternating pillars, and third zone with different shapes of mound-like structures (Figure~\ref{heightmaps}b).
  \item \textit{Neighborhood:} Unlike the two previous height maps, this environment is a photo-realistic rendering of a suburban residential area. The actor walks among structures like streets, houses, bushes, trees, and cars (Figure~\ref{heightmaps}c).
\end{itemize}

\begin{table}[h!]
\caption{Objectives and results for artistic shot selection.}
\centering
\begin{tabular}{l|l}
\hline   
\textbf{Objectives} & \textbf{Results}\\
\hline
\begin{tabular}[c]{@{}l@{}}Compare policy generalizability\\ to different environments\end{tabular} & 
\begin{tabular}[c]{@{}l@{}} Learned policies generalized well to new unseen\\ environments. Table~\ref{tab:rl_results_table} compares performances.\end{tabular}
\\
\hline
\begin{tabular}[c]{@{}l@{}}Analyze role of specific environment \\context in policy behavior\end{tabular} &
\begin{tabular}[c]{@{}l@{}} Policy learned to actively avoid potential occlusions\\ and switch often to keep video interesting (Figs.~\ref{fig:trajectory_rl_edited}-\ref{fig:heatmap})\end{tabular}
\\
\hline
\begin{tabular}[c]{@{}l@{}}Evaluate policies against baselines using \\ real human aesthetics in user study\end{tabular} &
\begin{tabular}[c]{@{}l@{}}Our policy outperformed constant shot types\\ or random actions (Table~\ref{tab:userstudy_results} and Fig.~\ref{fig:user_study_trajectories})  \end{tabular}
\\
\hline
\begin{tabular}[c]{@{}l@{}}Transfer policy learned in simulation \\ to real-life environments\end{tabular} &
\begin{tabular}[c]{@{}l@{}} We deployed the policy in additional \\field experiments (Fig.~\ref{fig:real_test_2}). \end{tabular}  
\\
\hline
\end{tabular}
\label{tab:artistic_summary}
\end{table}




\subsubsection{Learning an artistic policy}

\textit{Hand-crafted reward:} Using the hand-crafted reward definition from Section~\ref{sec:artistic}, we train a total of six policies in different environments. For all policies except \textit{Neighborhood 1 roam}, the actor walks along a pre-defined path. We define each episode as a concatenation of 5 consecutive time steps, each with a duration of $6$ seconds, and we train each policy between 300 to 2000 episodes, depending on the complexity of the environment:

\begin{itemize}
  \item \textit{BlockWorld 1 and 2:} Trained in different randomized BlockWorld environments, BW1 and BW2, for 300 episodes.
  \item \textit{BigMap:} Trained in a randomized BigMap environment, BM, for 1500 episodes.
  \item \textit{Neighborhood 1 and 2:} Trained in different sections of the Neighborhood environment, NH1 and NH2. Trained for 500 episodes.
  \item \textit{Neighborhood roam:} Trained in the entirety of the Neighborhood environment NH, with actor walking in random motion, for 2000 episodes.
\end{itemize}


We test all policies in all environments to evaluate generalizability. Table~\ref{tab:rl_results_table} summarizes the quantitative results. As expected, all policies perform better than random choice, and we achieve highest testing rewards in the same environments the policies were trained in. We also verify that the best generalization performance occurs when policies are trained and tested on the same environment category. It is interesting to note that policies trained on the Neighborhood environments tested on BigMap perform significantly better than those trained on BlockWorld, likely due to the simple geometry of the BlockWorld obstacles.

\begin{table}[h!]
\caption{Average reward per time step. As expected, policies have the highest rewards when trained and tested on the same environment (bold diagonal). The second-best policy for each environment is underlined and italicized.}
\label{tab:rl_results_table}
\begin{center}
\begin{tabularx}{\linewidth}{l | L L L L L}
\hline
\backslashbox{\textbf{Policy}}{\textbf{Test} \\ \textbf{Env.}} & \textbf{BW1} & \textbf{BW2} & \textbf{BM} & \textbf{NH1} & \textbf{NH2} \\
\hline
\textbf{BlockWorld 1} & \textbf{0.3444} & \underline{\textit{0.3581}} & 0.3635 & 0.3622 & 0.2985 \\
\textbf{BlockWorld 2} & \underline{\textit{0.3316}} & \textbf{0.3718} & 0.3918 & 0.4147 & 0.3673 \\
\textbf{BigMap} & 0.2178 & 0.2506 & \textbf{0.5052} & 0.5142 & 0.5760 \\
\textbf{Neighborhood 1} & 0.1822 & 0.1916 & 0.4311 & \textbf{0.5398} & \underline{\textit{0.5882}} \\
\textbf{Neighborhood 2} & 0.0813 & 0.1228 & \underline{\textit{0.4488}} & 0.4988 & \textbf{0.5897} \\
\textbf{Neighborhood 1 roam} & 0.1748 & 0.1779 & 0.4394 & \underline{\textit{0.5221}} & 0.5546 \\
\textbf{Random choice} & -0.0061 & 0.0616 & 0.1944 & 0.2417 & 0.2047 \\
\hline
\end{tabularx}
\end{center}
\end{table}

\begin{figure}[h!]
    \centering
    \includegraphics[width=0.4\columnwidth]{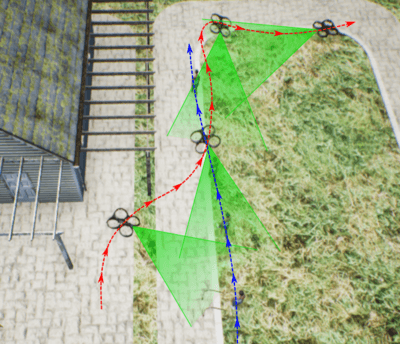}
    \caption{Time lapse of drone trajectory during filming in photo-realistic environment. Since the left hand side is occupied, the drone switches from left to front and then right shot.}
    \label{fig:trajectory_rl_edited}
\end{figure}

\begin{figure}[h!]
    \centering
    \includegraphics[width=0.8\columnwidth]{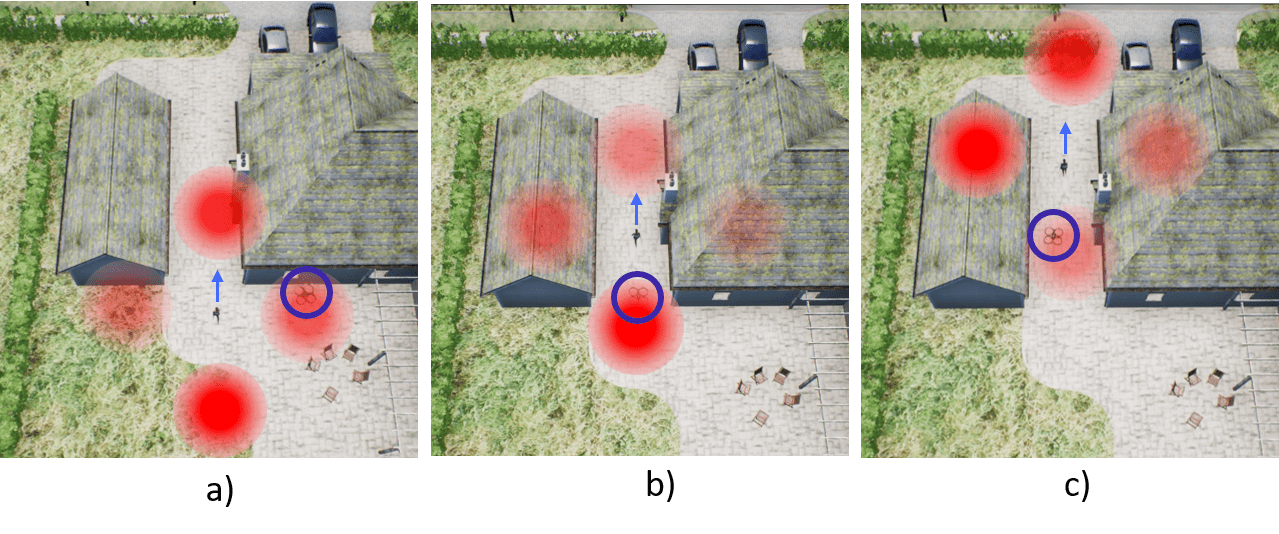}
    \caption{Visualization of the Q-values of the DQN during testing corresponding to the 4 shot types. The more opaque a circle is, the higher is the action's Q-value. The drone position before each decision is indicated by a blue ring around the drone. The drone starts on the right side of the actor and switches to back shot mode to traverse a narrow passage (a) where it stays in the actor's back for one time step (b). Finally it decides to switch to a left side shot once the obstacles are passed (c).}
    \label{fig:heatmap}
\end{figure}

Figures~\ref{fig:trajectory_rl_edited} and \ref{fig:user_study_trajectories} show examples of trajectories generated with trained policies. In addition, Figure~\ref{fig:heatmap} shows a heatmap with different actions, providing insights into the learned policy's behavior.
Qualitatively, we observe that the learned behavior matches the intended goals:

\begin{itemize}
    \item Keeps the actor in view by avoiding drastic shot mode switches between opposite positions such as left and right or front and back, which often cause visual loss of the actor;
    \item Switches shot types regularly to keep the viewer's interest;
    \item Avoids flying above high obstacles to keep the shot angle within desirable limits;
    \item Avoids high obstacles to maintain aircraft safety.
\end{itemize}

\textit{Human-generated rewards:} Using the interface described in Section~\ref{sec:artistic}, we use human aesthetics evaluations as rewards to train two new policies in the BlockWorld and Neighborhood environments for 300 and 500 episodes respectively. Comparisons between both reward schemes are presented next.

\subsubsection{User study results}

We asked ten participants to watch 30-seconds video clips taken from four different policies in five different scenes. Each participant ranked clips from most to least visually pleasing, and wrote open-ended comments on each clip. We chose two scenes from the BlockWorld and three from the Neighborhood environments, and compared the highest-performing handcrafted reward policy for each environment against the human-generated reward policy. In addition, we included a constant back shot policy and a random action policy for comparison.

\begin{figure}[h!]
    \centering
    \includegraphics[width=0.95\columnwidth]{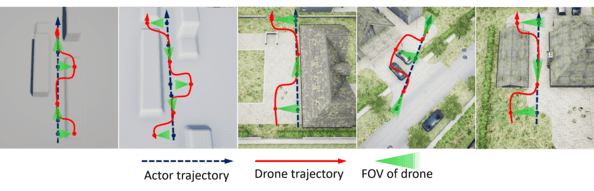}
    \caption{Drone trajectories of the highest rated policies in the user study in scenes 1 to 5.}
    \label{fig:user_study_trajectories}
\end{figure}

Table~\ref{tab:userstudy_results} summarizes the user study results, and Figure~\ref{fig:user_study_trajectories} shows the best-rated drone trajectories for each scene. As seen in Table~\ref{tab:userstudy_results}, the trained policies using both reward schemes have higher ratings in all scenes than the random or constant back shot policies. 

On average, the hand-crafted rewards trained policies which were ranked better than those trained with human-defined rewards, although participants' opinions were the opposite in some cases. We also summarize the participant's comments on the clips: 

\begin{itemize}
    \item All participants criticize the back shot policy as `boring' or `unexciting';
    \item All participants mention that the random policy loses view of the actor too often;
    \item The most common aesthetics complaint is due to loss of actor visibility;
    \item Participants often complained about too little camera movement when only one shot type is used for the entire $30$s clip. They also complained about camera movements being visually unpleasing when the shot type changes at every time step (every $6$s);
    \item Participants frequently mention that they like to see an overview of the surrounding environment, and not only viewpoints with no scene context where the actor's background is just a building or wall. Clips where the UAV provides multiple multiple viewpoints were positively marked;
    \item The human reward policy was often described as the most exciting one, while the handcrafted reward policy was described as very smooth.
\end{itemize}

\begin{table}[h!]
\caption{Average normalized score of video clips between 0 (worst) and 10 (best).}
\centering
\setlength{\tabcolsep}{0.5\tabcolsep}
\begin{tabular}{l|cccccc}
\hline
 & \textbf{Average} & \textbf{Scene 1} & \textbf{Scene 2} & \textbf{Scene 3} & \textbf{Scene 4} & \textbf{Scene 5}\\
\hline
\textbf{Hand-crafted reward} & \textbf{8.2} & \textbf{10.0} & 5.3 & \textbf{9.3} & \textbf{7.7} & \textbf{8.7} \\
\textbf{Human reward} & 7.1 & 5.0 & \textbf{9.0} & 6.0 & \textbf{7.7} & 8.0 \\
\textbf{Back shot} & 3.8 & 4.0 & 4.7 & 4.3 & 4.0 & 2.0\\
\textbf{Random} & 0.9 & 1.0 & 1.0 & 0.3 & 0.7 & 1.3\\
\hline
\end{tabular}
\label{tab:userstudy_results}
\end{table}

The main perceived difference between the handcrafted reward policy and the human reward policy was the consistency of switching shots. 
While the former tries to switch shots every 2 times steps ($12$s) if not disturbed by an obstacle, the latter is more irregular in timing. From the user studies it is evident that a more regular period that is neither too short nor too long improves aesthetic scores.

\subsubsection{Extended results in field experiments}

\begin{figure}[h!]
    \centering
    \includegraphics[width=0.7\columnwidth]{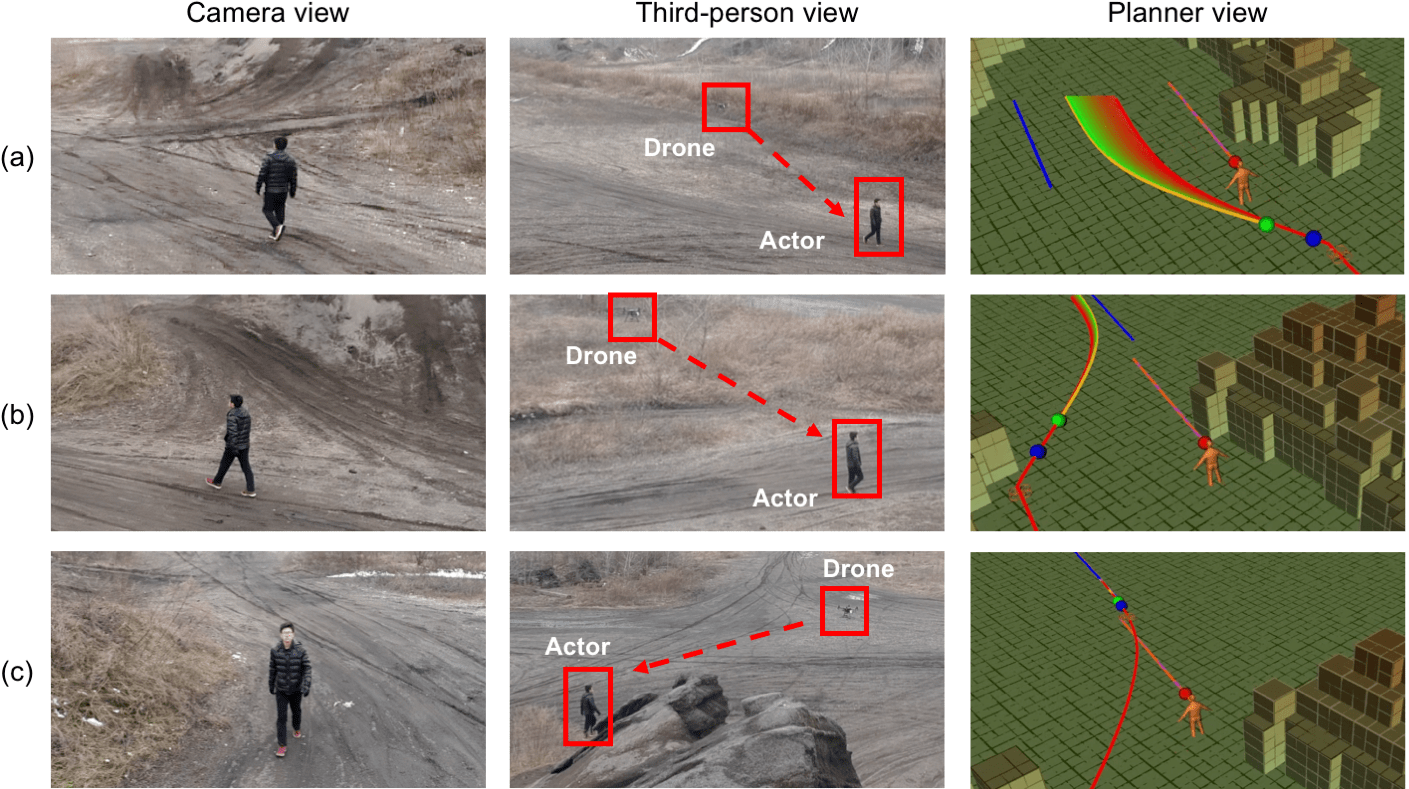}
    \caption{UAV follows an actor while switching shot types autonomously with our trained policy: a) UAV starts at the back of the actor; b) The mound's presence on the right side of the actor leads to a left shot selection; c) UAV switches to a frontal shot in the open area.}
    \label{fig:real_test_2}
\end{figure}

We tested our trained policies in real-life settings. Since these tests were only focused on the shot selection module we operated using a pre-mapped environment, differently than the integrated results from Subsection~\ref{subsec:experiments_integrated} that uses online mapping. We filmed scenes using three distinct policies: one trained with handcrafted rewards on the \textit{BigMap} environment, a second fixed back shot policy and a third random policy. Figure \ref{fig:real_test_2} summarizes the results.




Similar to the simulation results, the random policy results in constant loss of actor due to drastic position changes and close proximity to obstacles. For example, a random action of right shot will cause the UAV to climb much above the actor if it is too close to a large mound. The back shot policy, although stable in vehicle behavior, results in visually unappealing movies. Finally, our trained policy provides a middle ground, resulting in periodic changes in UAV viewpoint, while providing stable visual tracking.



\section{Discussion}
\label{sec:discussion}

In this section we discuss lessons learned throughout the development of our work, and also present comments on how our methods can be used with different types of sensors and UAV platforms.




\subsection{Lessons Learned}

During development of our autonomous aerial cinematographer platform we learned several lessons and gained insights into problem specificities, which we summarize below. We expect these lessons to not only be useful to researchers in the field of aerial vehicles, but also to generalize to other related areas.

\paragraph{Cascading errors can destabilize the robot:} Estimation errors in a module get amplified if used downstream in decision making. For example, we learned that jerky UAV movements lead to mis-registration of camera pose, which leads to poor actor detection. This in turn leads to poor actor prediction, which can be off by meters. Unlike previous works that operate under highly precise motion capture systems, in real scenarios we observe that controlling the camera orientation towards the position estimates causes the robot to completely lose the actor. To mitigate this effect, we chose to decouple motion planning, which uses the actor world projection estimates, from camera control, which uses only object detections on the current image as feedback. To validate the quality of both threads independently, we performed statistical performance evaluations, as seen throughout Section~\ref{sec:experiments}. 




\paragraph{Long-range sensors are beneficial to planning performance:} The planner relies on the online map. Slow online map updates slow down the planner. This typically happens when the robot moves near large pockets of unknown areas, which triggers large updates for the TSDT. We learned that a relatively long range LiDAR sensor maps out a significantly larger area. Hence almost always, the area near the robot is mapped out fully and the planner does not have to wait for map updates to enter an unknown region. While our system used a relatively large map of $250\times250\times100$m, significantly faster mapping-planning frequencies could be achieved with the use of a smaller local map. Such change would likely be necessary with the use of shorter-range sensors like stereo pairs, which are common in commercial aircrafts due to their reduced price.

\paragraph{Semi-supervised methods can improve learning generalization:} While deep learning methods are ubiquitous in computer vision, they rely on massive amounts of labeled data due to the complexity of the model class. Moreover, these models do not generalize to varying data distributions. We learned that one can reduce sample complexity by enforcing regularization / additional structure. In our case, we enforce temporal continuity on the network output. We show that a combination of \emph{small labeled} dataset for supervisory loss and a \emph{large unlabeled} dataset for temporal continuity loss is enough to solve the problem. Exploring other regularization techniques such as consistency between different sensory modalities is also an interesting area to be investigated in the future. 

\paragraph{Height estimate using IMU and barometer is not enough for long operations:} 
During extended vehicle operations (over $5-10$ minutes), we learned that the UAV's height calculated by fusing IMU and barometer data drifts significantly, especially after large vertical maneuvers. Inaccurate height estimates degrade pointcloud registration, thereby degrading the overall system performance. In the future, we plan to use LiDAR or visual SLAM to provide more accurate 3D localization. 

\paragraph{Real-world noise reduces transferability of the artistic policy trained in simulation:} 
The noise present in real-world testing conditions, in particular for the map registration and actor localization, affected the results generated by the policy that was trained purely in simulation using ground-truth data. Shot selection in simulation highly prioritized viewpoints that drew the UAV away from tall obstacles, while in deployment we observed that the drone avoided proximity to tall objects with a significantly smaller frequency. In the future we will consider artificially adding noise to simulations for better transferability, or training the artistic policy with a combination of simulated and real-life data.

\subsection{Adapting Our Work to Different UAVs and Sensors}

In this work we employed a long-range LiDAR sensor for mapping the environment, and a DJI M210 UAV as the base platform. Even though we used relatively standard robotics development platform and sensor, researchers and developers who work on problems similar to aerial cinematography may face different constraints in terms of payload capacity, vehicle size, sensor modalities and costs. We argue that our problem formulation can be easily extended to other contexts.

First, we argue that the LiDAR sensor used in the online mapping module (Section~\ref{sec:mapping}) can be replaced by other sensors. Stereo cameras and depth sensors, for example, can be light-weight and significantly cheaper alternatives. The incoming hits for the mapping pipeline can then be acquired by using each pixel from a depth image, or each match from the stereo pair. The main advantage of LiDAR is the relatively long range, in the order of hundreds of meters. When using lighter sensors, the developer needs to take into account the new sensor range in order to keep the UAV safe. They must consider the expected vehicle speed and the scale of obstacles in the environment, so that the planner can reason about obstacles far from the UAV's current position.

In addition, our system architecture is platform-agnostic. Our methods can easily be adapted to smaller or potentially cheaper platforms. To do so, one only needs to care for the software interface between the trajectory controller and the aircraft's internal attitude or velocity controller.

In the future, we hope to see our architecture extended to other UAV types: our framework is not constrained to uniquely multi-rotors. With the appropriate changes in the motion planner's trajectory parametrization and cost functions, our pipeline can also be employed by fixed-wing or hybrid aircrafts. More generally, despite the lower path dimensionality, even ground robots can employ the same methodology for visually tracking dynamic targets.

\section{Conclusion}
\label{sec:conclusion}

In this work we presented a complete system for robust autonomous aerial cinematography in unknown, unstructured environments while following dynamic actors in unscripted scenes. Current approaches do not address all theoretical and practical challenges present in real-life operation conditions; instead, they rely on simplifying assumptions such as requiring ground truth actor position, using prior maps of the environment, or only following one shot type specified before flight. To solve the entirety of the aerial cinematography, our system revolves around two key ideas. First, we frame the filming task as an efficient cost optimization problem, which allows trajectories with long time horizons to be computed in real time, even under sensor noise. Second, instead of using hand-defined heuristics to guide the UAV's motion, we approach the artistic decision-making problem as a learning problem that can directly use human feedback.

We developed a system with four modules that work in parallel. (1) A vision-based actor localization module with motion prediction. (2) A real-time incremental mapping algorithm using a long-range LiDAR sensor. (3) A real-time optimization-based motion planner that exploits covariant gradients to efficiently calculate safe trajectories with long time horizons while balancing artistic objectives and occlusion avoidance for arbitrary obstacle shapes. (4) Finally, a deep reinforcement learning policy for artistic viewpoint selection. 

We offered extensive detailed experiments to evaluate the robustness and real-time performance of our system both in simulation and real-life scenarios. These experiments occur among a variety of terrain types, obstacle shapes, obstacle complexities, actor trajectories, actor types (i.e., people, cars, bikes) and shot types.

Based on our results, we identify several key directions for possible future work. One clear direction is the extension of our theory to multi-drone, multi-actor scenarios. This improvement can be achieved by the addition of new cost functions that penalize inter-drone collisions, inter-drone sight, and a metric for multi-actor coverage. In addition, multi-actor scenarios require a slight modification in the definition of artistic parameters that define the desired artistic shot for our motion planner. Another interesting direction to follow lies among the reconstruction of dynamic scenes. While systems such as the CMU PanOptic Studio \cite{joo2015panoptic} can precisely reconstruct scenes volumetrically in indoor and static scenarios, to our knowledge, no current system offers good volumetric reconstruction of dynamic scenes in natural environments in real-life conditions. Lastly, we envision further research in learning the artistic reasoning behind human choices. More broadly, as robotics evolves, autonomous agents are required to operate in a large variety of tasks in proximity to humans, where success is in great part measured by the ability of the robot to execute \textit{aesthetic} and \textit{human-like} behaviors. We identify important related areas to cinematography, such as autonomous driving and human-robot interaction, where fine nuances of human behavior modeling are important for the development of autonomous agents.

\subsubsection*{Acknowledgments}
We thank Xiangwei Wang and Greg Armstrong for the assistance in field experiments and robot construction. Research presented in this paper was funded by Yamaha Motor Co., Ltd. under award \#A019969.

\bibliographystyle{apalike}
\bibliography{IEEEexample}

\clearpage

\begin{appendices}

\section{Derivations of planning cost functions}
\label{sec:appendix_costs}

\subsection{Smoothness cost}
\label{subsec:appendix_smooth}

We can discretize Equation~\ref{eq:smooth_cont} to compute the smoothness for $\Path{q}$:

\begin{equation}
  \label{eq:smooth_discrete}
      \costFnSmooth{\Path{q}} = 
       \frac{1}{(n-1)} \frac{1}{2} \sum_{t=1}^{n-1} 
       \left [ 
       \alpha_0 \abs{\frac{p_{t}-p_{t-1}}{\Delta t}}^2 + 
       \alpha_1 \abs{\frac{\dot{p_{t}}-\dot{p_{t-1}}}{\Delta t}}^2 + 
       \alpha_2 \abs{\frac{\ddot{p_{t}}-\ddot{p_{t-1}}}{\Delta t}}^2 + \dots
       \right ] 
\end{equation}

To simplify Equation~\ref{eq:smooth_discrete}, we define: $\Delta t = \frac{t_f}{n-1}$, a finite differentiation operator $K$, and an auxiliary matrix $e$ for the contour conditions:

\begin{equation}
  \begin{aligned}
    K_{(n-1) \times (n-1)} = 
    \begin{bmatrix}
      1&0&0&\cdots&0&0&0\\-1&1&0&\cdots&0&0&0\\0&-1&1&\cdots&0&0&0\\&&&\vdots&&&\\0&0&0&0\cdots&0&-1&1
    \end{bmatrix}
    \quad
    e_{(n-1)\times3} = 
    \begin{bmatrix}
      -p_{0x} & -p_{0y} & -p_{0z}\\
      0&0&0\\\vdots&\vdots&\vdots\\0&0&0
    \end{bmatrix}
  \end{aligned}
\end{equation}

By manipulating the terms $K$, $e$, and $\Delta t$ we obtain the auxiliary terms:

\begin{equation}
\begin{aligned}
  &K_0 = \frac{K}{\Delta t}, \quad &e_0 = \frac{e}{\Delta t}\\
  &K_1 = \frac{K^2}{\Delta t^2}, \quad &e_1 = \frac{K e}{\Delta t^2} + \frac{\dot{e}}{\Delta t^2}\\
  &K_2 = \frac{K^3}{\Delta t^3}, \quad &e_2 = \frac{K^2 e}{\Delta t^3} + \frac{K \dot{e}}{\Delta t^3} + \frac{\ddot{e}}{\Delta t^3}
\end{aligned}
\end{equation}

\begin{equation}
\begin{aligned}
  &A_0 = K_0^TK_0, \quad b_0 = K_0^Te_0, \quad c_0 = e_0^Te_0\\
  &A_1 = K_1^TK_1, \quad b_1 = K_1^Te_1, \quad c_1 = e_1^Te_1\\
  &A_2 = K_2^TK_1, \quad b_2 = K_2^Te_2, \quad c_2 = e_2^Te_2
\end{aligned}
\end{equation}

Finally, we can analytically write the smoothness cost as a quadratic objective: 

\begin{equation}
\begin{aligned}
 \costFnSmooth{\Path{q}} &= 
 \frac{1}{2(n-1)} Tr(\Path{q}^T A_\mathrm{smooth} \Path{q} + 2\Path{q}^T b_\mathrm{smooth} + c_\mathrm{smooth}) \\
 \text{where:} \quad & A_\mathrm{smooth} = \alpha_0 A_0 + \alpha_1 A_1 + \dots\\
 & b_\mathrm{smooth} = \alpha_0 b_0 + \alpha_1 b_1 + \dots\\
 & c_\mathrm{smooth} = \alpha_0 c_0 + \alpha_1 c_1 + \dots
\end{aligned}
\end{equation}

Since $\costFnSmooth{\Path{q}}$ is quadratic, we find analytic expressions for its gradient and Hessian. Note that the Hessian expression does not depend on the current trajectory, which is a property used serve to speed up the optimization algorithm described in Subsection~\ref{subsec:opt_algorithm}:

\begin{equation}
\begin{aligned}
  \nabla \costFnSmooth{\Path{q}} &= \frac{1}{(n-1)}(A_\mathrm{smooth}\Path{q}+b_\mathrm{smooth})\\
  \nabla^2 \costFnSmooth{\Path{q}} &= \frac{1}{(n-1)} A_\mathrm{smooth}
\end{aligned}
\end{equation}

\subsection{Shot quality cost}
\label{subsec:appendix_shotqual}

We can calculate $\costShot$ in the discrete form:

\begin{equation}
    \costFnShot{\Path{q},\Path{a}} = \frac{1}{(n-1)} \frac{1}{2} \sum_{t=1}^{n} \abs{p_{t}-p_\mathrm{t\ shot}}^2
\end{equation}

By defining auxiliary matrices, we can also define a quadratic expression:

\begin{equation}
  \costFnShot{\Path{q},\Path{a}} = 
  \frac{1}{2(n-1)} Tr(\Path{q}^T A_\mathrm{shot} \Path{q} + 2\Path{q}^T b_\mathrm{shot} + c_\mathrm{shot}), \\
\end{equation}

where:

\begin{equation}
\begin{aligned}
  &K_\mathrm{shot} = -I_{(n-1) \times (n-1)}\\
  &e_\mathrm{shot} = \Path{\mathrm{shot}} = 
  \begin{bmatrix}
    p_\mathrm{1x\ shot} & p_\mathrm{1y\ shot} & p_\mathrm{1z\ shot}\\
    p_\mathrm{1x\ shot} & p_\mathrm{1y\ shot} & p_\mathrm{1z\ shot}\\
    \vdots&\vdots&\vdots\\
    p_\mathrm{(n-1)x\ shot} & p_\mathrm{(n-1)y\ shot} & p_\mathrm{(n-1)z\ shot}\\
  \end{bmatrix},
\end{aligned}
\end{equation}

and:
\begin{equation}
  A_\mathrm{shot} = K_\mathrm{shot}^TK_\mathrm{shot}, \quad
  b_\mathrm{shot} = K^Te_\mathrm{shot}, \quad
  c_\mathrm{shot} = e_\mathrm{shot}^Te_\mathrm{shot}
\end{equation}

We can again find analytic expressions for the shot quality gradient and Hessian, which is independent from the current trajectory:

\begin{equation}
\begin{aligned}
  \nabla \costFnShot{\Path{q}} &= \frac{1}{(n-1)}(A_\mathrm{shot}\Path{q}+b_\mathrm{shot})\\
  \nabla^2 \costFnShot{\Path{q}} &= \frac{1}{(n-1)} A_\mathrm{shot}
\end{aligned}
\end{equation}
\end{appendices}

\end{document}